\documentclass[journal]{IEEEtran}

\usepackage{graphicx}

\usepackage{amsmath}
\usepackage{amssymb}
\usepackage{amsfonts}

\usepackage{booktabs}
\usepackage{subfig}
\usepackage{floatflt}
\usepackage[linesnumbered, ruled,vlined]{algorithm2e}

\usepackage{moresize}
\usepackage{url}

\usepackage{tablefootnote}
\usepackage{bigfoot}

\usepackage{colortbl}
\usepackage{tabularx}

\def\vector#1{\mbox{\boldmath $#1$}}

\def\shyp{\mathchar`-}

\ifCLASSINFOpdf
\else
\fi
\begin{document}
%



\title{An Analysis of Quality Indicators Using Approximated Optimal Distributions in a Three-dimensional Objective Space}






%
%
%

\author{
Ryoji~Tanabe,~\IEEEmembership{Member,~IEEE,}and~Hisao~Ishibuchi,~\IEEEmembership{Fellow,~IEEE}
\thanks{R. Tanabe and H. Ishibuchi are with
Shenzhen Key Laboratory of Computational Intelligence, University Key Laboratory of Evolving Intelligent Systems of Guangdong Province, Department of Computer Science and Engineering, Southern University of Science and Technology, Shenzhen 518055, China. e-mail: (rt.ryoji.tanabe@gmail.com, hisao@sustech.edu.cn). (Corresponding author: Hisao Ishibuchi)}
}
\maketitle


\begin{abstract}

Although quality indicators play a crucial role in benchmarking evolutionary multi-objective optimization algorithms, their properties are still unclear.
One promising approach for understanding quality indicators is the use of the optimal distribution of objective vectors that optimizes each quality indicator.
However, it is difficult to obtain the optimal distribution for each quality indicator, especially when its theoretical property is unknown.
Thus, optimal distributions for most quality indicators have not been well investigated.
To address these issues, first, we propose a problem formulation of finding the optimal distribution for each quality indicator on an arbitrary Pareto front.
Then, we approximate the optimal distributions for nine quality indicators using the proposed problem formulation.
We analyze the nine quality indicators using their approximated optimal distributions on eight types of Pareto fronts of three-objective problems.
Our analysis demonstrates that uniformly-distributed objective vectors over the entire Pareto front are not optimal in many cases.
Each quality indicator has its own optimal distribution for each Pareto front.
We also examine the consistency among the nine quality indicators.

\end{abstract}


%
\IEEEpeerreviewmaketitle

\section{Introduction}
\label{sec:introduction}

\IEEEPARstart{M}ULTI-OBJECTIVE optimization problems (MOPs) appear in real-world applications.
Since no solution can simultaneously optimize multiple objective functions in general, the goal of MOPs is usually to find a Pareto optimal solution preferred by a decision maker \cite{Miettinen98}.
When no information about the decision maker's preference is available, an ``a posteriori'' decision making is performed.
The decision maker selects a final solution from a non-dominated solution set that approximates the Pareto front in the objective space.
An evolutionary multi-objective optimization algorithm (EMOA) is frequently used for the ``a posteriori'' decision making \cite{Deb01}.

In multi-objective optimization, a vector that consists of all objective values of a solution is referred to as an objective vector.
In this paper, we are interested in the distribution of objective vectors in the objective space, rather than the distribution of solutions in the solution space.
Two terms ``objective vector'' and ``solution'' are used synonymously in some previous studies (e.g., \cite{IshibuchiISN18,IshibuchiISN18ecj,LiY19}). For the sake of clarity, we distinguish the two terms.
That is, solutions in the objective space are always referred to as objective vectors in this paper. 
We also assume that the number of objective vectors is bounded by the population size $\mu$.

Most quality indicators evaluate the quality of an objective vector set in terms of at least one of the three criteria (the convergence, the uniformity, and the spread) \cite{KnowlesC02,ZitzlerTLFF03,OkabeJS03,LiY19}.
Although a number of quality indicators have been proposed in the literature, we focus on unary quality indicators that map $\mu$ objective vectors to a real number.
Representative quality indicators include hypervolume (HV) \cite{ZitzlerT98}, the additive $\epsilon$-indicator ($I_{\epsilon+}$) \cite{ZitzlerTLFF03}, and inverted generational distance (IGD) \cite{CoelloS04}.
Since the performance of EMOAs is discussed based on the quality of obtained objective vector sets, quality indicators play a central role in benchmarking EMOAs.
Although this paper does not address the cardinality of an objective vector set, it does not mean that the cardinality is less important.
As pointed out in \cite{LiY19}, the cardinality is one of important aspects when the size of the Pareto front is relatively small (e.g., combinatorial MOPs \cite{IshibuchiM98}).
Since we address only non-dominated objective vectors on the Pareto front of a continuous MOPs in this paper, we do not analyze the cardinality-based quality indicators.

One of the critical issues in quality indicators is that their properties are not always obvious.
Thus, some quality indicators are likely to provide misleading information about the quality of objective vector sets.
HV is maximized by non-uniform objective vectors when the Pareto front is nonlinear \cite{AugerBBZ09}.
Although the original generalized spread ($\Delta$) for bi-objective problems \cite{DebAPM02} can evaluate the uniformity of given objective vectors, its extended version for problems with more than two objectives \cite{WangWY10} overestimates the uniformity of non-uniform objective vectors in some cases \cite{JiangOZF14,LiY19}.





Since understanding the properties of quality indicators is necessary for the development of EMOAs in the right direction, some analysis studies have been presented in the literature.
One of the most promising approaches is the use of the optimal distribution of objective vectors that optimizes a given quality indicator.
A distribution of $\mu$ objective vectors is said to be the optimal $\mu$-distribution if it optimizes a quality indicator \cite{AugerBBZ12}.
The optimal $\mu$-distribution can explain which distribution of objective vectors is highly evaluated by each quality indicator.
For example, the optimal $\mu$-distribution for IGD does not contain objective vectors on the edges of the linear Pareto front in some cases \cite{IshibuchiISN18}.
Thus, IGD may overestimate an objective vector set with a poor spread.
The optimal $\mu$-distribution helps an indirect analysis of each quality indicator in this manner.






Although the optimal $\mu$-distributions for HV and IGD can be obtained on the linear Pareto front of two objective problems \cite{AugerBBZ09,IshibuchiISN18}, it is difficult to obtain the optimal $\mu$-distributions for other quality indicators in an exact manner, especially problems with more than two objectives.
Thus, approximated optimal $\mu$-distributions are used for the analysis of quality indicators.
In general, an approximated optimal $\mu$-distribution is obtained in an analytical or empirical manner \cite{AugerBBZ09,BrockhoffWT12,Glasmachers14,BringmannFK15,IshibuchiISN18}.
However, existing analytical and empirical approaches have some limitations.
Analytical approaches utilize some specific properties of a quality indicator (e.g., the decomposability of the HV calculation \cite{Glasmachers14}).
For this reason, analytical approaches can be applied only to quality indicators whose theoretical properties are clear (e.g., HV for two- and three-objective problems \cite{AugerBBZ09,Glasmachers14} and $I_{\epsilon+}$ only for two-objective problems \cite{BringmannFK15}).
Empirical approaches translate the problem of finding the optimal $\mu$-distribution on the Pareto front into a black-box single-objective continuous optimization problem.
Then, the optimal $\mu$-distribution is approximated using a derivative-free optimizer.
For example, the optimal $\mu$-distribution for R2 was approximated  by CMA-ES \cite{HansenO01} in \cite{BrockhoffWT12}.
Although empirical approaches are more flexible than analytical approaches, empirical approaches have been applied to only a few quality indicators, including HV \cite{AugerBBZ09} and R2 \cite{BrockhoffWT12}.
In addition, most previous studies addressed only two-objective problems.
This is because it is not obvious how to formulate the problem of finding the optimal $\mu$-distribution on a multi-objective problem with more than two objectives.
For these reasons, the (approximated) optimal $\mu$-distributions for most quality indicators on problems with more than two objectives have not been well analyzed in the literature.

To address these issues, first, we propose a problem formulation of finding the optimal $\mu$-distribution for each quality indicator on a Pareto front in an arbitrary-dimensional objective space.
Then, we approximate the optimal $\mu$-distribution for each of the following nine quality indicators on a Pareto front in a three-dimensional objective space: HV, IGD, IGD plus (IGD$^+$) \cite{IshibuchiMTN15}, R2, new R2 (NR2) \cite{ShangIZL18}, $I_{\epsilon+}$, $s$-energy (SE) \cite{HardinS04}, $\Delta$, and pure diversity (PD) \cite{WangJY17}.
We analyze the nine quality indicators using their approximated optimal $\mu$-distributions on eight types of Pareto fronts.

Our main contributions in this paper are as follows:

\begin{enumerate}
\item We propose a general problem formulation of finding the optimal $\mu$-distribution. In contrast to the existing empirical approaches, the proposed formulation can handle any number of objectives. The proposed formulation is applicable to any types of the Pareto front (under the condition that its front shape function is given).
Note that the formulation proposed in \cite{AugerBBZ09} and our formulation can be categorized into set-based optimization \cite{ZitzlerTB10}.
Details are discussed in Section \ref{sec:proposed_method}.


\item We approximate the optimal $\mu$-distributions for the nine quality indicators on the Pareto fronts of eight problems with three objectives.
  This is the first study to present the approximated optimal $\mu$-distributions for NR2, $\Delta$, SE, PD, R2, and $I_{\epsilon+}$ for three-objective problems.  
\item We analyze the nine quality indicators from the viewpoint of the approximated optimal $\mu$-distributions.
  We provide insightful information about the nine quality indicators.
  For example, PD may overestimate an objective vector set with a small dissimilarity.
  Our observations are summarized in Section \ref{sec:conclusion}.


\end{enumerate}



The rest of this paper is organized as follows.
Section \ref{sec:preliminaries} provides some preliminaries.
Section \ref{sec:related_work} describes related studies.
Section \ref{sec:proposed_method} proposes a problem formulation to search for the optimal distribution of objective vectors.
Section \ref{sec:experimental_settings} describes the setting of our computational experiments.
Section \ref{sec:experimental_results} shows analysis results.
Section \ref{sec:conclusion} concludes this paper.


%


\section{Preliminaries}
\label{sec:preliminaries}




\subsection{Definition of multi-objective optimization problems (MOPs)}
\label{sec:def_MOPs}

We suppose a multi-objective minimization problem in this paper.
A continuous MOP is to find a solution $\vector{x} \in \mathbb{S}$ that minimizes a given objective function vector $\vector{f}: \mathbb{R}^n \rightarrow \mathbb{R}^m$.
Here, $\mathbb{S} \subseteq \mathbb{R}^n$ is the $n$-dimensional solution space, and $\mathbb{R}^m$ is the $m$-dimensional
objective space.
Thus, $n$ is the number of decision variables, and $m$ is the number of objective functions.


A solution $\vector{x}_1$ is said to dominate $\vector{x}_2$ iff $f_i (\vector{x}_1) \leq f_i (\vector{x}_2)$ for all $i \in \{1, ..., m\}$ and $f_i (\vector{x}_1) < f_i (\vector{x}_2)$ for at least one index $i$.
In addition, $\vector{x}_1$ is said to weakly dominate $\vector{x}_2$ iff $f_i (\vector{x}_1) \leq f_i (\vector{x}_2)$ for all $i \in \{1, ..., m\}$.
If $\vector{x}^*$ is not dominated by any other solutions in $\mathbb{S}$, $\vector{x}^*$ is a Pareto optimal solution.
The set of all $\vector{x}^*$ is the Pareto optimal solution set.
The set of all $\vector{f}(\vector{x}^*)$ is the Pareto front.
The goal of MOPs for the ``a posteriori'' decision making is to find a non-dominated solution set that approximates the Pareto front in the objective space.


\subsection{Quality indicators}
\label{sec:q_indicators}



For the sake of simplicity, we denote the objective vector $\vector{f}(\vector{x}) = \bigl(f_1(\vector{x}), ..., f_m(\vector{x})\bigr)^{\rm T}$ as $\vector{a} = (a_1, ..., a_m)^{\rm T}$, where $a_i=f_i(\vector{x})$ for $i \in \{1, ..., m\}$.
Let $\vector{A} = \{\vector{a}_1, ..., \vector{a}_{\mu}\}$ be a set of $\mu$ objective vectors, where $\mu$ is the population size.


Let $\Omega$ be a set of all possible objective vector sets.
A unary quality indicator $I: \Omega \rightarrow \mathbb{R}$ evaluates the quality of $\vector{A}$ in terms of at least one of the convergence, the uniformity, and the spread.
$I$ is said to be Pareto-compliant iff the ranking of all objective vector sets in $\Omega$ by $I$ is consistent with the Pareto dominance relation (for details, see \cite{KnowlesTZ06}).
In this paper, we define the convergence, the uniformity, and the spread as follows.
The convergence of $\vector{A}$ means how close objective vectors in $\vector{A}$ are to the Pareto front.
The uniformity of $\vector{A}$ means how uniform the distribution of objective vectors in $\vector{A}$ is.
The spread of $\vector{A}$ means how well objective vectors in $\vector{A}$ cover the entire Pareto front.
As pointed out in \cite{LiY19}, the spread is not equal to the extensity, which represents the length of the boundaries of $\vector{A}$.  
A combination of the uniformity and the spread is referred to as ``diversity'' in the evolutionary multi-objective optimization (EMO) community.
Fig. S.1 in the supplementary file shows examples of distributions of objective vectors in the three cases on the linear Pareto front with $m=2$.

Below, we explain the following nine quality indicators used in this paper: HV, IGD, IGD$^+$, R2, NR2, $I_{\epsilon+}$, SE, $\Delta$, and PD.
We do not consider convergence-based quality indicators, such as generational distance (GD) \cite{VeldhuizenL98}, which evaluate the quality of an objective vector set $\vector{A}$ in terms only of the convergence.
This is because the optimal $\mu$-distribution for such a quality indicator is obvious.
$\vector{A}$ is optimal for convergence-based quality indicators if all elements in $\vector{A}$ are on the Pareto front regardless of their distribution.
For the same reason, we do not consider cardinality-based quality indicators, such as overall non-dominated vector generation (ONVG) \cite{Veldhuizen99}, which evaluate the quality of $\vector{A}$ based on the number of non-dominated solutions.
For example, all of $\vector{A}_1$, $\vector{A}_2$, and $\vector{A}_3$ in Fig. S.1 are optimal for both GD and ONVG.


Below, $\vector{R}$ is a set of $m$-dimensional reference vectors that approximate the Pareto front in the objective space.
The reference point $\vector{q} = (q_1, ..., q_m)^{\rm T} \in \mathbb{R}^m$ is used for the HV and NR2 calculations.
It should be noted that $\vector{q} \not\in \vector{R}$.

\subsubsection{Hypervolume (HV)}

Since HV \cite{ZitzlerT98} is the only Pareto-compliant quality indicator known so far \cite{ZitzlerBT06}, HV is one of the most popular quality indicators.
The HV value of $\vector{A}$ is the volume of the area that is dominated by objective vectors in $\vector{A}$ and bounded by the reference point $\vector{q}$ as follows:
%
%
\begin{align}
\label{eqn:hv}
{\rm HV} (\vector{A}) = {\rm volume}\Biggl(\bigcup_{\vector{a} \in \vector{A}} [a_1, q_1] \times ... \times [a_m, q_m] \Biggr),
\end{align}
%
where the function ``${\rm volume}$'' in \eqref{eqn:hv} is the Lebesgue measure.
A large HV value indicates that $\vector{A}$ approximates the Pareto front well in terms of both convergence and diversity.


\subsubsection{Inverted generational distance (IGD)}
\label{subsubsec:igd}
IGD \cite{CoelloS04} evaluates the quality of $\vector{A}$ in terms of both convergence and diversity.
IGD measures the average distance from each reference vector $\vector{r}$ in $\vector{R}$ to its nearest objective vector $\vector{a}$ in $\vector{A}$ as follows:
%
\begin{align}
\label{eqn:igd}
{\rm IGD} (\vector{A}) = \frac{1}{|\vector{R}|} \left(\sum_{\vector{r} \in \vector{R}} \min_{\vector{a} \in \vector{A}} \Bigl\{ d(\vector{a}, \vector{r}) \Bigr\} \right),
\end{align}
where $d(\vector{a}, \vector{r})$ in \eqref{eqn:igd} is the distance between $\vector{a}$ and $\vector{r}$.
IGD uses the following Euclidean distance: $d_{\rm IGD}(\vector{a}, \vector{r}) = \sqrt{\sum^m_{i=1} (a_i - r_i)^2}$.
%
%

\subsubsection{IGD plus (IGD$^+$)}
\label{subsubsec:igdp}


Since IGD is not Pareto compliant, IGD may incorrectly evaluate the quality of $\vector{A}$ that contains non-converged objective vectors \cite{IshibuchiMTN15,SchutzeELC12}.
IGD$^+$ \cite{IshibuchiMTN15} addresses the issue in IGD.
While IGD is Pareto non-compliant, IGD$^+$ is weakly Pareto compliant.
The IGD$^+$ value of $\vector{A}$ is the average distance from each reference vector $\vector{r}$ to its nearest objective vector in the dominated region by $\vector{A}$ \cite{IshibuchiIMN19}.
Although the IGD$^+$ value of $\vector{A}$ is calculated using \eqref{eqn:igd}, $d$ is replaced by the following distance function: $d_{\rm IGD^+}(\vector{a}, \vector{r}) = \sqrt{\sum^m_{i=1} \bigl(\max\{a_i - r_i, 0\}\bigr)^2}$.
It should be noted that small IGD and IGD$^+$ values are preferable whereas large HV values are preferable.





%
In addition to IGD$^+$, the averaged Hausdorff distance indicator ($\Delta_p$) \cite{SchutzeELC12} addresses the issue in IGD.
When all elements in $\vector{A}$ are on the Pareto front and $\vector{R}$ has a large number of uniformly distributed reference vectors over the entire Pareto front (as in our computational experiments in this paper), the $\Delta_p$ value of $\vector{A}$ is equal to the IGD value of $\vector{A}$ in general.
Actually, the IGD and $\Delta_p$ values are the same for all of the obtained solution sets in this paper.
For this reason, we do not report experimental results for $\Delta_p$.
Of course, the above-mentioned relationship between $\Delta_p$ and IGD does not always hold depending on $\vector{A}$ and $\vector{R}$. 








\subsubsection{The additive $\epsilon$-indicator ($I_{\epsilon+}$)}
Here, we describe the unary version of $I_{\epsilon+}$ \cite{ZitzlerTLFF03}.
The $I_{\epsilon+}$ value of two objective vectors $\vector{a}$ and $\vector{r}$ is given as follows:
\begin{align}
\label{eqn:epsilon_point}
I_{\epsilon+}^{\rm vec} (\vector{a}, \vector{r}) = \max_{i \in \{1, ..., m\}} \{a_i - r_i\},
\end{align}
%
where the $I_{\epsilon+}^{\rm vec} (\vector{a}, \vector{r})$ value is the minimal shift such that $\vector{a}$ weakly dominates $\vector{r}$.
It should be noted that $I_{\epsilon+}^{\rm vec} (\vector{a}, \vector{r}) \neq I_{\epsilon+}^{\rm vec} (\vector{r}, \vector{a})$ in \eqref{eqn:epsilon_point}.
$I_{\epsilon+}$ evaluates how well $\vector{A}$ approximates the reference vector set $\vector{R}$ as follows:
\begin{align}
  \label{eqn:epsilon_set}
  I_{\epsilon+} (\vector{A}) = \max_{\vector{r} \in \vector{R}} \biggl\{ \min_{\vector{a} \in \vector{A}} \Bigl\{ I_{\epsilon+}^{\rm vec} (\vector{a}, \vector{r}) \Bigr\} \biggr\},
\end{align}
where the $I_{\epsilon+}$ value of $\vector{A}$ in \eqref{eqn:epsilon_set} is the minimal shift such that each reference objective vector in $\vector{R}$ is weakly dominated by at least one objective vector in $\vector{A}$.
A small $I_{\epsilon+}$ value indicates that the corresponding $\vector{A}$ is a good approximation of the Pareto front in terms of both convergence and diversity.






\subsubsection{R2}


Although three R indicators are proposed in \cite{HansenJ98}, the following R2 indicator is the most widely used one:
%
\begin{align}
  \label{eqn:r2}
  {\rm R2}(\vector{A}) &= \frac{1}{|\vector{W}|} \sum_{\vector{w} \in \vector{W}} \min_{\vector{a} \in \vector{A}} \Bigl\{g^{\rm tch}(\vector{a}, \vector{w}) \Bigr\},
\\
  \label{eqn:Tchebycheff}
  g^{\rm tch}(\vector{a}, \vector{w}) &= \max_{i \in \{1, ..., m\}} \bigl\{ w_i |a_i - z_i^*| \bigr\},
\end{align}
%
where $\vector{W}$ in \eqref{eqn:r2} is a set of weight vectors.
For each $\vector{w} = (w_1, ..., w_m)^{\rm T} \in \vector{W}$, $0 \leq w_i \leq 1$ for $i \in \{1, ..., m\}$ and $\sum^{m}_{i=1}w_i = 1$.
In \eqref{eqn:Tchebycheff}, $g^{\rm tch}: \mathbb{R}^m \rightarrow \mathbb{R}$  is the weighted Tchebycheff function, and $\vector{z}^* = (z^*_1, ..., z^*_m)^{\rm T}$ is the ideal point.
The $i$-th element $z^*_i$ of $\vector{z}^*$ is the minimum value of the $i$-th objective function over the Pareto front.
According to \cite{LiY19},  R2 with a tuned $\vector{z}^*$ works well on various problems with $m=2$. 
A small R2 value indicates that $\vector{A}$ approximates the Pareto front well in terms of both convergence and diversity.


%



\subsubsection{New R2 indicator (NR2)}



Although the Pareto compliance of HV is attractive, the computation time for the HV calculation increases exponentially with the number of objectives $m$.
To address this issue, the method of approximating the HV value using R2 is proposed in \cite{IshibuchiTSN10}.
The following NR2 \cite{ShangIZL18} is an improved version of R2 to obtain a better approximation of the HV value:
\begin{align}
\label{eqn:nr2}
  {\rm NR2}(\vector{A}) &= \frac{1}{|\vector{W}|} \sum_{\vector{w} \in \vector{W}} \min_{\vector{a} \in \vector{A}} \Bigl\{g^{\rm mtch}(\vector{a}, \vector{w}) \Bigr\}^m,\\
\label{eqn:mTchebycheff}
g^{\rm mtch}(\vector{a}, \vector{w}) &= \max_{i \in \{1, ..., m\}} \biggl\{ \frac{|q_i - a_i|}{w_i}  \biggr\},
\end{align} 
where $g^{\rm mtch}$ in \eqref{eqn:mTchebycheff} is a modified version of $g^{\rm tch}$.
If $w_i = 0$ in \eqref{eqn:mTchebycheff}, $w_i$ was set to $10^{-6}$ to avoid division by zero.
Similar to HV, NR2 uses the reference point $\vector{q}$.
A large NR2 value indicates that the corresponding $\vector{A}$ has a good HV value.
Since NR2 was designed to approximate the HV value, it is expected that their optimal $\mu$-distributions are similar to each other.
However, such an expectation has not been verified in the literature.
To address this issue, we examine the approximated optimal $\mu$-distribution for NR2 in this paper.





\subsubsection{$s$-energy (SE)}

Although SE \cite{HardinS04} was not originally proposed in the field of EMO, SE is used as a quality indicator in some studies (e.g., \cite{Falcon-CardonaC18}).
SE is given as follows:
\begin{align}
 \label{eqn:senergy}
 {\rm SE} (\vector{A}) = \sum_{\vector{a} \in \vector{A}} \sum_{\vector{b} \in \vector{A} \setminus \{\vector{a}\}} \|\vector{a}  - \vector{b}  \|_2^{-s},
 \end{align}
where $s$ is usually set to $m-1$, and $\|\vector{a}\|_p$ is the $L_{p}$ norm of $\vector{a}$.
If $\vector{a} = \vector{b}$ in \eqref{eqn:senergy}, we set the SE value of $\vector{A}$ to an infinitely large value to avoid division by zero.
A small SE value indicates that $\vector{A}$ has a good uniformity and a good spread.





\subsubsection{Generalized Spread ($\Delta$)}


The original $\Delta$ indicator \cite{DebAPM02} can handle only the two-dimensional objective space.
In this paper, we use an extension of $\Delta$ with respect to $m$ presented in \cite{WangWY10}.
$\Delta$ evaluates the quality of $\vector{A}$ in terms of both uniformity and spread as follows:
%
\begin{align}
  \label{eqn:delta}
        \Delta (\vector{A}) &= \frac{d^{\rm ext} + \sum_{\vector{a} \in \vector{A}} |{\rm dist}(\vector{a}, \vector{A}) - d^{\rm avg}|}{d^{\rm ext} + d^{\rm avg}(|\vector{A}|-m)},\\
\label{eqn:delta_dist}
{\rm dist}(\vector{u}, \vector{V}) &= \min_{\vector{v} \in \vector{V}, \vector{v} \neq \vector{u}} \|\vector{u} - \vector{v} \|_2,\\
\label{eqn:delta_ext}
d^{\rm ext} &= \sum^m_{i=1} {\rm dist}(\vector{r}^{\rm ext}_i, \vector{A}),\\
 \label{eqn:delta_avg}
  d^{\rm avg} &= \frac{1}{|\vector{A}|} \sum_{\vector{a} \in \vector{A}} {\rm dist}(\vector{a}, \vector{A}),  
\end{align}
%
where the function ${\rm dist}(\vector{u}, \vector{V})$ in \eqref{eqn:delta_dist} returns the Euclidean distance from $\vector{u}$ to its nearest vector in the set $\vector{V}$. 
In \eqref{eqn:delta_ext}, $\vector{r}^{\rm ext}_i$ is the $i$-th extreme objective vector with the maximum value of the $i$-th objective function in $\vector{R}$ ($i \in \{1, ..., m\}$).
In \eqref{eqn:delta}, $d^{\rm ext}$ aims to evaluate the spread of $\vector{A}$, while the other parts aim to evaluate the uniformity of $\vector{A}$.
A small $\Delta$ value indicates that $\vector{A}$ has good spread and uniformity.

\subsubsection{Pure diversity (PD)}

PD \cite{WangJY17} is an extended version of the Solow-Polasky diversity indicator \cite{SolowP94} for multi-objective optimization.
The PD value of $\vector{A}$ is based on the dissimilarity of each objective vector $\vector{a}$ to $\vector{A}$ as follows:
\begin{align}
 \label{eqn:pd}
  {\rm PD} (\vector{A}) &= \max_{\vector{a} \in \vector{A}} \Bigl\{{\rm PD} \bigl(\vector{A}\setminus\{\vector{a}\}\bigr) + d\bigl(\vector{a}, \vector{A}\setminus\{\vector{a}\}\bigr)\Bigr\},\\
  \label{eqn:pd_dist}
   d (\vector{u}, \vector{V}) &= \min_{\vector{v} \in \vector{V}} \bigl\{\|\vector{u} - \vector{v} \|_p \bigr\},
 \end{align}
where the recommended value of $p$ in the $L_{p}$ norm in \eqref{eqn:pd_dist} is $0.1$.
Most diversity-based quality indicators (including SE and $\Delta$) evaluate the uniformity of $\vector{A}$.
In contrast, PD evaluates the dissimilarity between objective vectors in $\vector{A}$.



\section{Related work}
\label{sec:related_work}




This section describes related work.
Analysis of quality indicators can be categorized into two approaches.
One is ranking information based approaches.
First, multiple objective vector sets are ranked by each quality indicator.
Then, the properties of each quality indicator are discussed based on which objective vector set is highly evaluated by the quality indicator.
The consistency of multiple quality indicators can also be investigated in this manner.
One main drawback of the ranking information based approaches is that only relative information can be obtained.

The other is optimal $\mu$-distribution based approaches.
As explained in Section \ref{sec:introduction}, the optimal $\mu$-distribution provides insightful information about the properties of each quality indicator.
One main drawback of the optimal $\mu$-distribution based approaches is that it is difficult to find the optimal $\mu$-distributions for quality indicators.


%


\subsection{Ranking information based approaches}
\label{sec:related_work_rank}




Okabe et al. \cite{OkabeJS03} analyzed 15 quality indicators using objective vector sets artificially generated on the linear Pareto front and objective vector sets found by EMOAs on the convex Pareto front.
The results show that some quality indicators can generate misleading results.
%
Jiang et al. \cite{JiangOZF14} examined six quality indicators using a pair of objective vector sets on the linear, convex, and concave Pareto fronts.
The results show that $I_{\epsilon+}$, IGD, and HV are consistent with each other on the convex Pareto front, but IGD is inconsistent with HV on the concave Pareto front.
Ravber et al. \cite{RavberMC17} investigated 11 quality indicators using a chess rating system.
They used objective vector sets found by EMOAs on some test problems.
The results show the robustness of HV, $I_{\epsilon+}$, and IGD$^+$.
However, the results also show that the three quality indicators do not equally evaluate the quality of objective vector sets in terms of the convergence, the uniformity, and the spread.

Wessing and Naujoks \cite{WessingN10} analyzed the correlation between HV, R2, and $I_{\epsilon+}$.
Their analysis was based on objective vector sets randomly generated on two- and three-objective problems.
The results based on Pearson's correlation coefficient show that HV is highly consistent with R2.
Liefooghe and Derbel \cite{LiefoogheD16} examined the correlation between six quality indicators.
They used objective vector sets found by the random search and NSGA-II \cite{DebAPM02} on test problems with $m \in \{2, 3\}$.
The results based on the Kendall rank correlation show that HV is consistent with R2 and R3.
In contrast, $I_{\epsilon+}$ and its multiplicative version are inconsistent with other quality indicators.
Ishibuchi et al. \cite{IshibuchiMN15gecco} compared IGD$^+$ with IGD on various objective vector sets found by NSGA-II on problems with up to 10 objectives.
The results show that IGD$^+$ can evaluate objective vector sets more accurately than IGD in terms of the Pareto compliance.
In addition to IGD$^+$ and IGD, three quality indicators (GD, $I_{\epsilon+}$, and D1 \cite{HansenJ98}) were examined.
Bezerra et al. \cite{BezerraLS17} analyzed the relation between IGD$^+$ and IGD on the linear Pareto front.
Various objective vector sets were used in their analysis.
The results show that IGD and IGD$^+$ are consistent with each other in most cases.
The results also show that IGD may favor objective vectors with a poor spread.





\subsection{Optimal $\mu$-distribution based approaches}
\label{sec:related_work_astar}



It is difficult to obtain the optimal $\mu$-distributions in an exact manner.
Thus, the optimal $\mu$-distributions are generally approximated by analytical or empirical methods.
For the differences between the two methods, see Section \ref{sec:introduction}.

Auger et al. \cite{AugerBBZ09} formulated finding the optimal $\mu$-distribution for HV on two-objective problems as a single-objective continuous optimization problem.
Below, we describe its problem formulation in a general manner.
Our description significantly differs from the original one.


Let $\vector{A} = \{\vector{a}_1, ...,\vector{a}_{\mu}\}$ be a set of $\mu$ objective vectors on the Pareto front.
Also, let $\theta_i$ be a real value that determines the first element (i.e., the value of the first objective $f_1$) of the $i$-th objective vector $\vector{a}_i$ ($i \in \{1, ..., \mu\}$) in $\vector{A}$.
All objective vectors in the optimal $\mu$-distributions must always be on the Pareto front.
For this restriction, the position of $\vector{a}_i$ for $f_1$ specifies that for the second objective $f_2$.
In \cite{AugerBBZ09}, a front shape function $F: \mathbb{R} \rightarrow \mathbb{R}$ is used to determine the position of $\vector{a}_i$ for $f_2$ as follows: $\vector{a}_i =\bigl(\theta_i, F(\theta_i)\bigr)^{\rm T}$.
For all $i \in \{1, ..., \mu\}$, $\theta_i$ is in the range $[\theta^{\rm min}, \theta^{\rm max}]$, where $\theta^{\rm min}$ and $\theta^{\rm max}$ are the lower and upper bounds for $\theta_i$.
Here, $\theta^{\rm min}$ and $\theta^{\rm max}$ correspond to the minimum and maximum values of $f_1$ in the Pareto front, respectively.
In this manner, the distribution of $\mu$ objective vectors in $\vector{A}$ is determined by a vector $\vector{\theta} = (\theta_1, ..., \theta_{\mu})^{\rm T}$.



\begin{figure}[t]
\newcommand{\widthvar}{0.15}
\centering
\subfloat[$F_{\rm DTLZ1}$]{\includegraphics[width=\widthvar\textwidth]{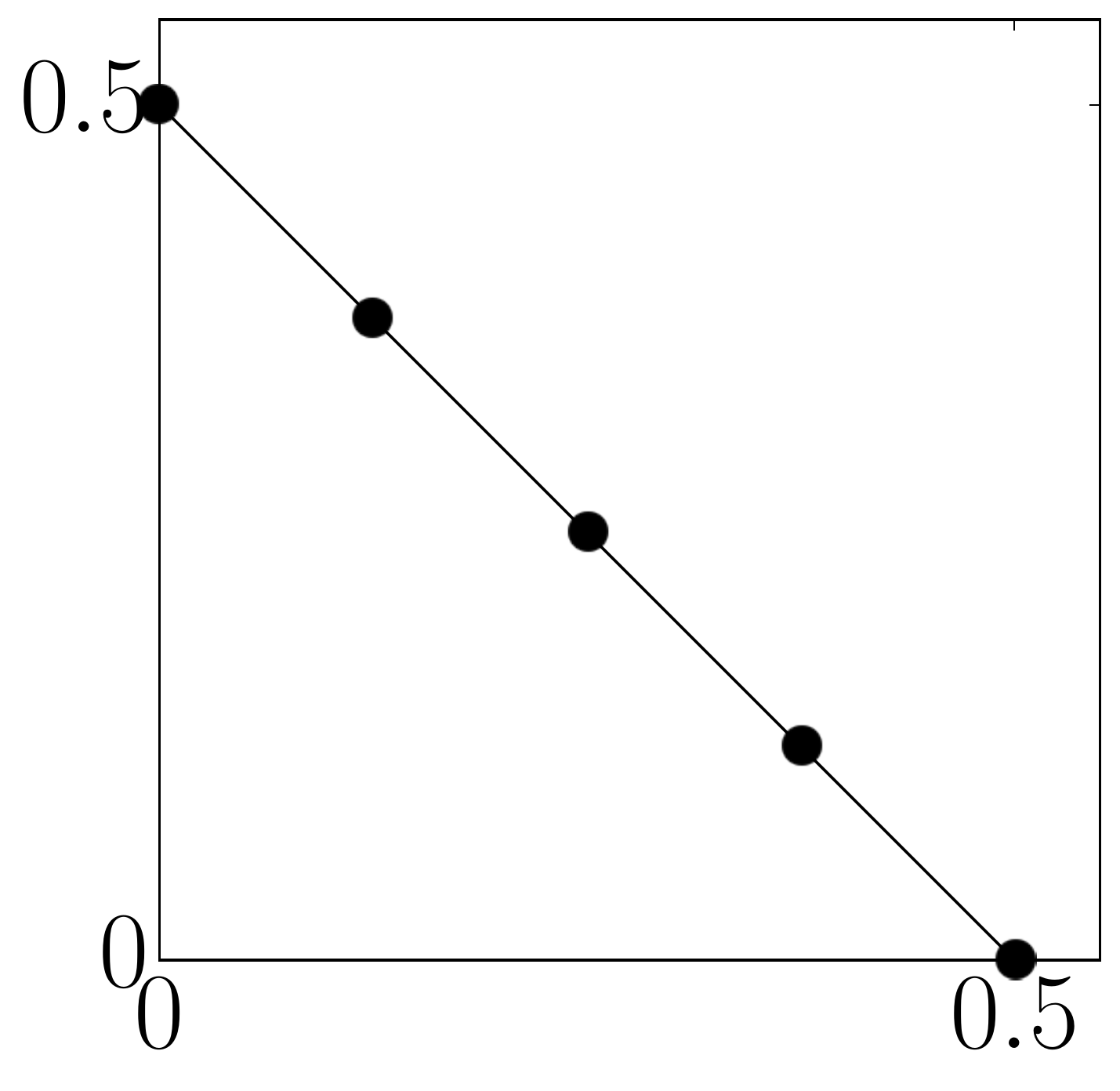}}
\subfloat[$F_{\rm DTLZ2}$]{\includegraphics[width=\widthvar\textwidth]{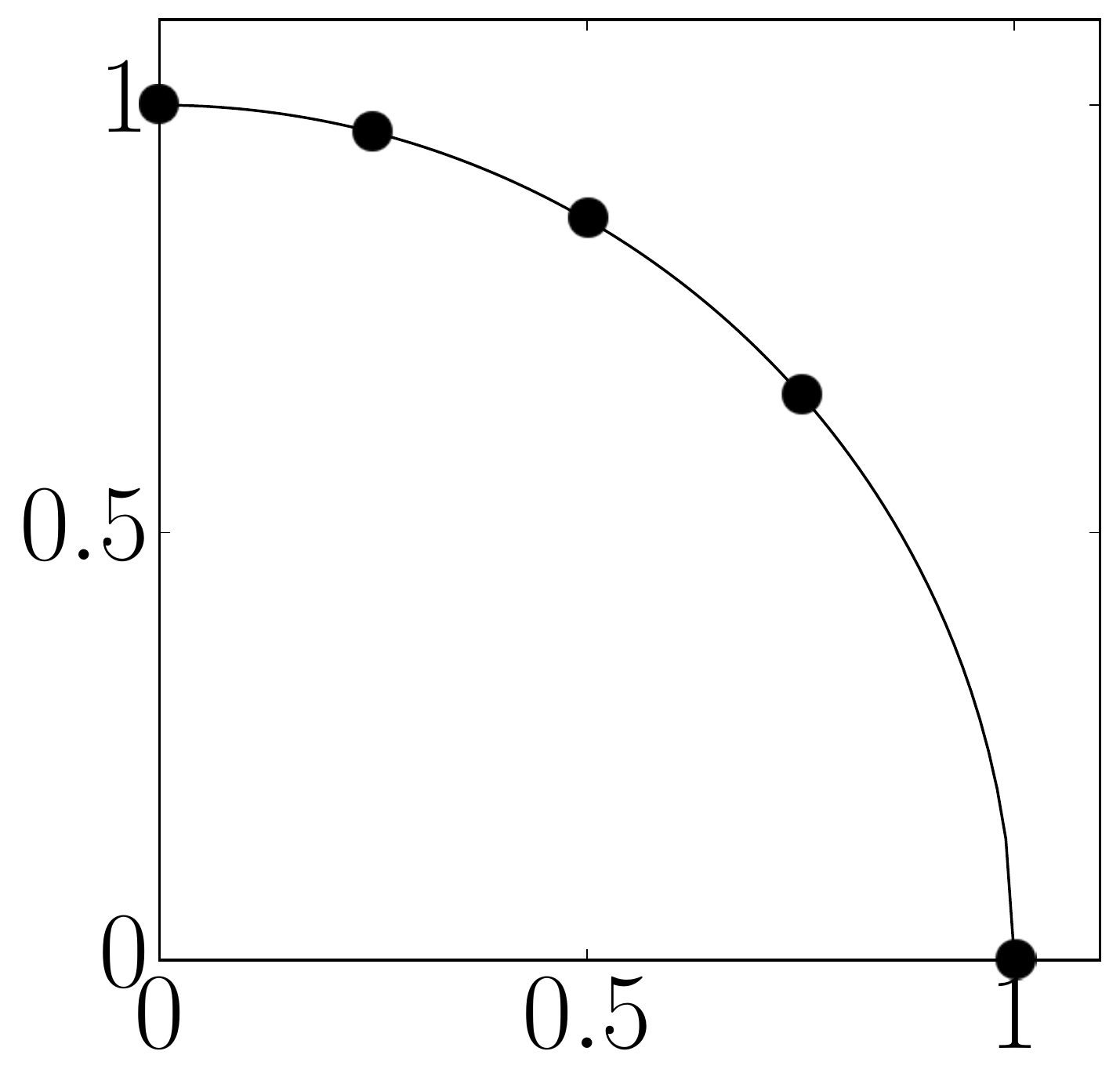}}
\subfloat[$F_{\rm ZDT1}$]{\includegraphics[width=\widthvar\textwidth]{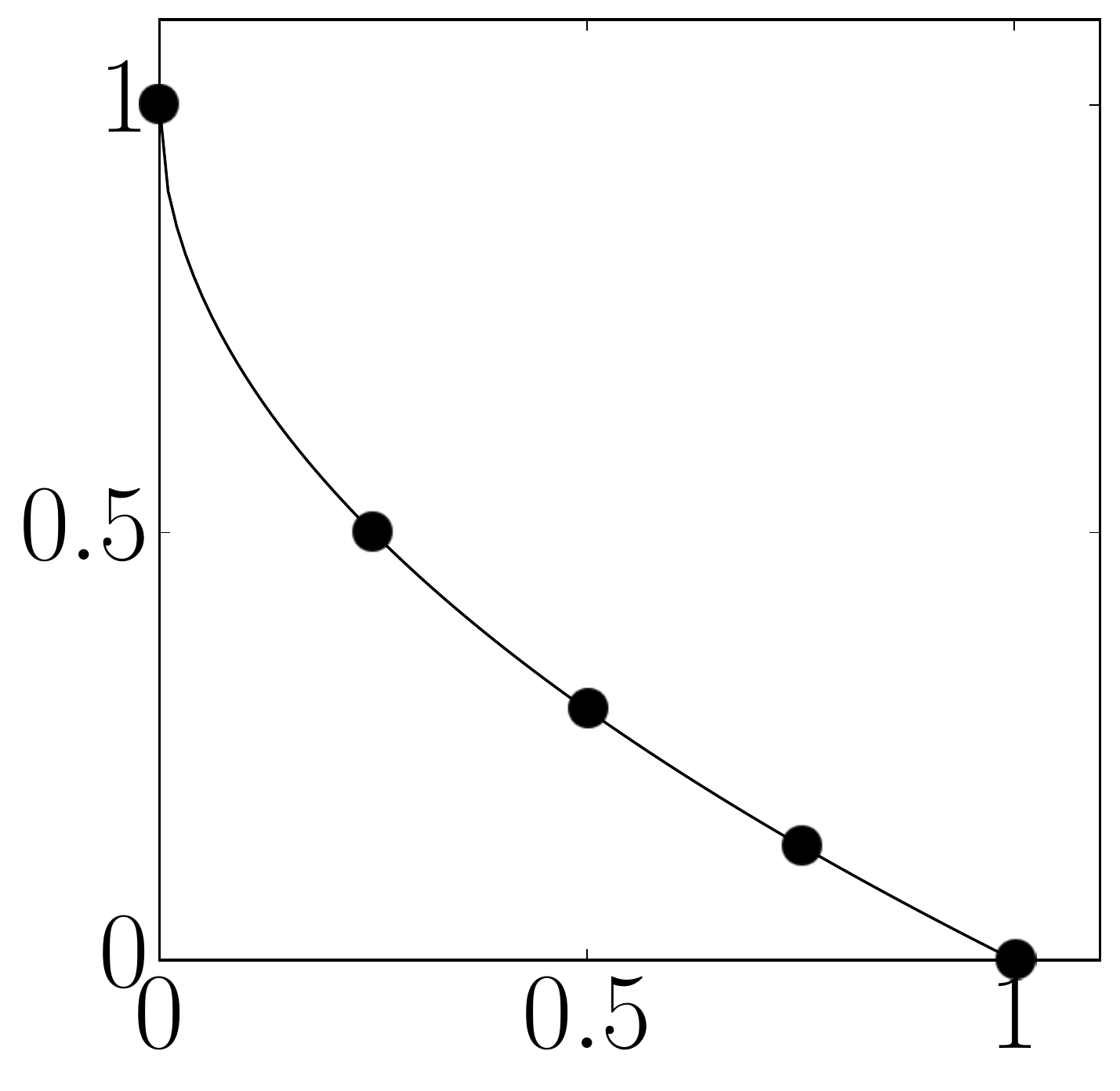}}
\caption{
\small
Distributions of five translated objective vectors. 
The $x$ and $y$ axes represent $f_1$ ($\theta$) and $f_2$ ($F(\theta)$), respectively.
}
\label{fig:example_translation}
\end{figure}

%
Fig. \ref{fig:example_translation} shows examples of five objective vectors.
In Fig. \ref{fig:example_translation}, the Pareto front is shown by 
\begin{align}
  \label{eqn:dtlz1}
  F_{\rm DTLZ1} (\theta) &= 0.5 - \theta &(\theta \in [0, 0.5]),\\
  \label{eqn:dtlz2}
  F_{\rm DTLZ2} (\theta) &= \sqrt{1 - \theta^2} &(\theta \in [0, 1]),\\
  \label{eqn:zdt1}
  F_{\rm ZDT1} (\theta) &= 1 - \sqrt{\theta} &(\theta \in [0, 1]),
\end{align}
where $F_{\rm DTLZ1}$, $F_{\rm DTLZ2}$, and $F_{\rm ZDT1}$ represent the Pareto front shapes of DTLZ1 \cite{DebTLZ05}, DTLZ2 \cite{DebTLZ05}, and ZDT1 \cite{ZitzlerDT00}, respectively.
In Fig. \ref{fig:example_translation}, $\mu=5$, $\vector{\theta} = (0, 0.125, 0.25, 0.375, 0.5)^{\rm T}$ for $F_{\rm DTLZ1}$, and $\vector{\theta} = (0, 0.25, 0.5, 0.75, 1)^{\rm T}$ for $F_{\rm DTLZ2}$ and $F_{\rm ZDT1}$.
%
As shown in Fig. \ref{fig:example_translation}, distributions of objective vectors on various Pareto fronts can be examined by changing $F$.


In summary, finding $\vector{A}$ that minimizes a quality indicator $I$ is the same as finding $\vector{\theta}$ that indirectly minimizes $I$ as follows:
\begin{align}
  \label{eqn:augers_m2_def_objvec}
    \text{minimize  }  I(\vector{A}) = I \biggl(\Bigl\{ \bigl(\theta_i, F(\theta_i)\bigr)^{\rm T}  \, | \, i \in \{1, ..., \mu \}\Bigr\}\biggr),
\end{align}
where $\vector{\theta}$ is a solution of the problem defined in \eqref{eqn:augers_m2_def_objvec}.
$I$ that needs to be maximized (e.g., HV, NR2, and PD) can be translated as $-I$ without loss of generality.
Any derivative-free optimizer can be used to minimize $I$ in \eqref{eqn:augers_m2_def_objvec}.

Auger et al. \cite{AugerBBZ09} proposed analytical and empirical methods of approximating optimal $\mu$-distributions for HV.
Details of the analytical method for each $F$ can be found in the supplementary website of \cite{AugerBBZ09} (\url{https://sop.tik.ee.ethz.ch/download/supplementary/testproblems/}).
The proposed problem-specific local search algorithm in \cite{AugerBBZ09} for \eqref{eqn:augers_m2_def_objvec} uses variable-wise perturbation.
For each iteration, each $\theta_i$ is perturbed by a value selected randomly from a normal distribution.
%
Brockhoff et al. \cite{BrockhoffWT12} applied CMA-ES \cite{HansenO01} to \eqref{eqn:augers_m2_def_objvec} to find the optimal $\mu$-distributions for R2.
Their results show that objective vectors in the optimal $\mu$-distributions for R2 are densely distributed in the center of the Pareto front.
They also examined the contribution of each objective vector in the optimal $\mu$-distribution for R2 to HV (and that for HV to R2).



In contrast to the above-mentioned studies \cite{AugerBBZ09,BrockhoffWT12}, the following studies do not use \eqref{eqn:augers_m2_def_objvec}.
Bringmann et al. \cite{BringmannFK15} proposed the analytical method for finding the optimal $\mu$-distributions for $I_{\epsilon+}$ on the Pareto front with $m=2$.
The proposed method explicitly exploits the properties of $I_{\epsilon+}$ presented in \cite{BringmannFK14gecco}.
The results show that the optimal $\mu$-distributions for $I_{\epsilon+}$ on most Pareto fronts with $m=2$ are uniform.
Glasmachers \cite{Glasmachers14} proposed a gradient-based analytical method for approximating the optimal $\mu$-distributions for HV on $m \in \{2, 3\}$.
For $m=3$, the proposed method in \cite{Glasmachers14} decomposes the three-dimensional objective space into cuboids.
Then, the proposed method in \cite{Glasmachers14} uses the gradient of HV based on the cuboids.


Ishibuchi et al. \cite{IshibuchiISN18ecj} used SMS-EMOA \cite{BeumeNE07} to approximate the optimal $\mu$-distributions for HV on Pareto fronts of three-objective problems.
SMS-EMOA uses the HV contribution in the steady-state selection.
For each iteration, the worst individual in terms of the HV contribution is removed from the last non-domination level.
In \cite{IshibuchiISN18ecj}, SMS-EMOA was applied to the DTLZ and inverted-DTLZ problems whose number of distance variables is zero.
Then, the influence of the reference vector on the optimal $\mu$-distributions for HV is examined.
Ishibuchi et al. \cite{IshibuchiISN18} also investigated the optimal $\mu$-distributions for IGD approximated by a similar approach.
The HV contribution-based selection in SMS-EMOA was replaced by the IGD contribution-based selection.
%
The optimal $\mu$-distributions for HV, IGD, and IGD$^+$ approximated by SMS-EMOA in a similar manner were analyzed in \cite{IshibuchiIMN19}.
The results show that the optimal $\mu$-distributions for IGD are almost uniform in all problems.
The results also show that the optimal $\mu$-distributions for HV and IGD$^+$ are similar in all problems when the reference vector for HV is set appropriately.

\section{Proposed formulation}
\label{sec:proposed_method}


Here, we introduce the proposed problem formulation of finding the optimal $\mu$-distribution for quality indicators on Pareto fronts for arbitrary number of objectives ($m \geq 2$).
As reviewed in Subsection \ref{sec:related_work_astar}, some methods of approximating the optimal $\mu$-distributions have been proposed in the literature.
The analytical methods \cite{AugerBBZ09,Glasmachers14,BringmannFK15} can efficiently approximate the optimal $\mu$-distributions in a computationally cheap manner.
Unfortunately, they can be applied to only quality indicators whose theoretical properties are clear (i.e., HV and $I_{\epsilon+}$).
It is questionable whether the empirical methods using SMS-EMOA \cite{IshibuchiISN18,IshibuchiISN18ecj,IshibuchiIMN19} can be applied to quality indicators except for HV, IGD, and IGD$^+$.
The empirical methods based on the problem formulation in \eqref{eqn:augers_m2_def_objvec} are promising.
Although the previous studies \cite{AugerBBZ09,BrockhoffWT12} addressed only HV and R2, the approaches using \eqref{eqn:augers_m2_def_objvec} can be applied to any quality indicators in principle.
However, the problem formulation in \eqref{eqn:augers_m2_def_objvec} can handle only Pareto fronts for $m=2$.
In contrast, the proposed problem formulation can handle Pareto fronts for $m \geq 2$.
We do not claim that the degenerated version of our formulation for $m=2$ is better than the formulation in \eqref{eqn:augers_m2_def_objvec}.
Note that  our formulation for $m=2$ is almost the same as the formulation in \eqref{eqn:augers_m2_def_objvec}.


The formulation proposed in \cite{AugerBBZ09} for $m=2$ and our formulation for $m\geq 2$ are essentially one of the set-based optimization methods \cite{ZitzlerTB10}.
Test problems constructed by bottom-up approaches (e.g., DTLZ) have two types of decision variables \cite{MartinezCAT19}.
One is position-related variables that specify the position of an objective vector on the Pareto front.
The other is distance-related variables that determine the distance between the objective vector and the Pareto front.
Thus, any solutions with only position-related variables are always on the Pareto front in the objective space.
The optimal $\mu$-distribution for a quality indicator can be approximated by a set-based method that finds a set of $\mu$ {\em solutions} (not objective vectors) with only position-related variables for minimization of the quality indicator.
While we here denote this approach as the {\em solution} set-based approach, we denote the formulations proposed in \cite{AugerBBZ09} and this paper as the {\em objective vector} set-based approach.
The main advantage of the objective vector set-based approach is its simplicity.
While the solution set-based approach needs to handle both solution and objective spaces, the objective vector set-based approach just needs to handle only the objective space.
We can focus only on the objective space in the objective vector set-based approach.
This simplicity is beneficial for an analysis of the optimal $\mu$-distributions as demonstrated in \cite{AugerBBZ09}.

%





\subsection{Proposed problem formulation}
\label{sec:proposed_method_sub}



Below, we explain the proposed formulation for general cases.
Only a disconnected front shape function $F_{\rm disconnected}$ explained in Subsection \ref{sec:self_T_function} requires an alternative formulation.
For details, see Subsection S.1-D in the supplementary file.
Finding the optimal $\mu$-distribution for a quality indicator $I$ on the Pareto front with $m$ objectives is formulated as follows:
\begin{align}
  \label{eqn:proposed_m_def}
  \text{minimize  } \:\, &I (\vector{A}),\\
  \text{subject to  } \, &G(\vector{A}) \leq 0,\notag
\end{align}
%
where $G$ is a constraint function for constrained front shape functions (e.g., $F_{\rm c\shyp concave}$ explained in Subsection \ref{sec:self_T_function}).
In contrast to \eqref{eqn:augers_m2_def_objvec}, $G$ is added to \eqref{eqn:proposed_m_def}.
If a given front shape function has no constraint, $G$ is eliminated from \eqref{eqn:proposed_m_def}.
An objective vector set $\vector{A} = \{\vector{a}_1, ...,\vector{a}_{\mu}\}$ is a translated version of a solution $\vector{\theta}$ of this problem in a two-phase manner as follows:
\begin{align}
  \label{eqn:proposed_m_def_objvec}
  \vector{A} &= F (\vector{B}),\\
  \label{eqn:proposed_translation}  
  \vector{B} &= {\rm translate} (\vector{\theta}),
\end{align}
where $\vector{\theta} = (\theta_1, ..., \theta_d)^{\rm T}$ is a $d$-dimensional vector, and $d=\mu (m-1)$.
Any single-objective black-box optimizer can be used to find $\vector{\theta}$ that minimizes $I$.
While the size of $\vector{\theta}$ is $\mu$ in \eqref{eqn:augers_m2_def_objvec}, that is $d$ in \eqref{eqn:proposed_m_def_objvec}.
%
For all $i \in \{1, ..., d\}$, $\theta_i$ is in $[0, 1]$.


The function ``${\rm translate}$'' in \eqref{eqn:proposed_translation} translates a $d$-dimensional vector $\vector{\theta}$ into a set $\vector{B} = \{\vector{b}_1, ..., \vector{b}_{\mu}\}$ that consists of $\mu$ $m$-dimensional vectors.
For each $i \in \{1, ..., \mu\}$, $\vector{b}_{i} = (b_{i,1}, ..., b_{i,m})^{\rm T}$, and $\sum^m_{j=1} b_{i,j} = 1$.
Although $\vector{b}_{i}$ is an $m$-dimensional vector, it is not an objective vector.
Subsection \ref{sec:self_S_function} explains the translation operator in detail.

After translating $\vector{\theta}$ into $\vector{B}$ in \eqref{eqn:proposed_translation}, $\vector{B}$ is further mapped to the objective vector set $\vector{A}$ by a front shape function $F$ in \eqref{eqn:proposed_m_def_objvec}.
For each $i \in \{1, ..., \mu\}$, an element $\vector{b}_i$ in $\vector{B}$ corresponds to an objective vector $\vector{a}_i$ in $\vector{A}$.
All objective values are normalized into $[0,1]$ in our study by using the ideal point $\vector{z}^*$ and the nadir point $\vector{z}^{\rm nadir} = (z^{\rm nadir}_1, ..., z^{\rm nadir}_m)^{\rm T}$ of each $F$ as follows: $a_i := (a_i - z^*_i)/(z^{\rm nadir}_i - z^*_i)$ for each $i \in \{1, ..., m\}$.
Here, $z^{\rm nadir}_i$ is the maximum value of the $i$-th objective function of the Pareto front.
We use the normalization in order to examine the approximated optimal $\mu$-distributions for the quality indicators without worrying about the influence of differently-scaled objective values.
Subsection \ref{sec:self_T_function} specifies how to map $\vector{B}$ to $\vector{A}$ by the front shape function $F$.



\subsection{Translation operator}
\label{sec:self_S_function}







We explain how the translation operator in \eqref{eqn:proposed_translation} maps $\vector{\theta}$ to $\vector{B}$.
First, all $d$ elements in $\vector{\theta}$ are divided into $\mu$ vector groups as $(\theta_1, ..., \theta_{1+m-2})^{\rm T}$, ..., $(\theta_{d-m+2}, ..., \theta_{d})^{\rm T}$.
 For example, when $\mu=2$, $m=3$, $d=\mu (m-1) = 4$, and $\vector{\theta} = (0.1, 0.2, 0.3, 0.4)^{\rm T}$, 4 elements in $\vector{\theta}$ are divided as follows: $(0.1, 0.2)^{\rm T}$ and $(0.3, 0.4)^{\rm T}$.
For the sake of simplicity, we denote $(\theta_i, ..., \theta_{i+m-2})^{\rm T}$ as $\vector{y}= (y_1, ..., y_{m-1})^{\rm T}$ for each $i \in \{1, 1 (m-1), 2 (m-1), ...,  (d-m+2)\}$.

Below, we describe how the $(m-1)$-dimensional vector $\vector{y}$ is translated into an $m$-dimensional vector $\vector{b}$ in $\vector{B}$.
To obtain all $\mu$ vectors in $\vector{B}$, the same operation is repeatedly applied to $\vector{y}$ for each $i \in \{1, 1 (m-1), 2 (m-1), ...,  (d-m+2)\}$.
Each element in $\vector{b}$ is obtained using the Jaszkiewicz's method of generating random weight vectors \cite{Jaszkiewicz02} as follows:
%
\begin{align}
\label{eqn:uniform}
b_1 &= 1 - \sqrt[m-1]{y_1},\\
b_j &= \Biggl(1 - \sum^{j-1}_{k=1}b_{k}\Biggr)  \bigl(1 - \sqrt[m-1-j]{y_j}\bigr), \notag\\
b_m &= 1 - \sum^{m-1}_{k=1} b_k,\notag
\end{align}
%
where $j \in \{2, ..., m-1\}$ in \eqref{eqn:uniform}.
It should be noted that $\vector{b}$ obtained by \eqref{eqn:uniform} satisfies the condition $\sum^m_{k=1} b_{k} = 1$.
Although the Jaszkiewicz's method was originally proposed for the weight vector generation in decomposition-based EMOAs, it can be used to translate $\vector{y}$ into $\vector{b}$ in a straightforward manner.

\subsection{Front shape function $F$}
\label{sec:self_T_function}



The proposed formulation can be applied to any Pareto front when its front shape function $F$ is available.
In other words, the proposed formulation cannot be applied to the Pareto front whose analytical expression (i.e., its front shape function) is unavailable such as WFG3 \cite{IshibuchiMN16}.
Although various front shape functions were presented in the previous study \cite{AugerBBZ09}, they are only for $m=2$.
We design front shape functions for $m\geq3$ based on a method of generating uniformly distributed reference vectors in representative test problems (e.g., DTLZ and WFG) presented in \cite{TianXZCJ18}.

We use eight front shape functions shown in Table \ref{tab:front_shape_functions}.
The name of the original problems and the shape of the Pareto fronts are also shown in Table \ref{tab:front_shape_functions}.
$F_{\rm linear}$, $F_{\rm concave}$, and $F_{\rm convex}$ are the most basic front shape functions.
$F_{\rm i\shyp linear}$, $F_{\rm i\shyp concave}$, and $F_{\rm i\shyp convex}$ are inverted versions of  $F_{\rm linear}$, $F_{\rm convex}$, and $F_{\rm concave}$, respectively.
It should be noted that an inverted version of the convex Pareto front is concave, and vice versa.
$F_{\rm disconnected}$ has the disconnected and mixed Pareto front.
Some parts of the Pareto front of $F_{\rm c\shyp concave}$ are infeasible due to the constraint.
We select $F_{\rm c\shyp concave}$ to demonstrate that the proposed formulation can handle the constrained Pareto front.
%
In our preliminary experiments, we used the degenerate front shape function $F_{\rm degenerate}$ whose original problem is DTLZ5.
$F_{\rm degenerate}$ is not a surface but a curve since it is degenerate.
Because the shape of $F_{\rm degenerate}$ is concave, the optimal $\mu$-distribution on $F_{\rm degenerate}$ is similar to that for a two-objective problem with $F_{\rm concave}$.
For this reason, we omit $F_{\rm degenerate}$.




\begin{table}[t]
\begin{center}
  \caption{\small Properties of the eight front shape functions.}
{\footnotesize
  \label{tab:front_shape_functions}
\scalebox{0.98}[1]{ 
\begin{tabular}{lll}
\toprule
$F$ & Original problem & Shape\\
  \midrule
$F_{\rm linear}$ & DTLZ1 \cite{DebTLZ05} & Linear\\ 
$F_{\rm concave}$ & DTLZ2 \cite{DebTLZ05} & Concave\\
  $F_{\rm convex}$ & Convex DTLZ2 \cite{DebJ14} & Convex\\
    \midrule
$F_{\rm i\shyp linear}$ & Inverted DTLZ1  \cite{JainD14} & Inverted, linear\\
$F_{\rm i\shyp concave}$ & Inverted convex DTLZ2 \cite{IshibuchiSMN16} & Inverted, concave\\
$F_{\rm i\shyp convex}$ & Inverted DTLZ2 \cite{IshibuchiSMN16} & Inverted, convex\\
  \midrule
$F_{\rm disconnected}$ & DTLZ7 \cite{DebTLZ05} & Disconnected, mixed\\
$F_{\rm c\shyp concave}$ & C2-DTLZ2 \cite{JainD14} & Disconnected, concave\\
\toprule
\end{tabular}
}
}
\end{center}
\end{table}




Due to the paper length limitation, we explain only $F_{\rm concave}$ in this paper.
We describe other front shape functions in Section S.1 in the supplementary file.
In $F_{\rm concave}$, the $i$-th element $b_i$ of $\vector{b}$ in $\vector{B}$ is translated into the $i$-th element $a_i$ of $\vector{a}$ in $\vector{A}$ on the concave Pareto front as follows:
\begin{align}
\label{eqn:t_nonconvex}
a_i = \frac{b_i}{t},
\end{align}
where $t = \sqrt{\sum^m_{j=1} b_j^2}$.
This translation is based on the method of generating reference vectors in DTLZ2 described in \cite{TianXZCJ18}.
All objective vectors obtained by \eqref{eqn:t_nonconvex} satisfy the following relation of the Pareto front of DTLZ2: $\sum^{m}_{i=1} (f_i(\vector{x}^*))^2 = 1$.
In addition to the eight front shape functions in Table \ref{tab:front_shape_functions}, we believe that it is possible to design other front shape functions using \cite{TianXZCJ18} as reference.

\section{Experimental settings}
\label{sec:experimental_settings}


%


This section describes experimental settings.
We examine the properties of the nine quality indicators using the eight front shape functions in Table \ref{tab:front_shape_functions}.
Although the number of objectives $m$ can be arbitrarily specified in the proposed formulation, $m$ is set to three in order to visually discuss the distributions of objective vectors.
We investigate the optimal $\mu$-distributions with $\mu =10, 15, 21, 28, 36, 45$ ($d=20, 30, 42, 56, 72, 90$, respectively).
We select these $\mu$ values so that we can visually discuss the distribution of objective vectors.
The distribution of objective vectors with a too large $\mu$ value is unclear.
We can also examine the influence of $\mu$ on the optimal $\mu$-distributions by using the various $\mu$ values.





In the proposed formulation in \eqref{eqn:proposed_m_def}, a black-box optimizer is necessary to find $\vector{\theta}$ that minimizes a given quality indicator.
We use L-SHADE \cite{TanabeF14CEC}, which is an improved version of differential evolution (DE) \cite{StornP97}.
L-SHADE was the winner of the IEEE CEC2014 competition on single-objective real-parameter optimization.
We used the Java source code of L-SHADE provided by the authors of \cite{TanabeF14CEC}.
The default parameter setting was used for L-SHADE.
The maximum number of function evaluations was set to $10^4 \times d$, and 31 independent runs were performed.
For the constrained front shape function $F_{\rm c\shyp concave}$, we use the death-penalty constraint handling method.
This is because objective vector sets are just classified into feasible and infeasible groups as described in Subsection S.1-E in the supplementary file.
The fitness value of an infeasible individual is an infinitely large value so that all infeasible individuals cannot survive to the next iteration.
We compared L-SHADE to the analytical approach that approximates the optimal $\mu$-distributions for HV for $m=2$ \cite{AugerBBZ09} (see Subsection \ref{sec:related_work_astar}).
We confirmed that L-SHADE can find the approximated optimal $\mu$-distribution for HV with acceptable quality for $m=2$.
For details, see Section S.3 in the supplementary file.





%


We implemented the nine quality indicators ourselves, except for HV.
The WFG algorithm \cite{WhileBB12} was used for the HV calculation.
We used the source code of the WFG algorithm provided by the authors of \cite{WhileBB12}.
For PD, we carefully translated the Matlab source code provided by the authors of \cite{WangJY17} into our implementation in order to speed up the PD calculation.
We confirmed that our version and the original version output exactly the same PD value on all the front shape functions used in our study.
The source codes used in our experiments can be downloaded from the supplementary website ({\url{https://sites.google.com/view/optmudist}).



We set the reference vector $\vector{q}$ for HV and NR2 to $(1.2, ..., 1.2)^{\rm T}$ according to \cite{IshibuchiISN18ecj}.
We set $\vector{z}^*$ for R2 to $(0, ..., 0)^{\rm T}$.
As mentioned in Subsection \ref{sec:proposed_method_sub}, all objective values are in the range $[0,1]$ for all front shape functions.
For R2 and NR2, we generated the weight vector set $\vector{W}$ using simplex-lattice design \cite{DasD98}.
For IGD, IGD$^+$, $I_{\epsilon+}$, and $\Delta$, we generated the reference vector set $\vector{R}$ by applying each front shape function $F$ to $\vector{W}$ as follows $\vector{R} = F(\vector{W})$.
That is, in order to generate $\vector{R}$, $\vector{W}$ is used in \eqref{eqn:proposed_m_def_objvec} instead of $\vector{B}$ (the resulting $\vector{A}$ is used as $\vector{R}$).
For details, see Section S.2 in the supplementary file.
The size of $\vector{W}$ and $\vector{R}$ was set to $1\,035$.
For $F_{\rm disconnected}$ and $F_{\rm c\shyp concave}$, we set the size of $\vector{R}$ to $1\,089$ and $1\,087$, respectively.
This is due to the properties of $F_{\rm disconnected}$ and $F_{\rm c\shyp concave}$.






%

\begin{figure*}[htp]
\begin{tabular}{ccc}
  \newcommand{\widthvar}{0.41}    
 \begin{minipage}[t]{0.33\hsize}
   \centering
   \subfloat[$\vector{A}_{\rm HV}$]{\includegraphics[width=\widthvar\textwidth]{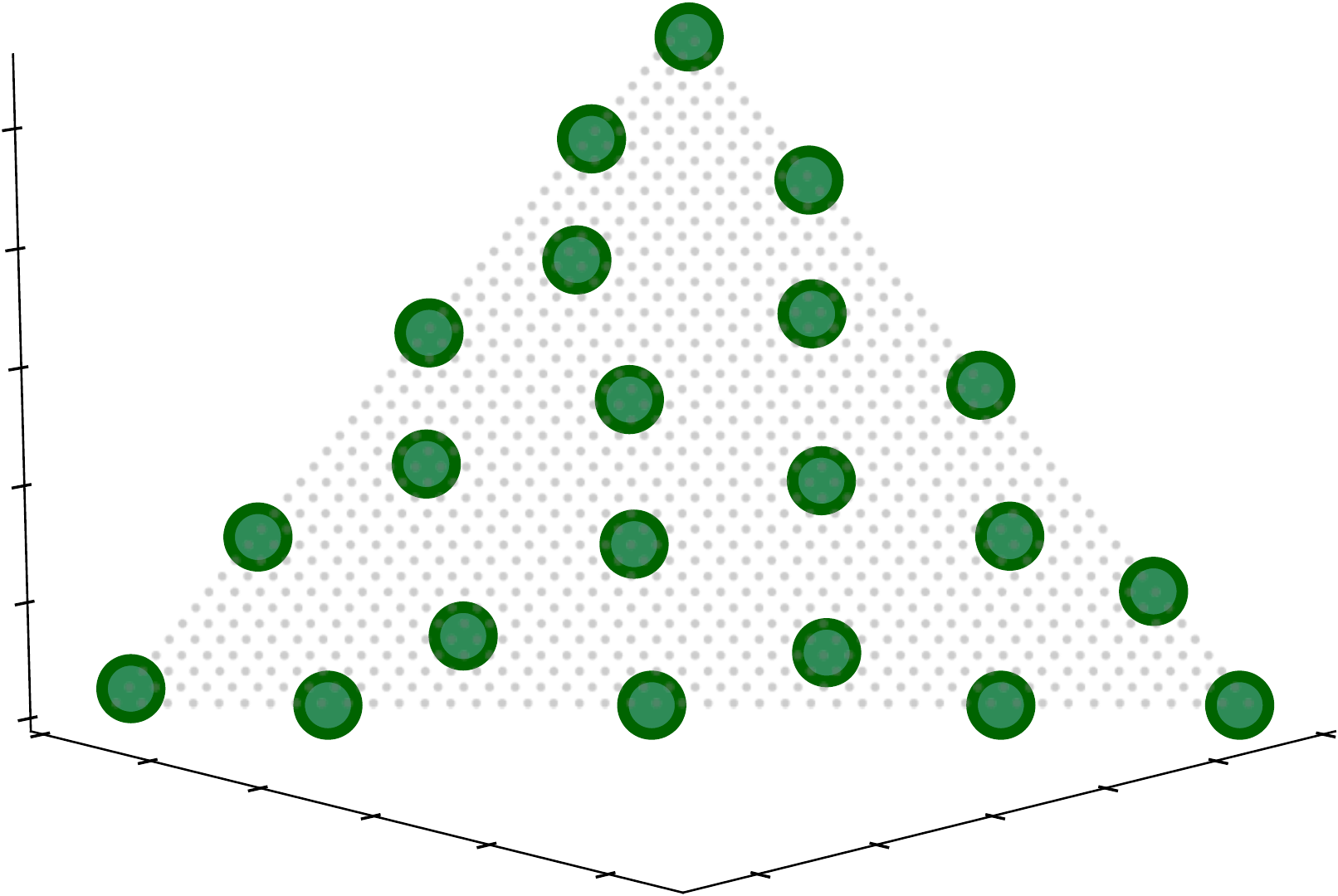}}
\subfloat[$\vector{A}_{\rm IGD}$]{\includegraphics[width=\widthvar\textwidth]{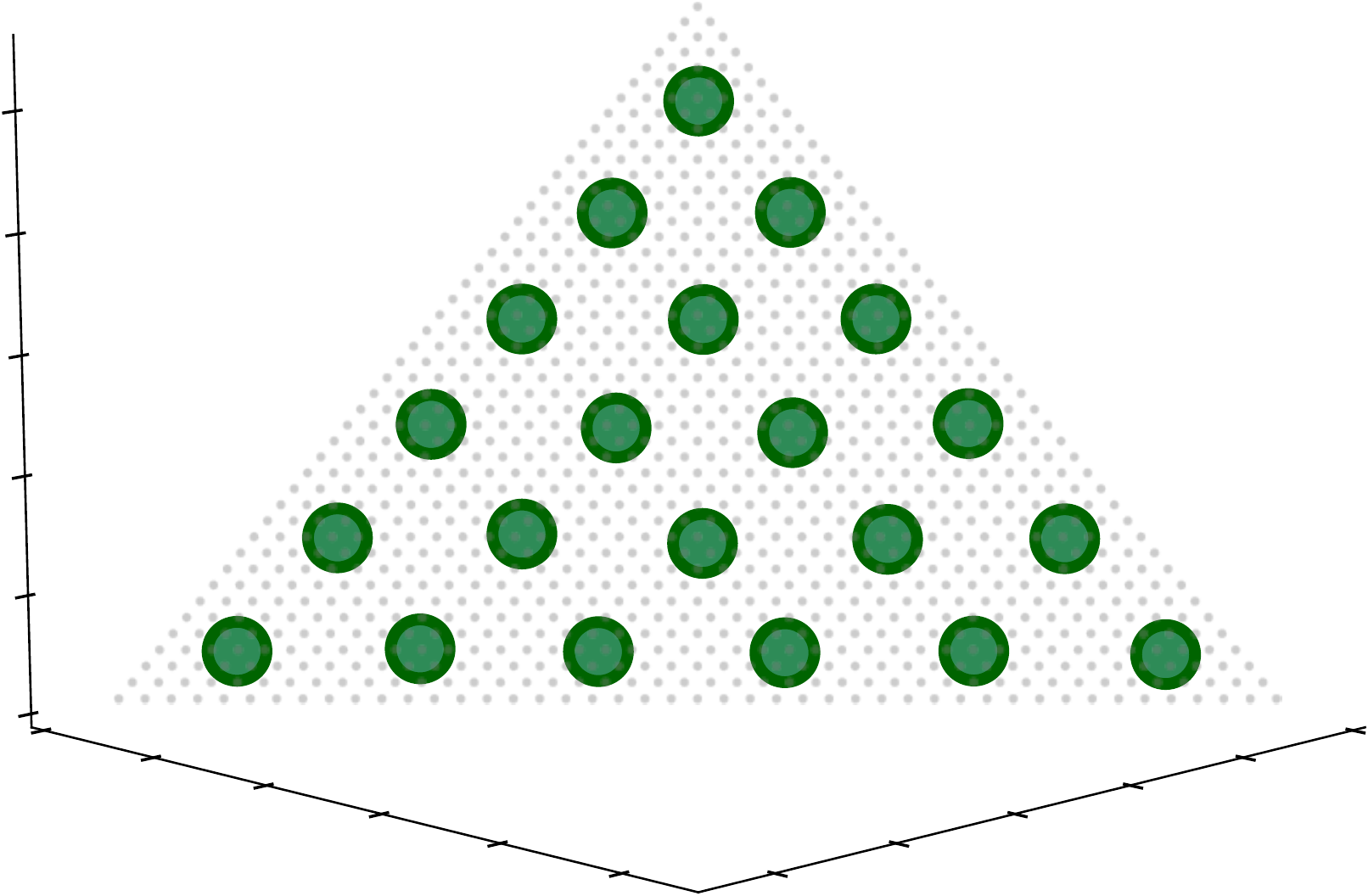}}
\\
\subfloat[$\vector{A}_{\rm IGD^+}$]{\includegraphics[width=\widthvar\textwidth]{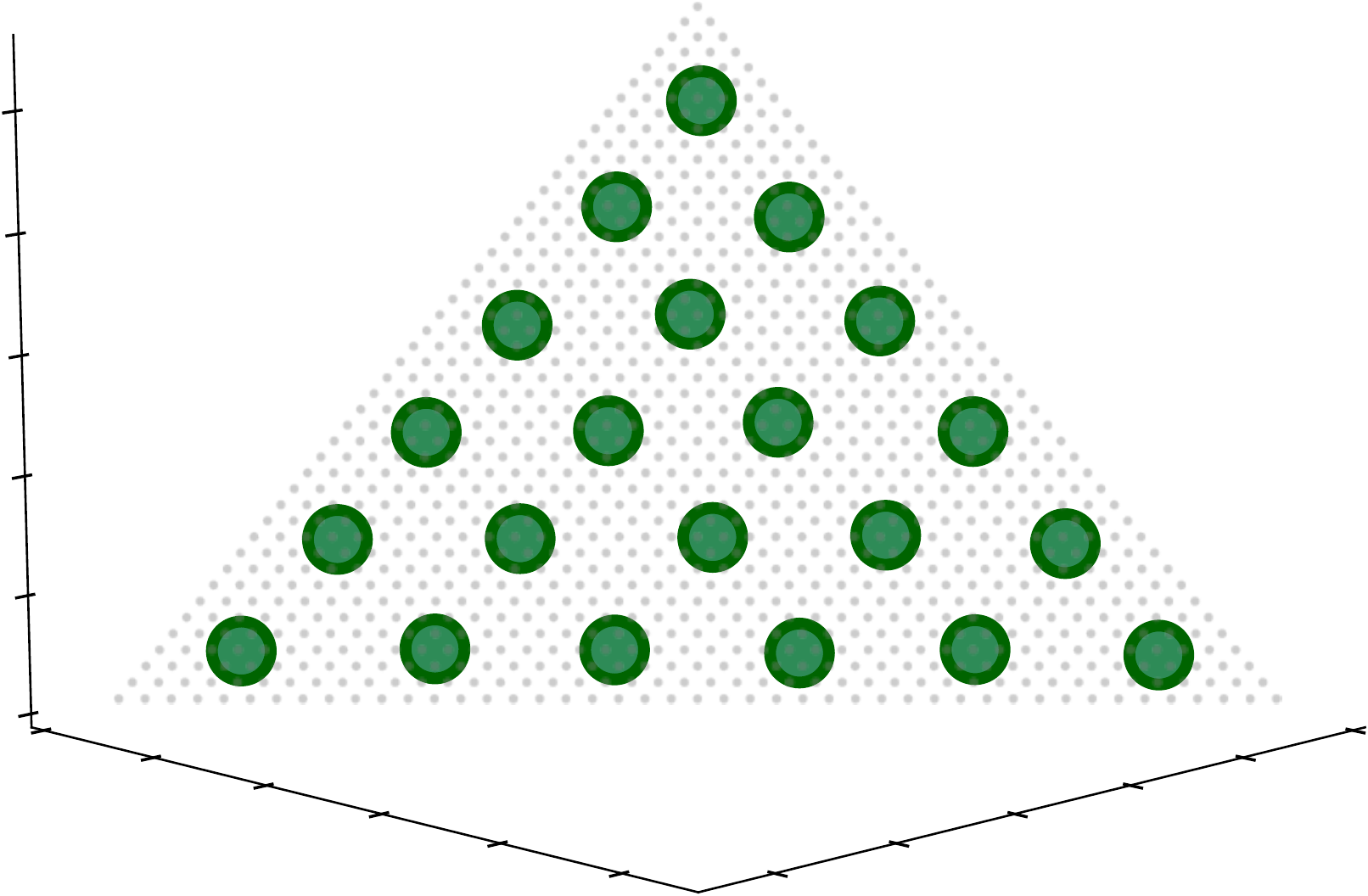}}
\subfloat[$\vector{A}_{\rm R2}$]{\includegraphics[width=\widthvar\textwidth]{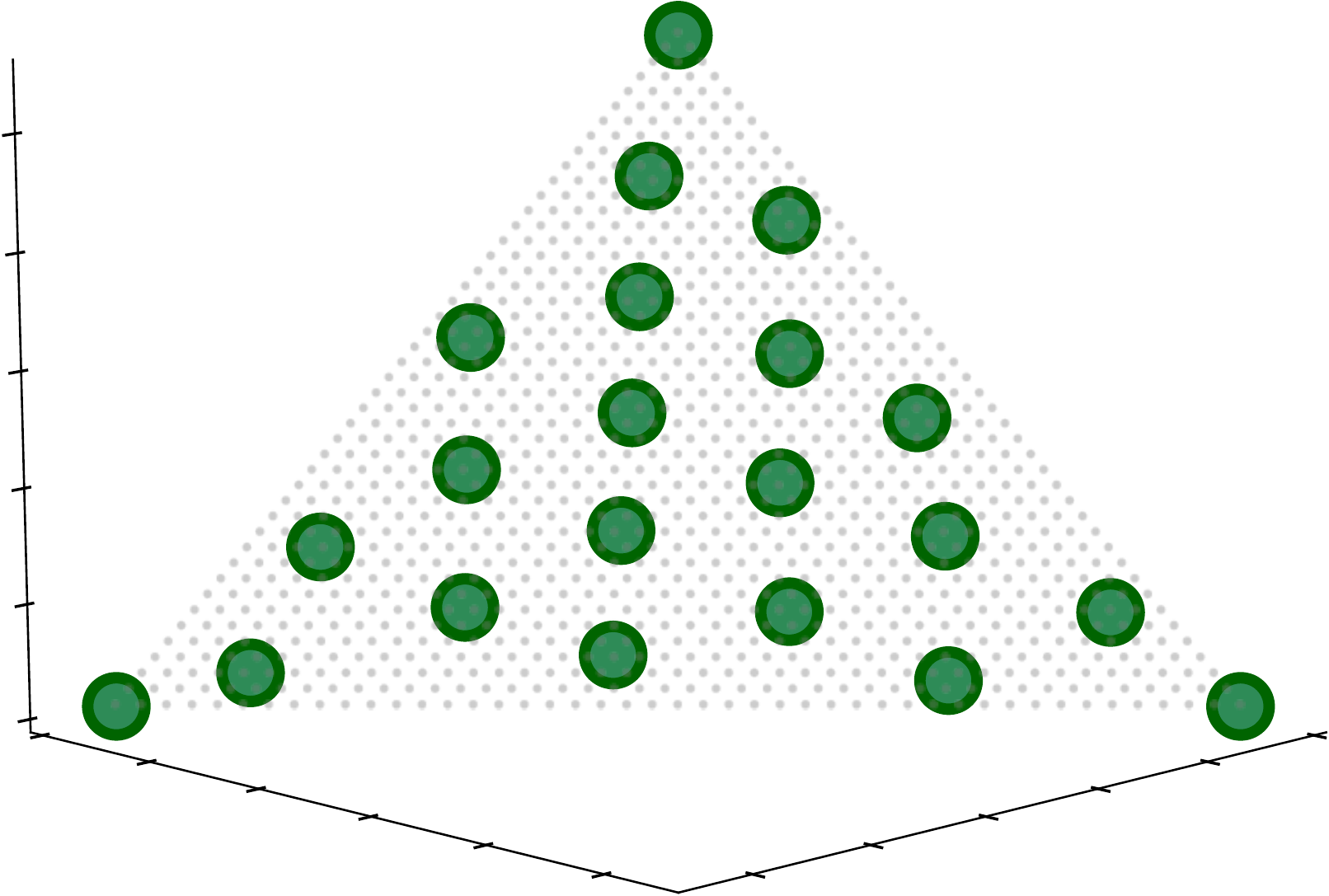}}
\\
\subfloat[$\vector{A}_{\rm NR2}$]{\includegraphics[width=\widthvar\textwidth]{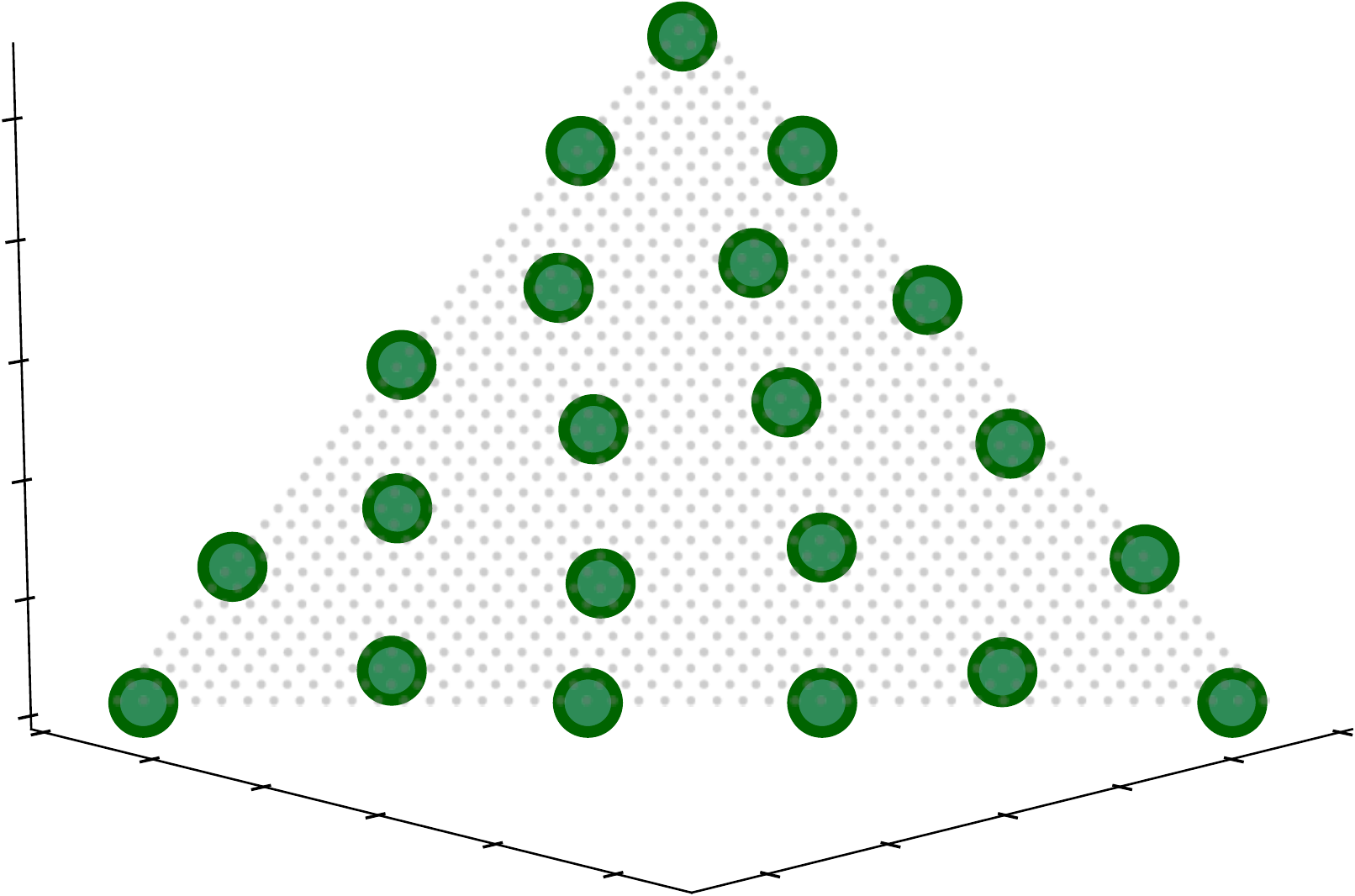}}
\subfloat[$\vector{A}_{I_{\epsilon+}}$]{\includegraphics[width=\widthvar\textwidth]{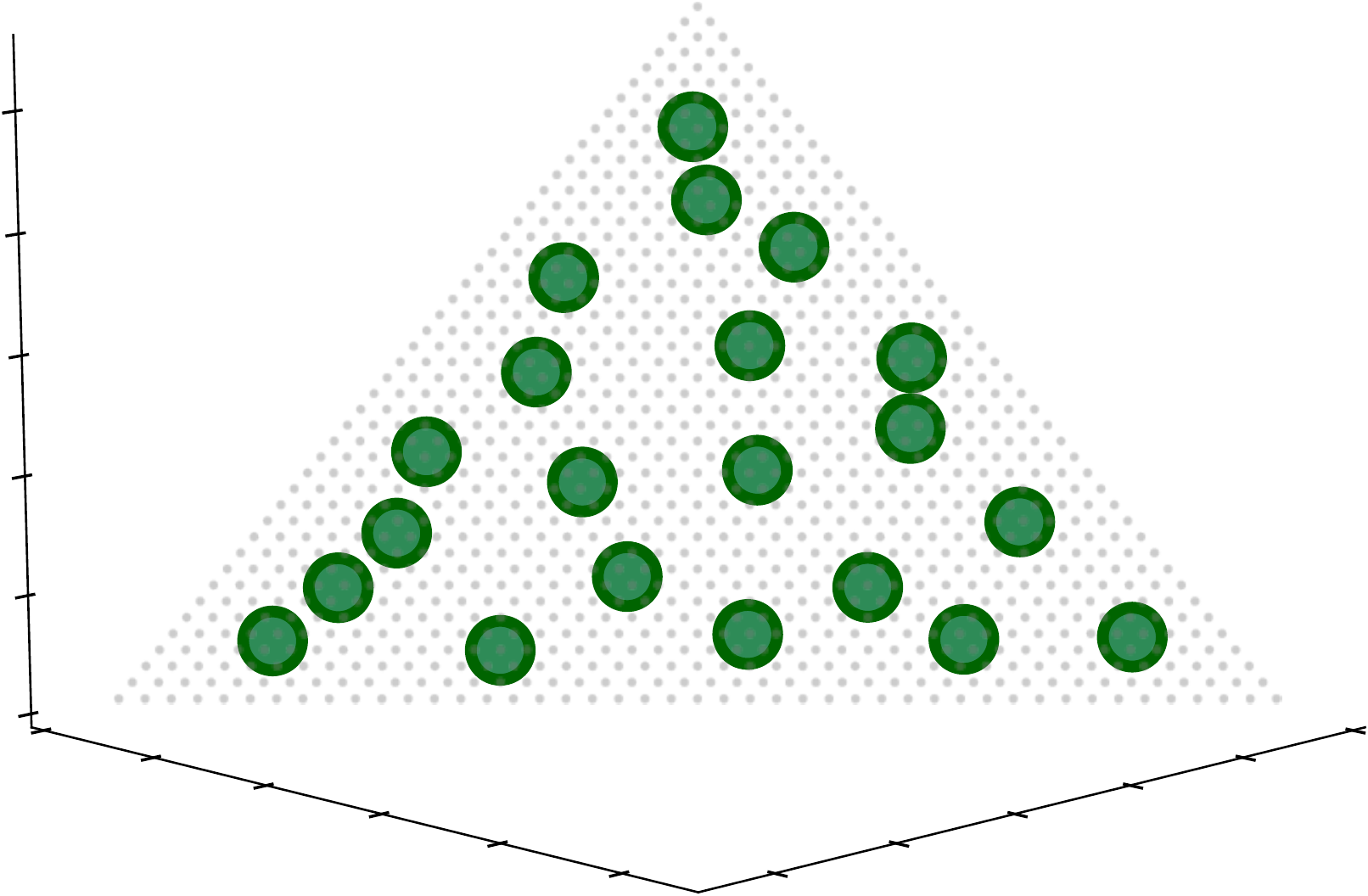}}
\\
\subfloat[$\vector{A}_{\rm SE}$]{\includegraphics[width=\widthvar\textwidth]{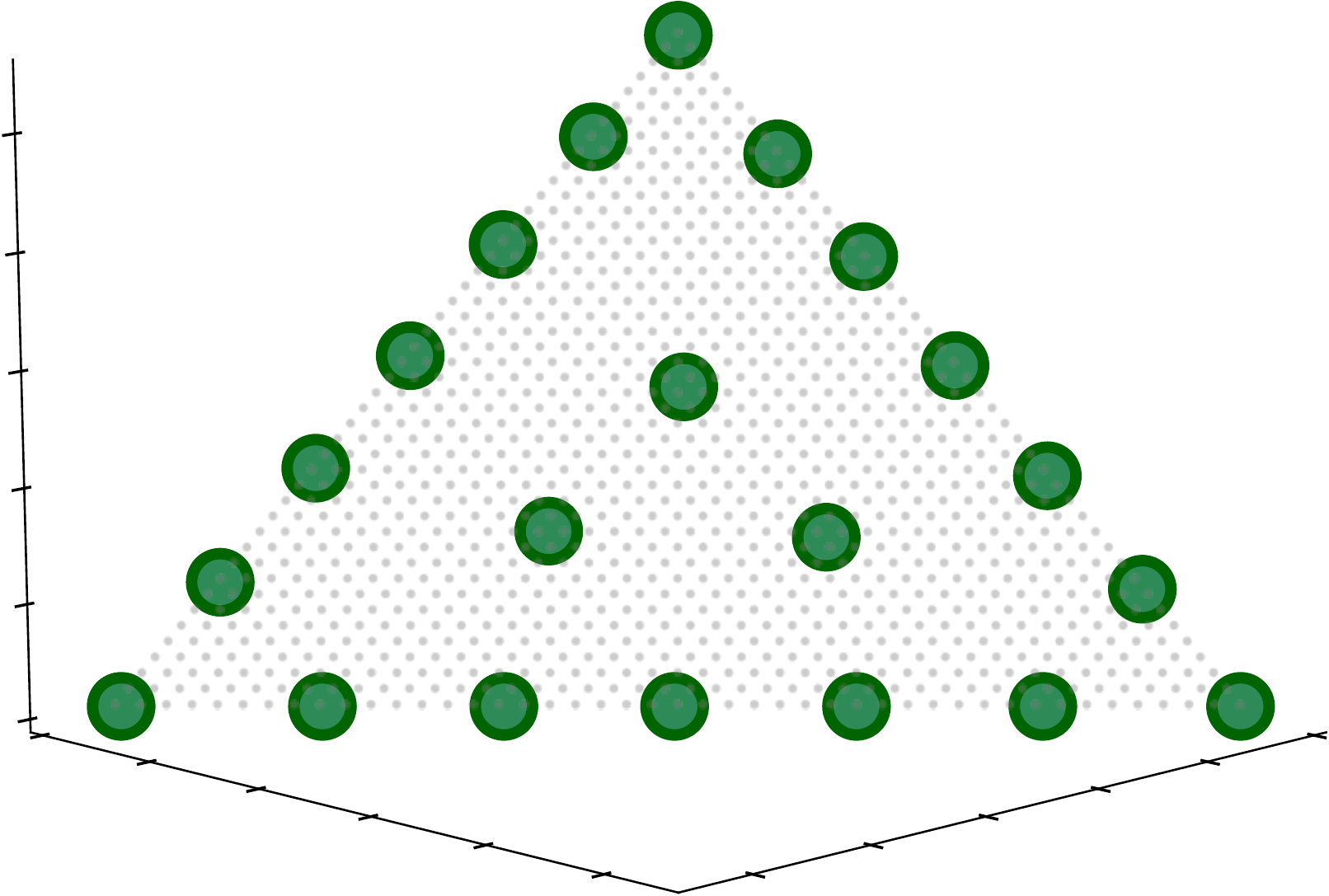}}
\subfloat[$\vector{A}_{\Delta}$]{\includegraphics[width=\widthvar\textwidth]{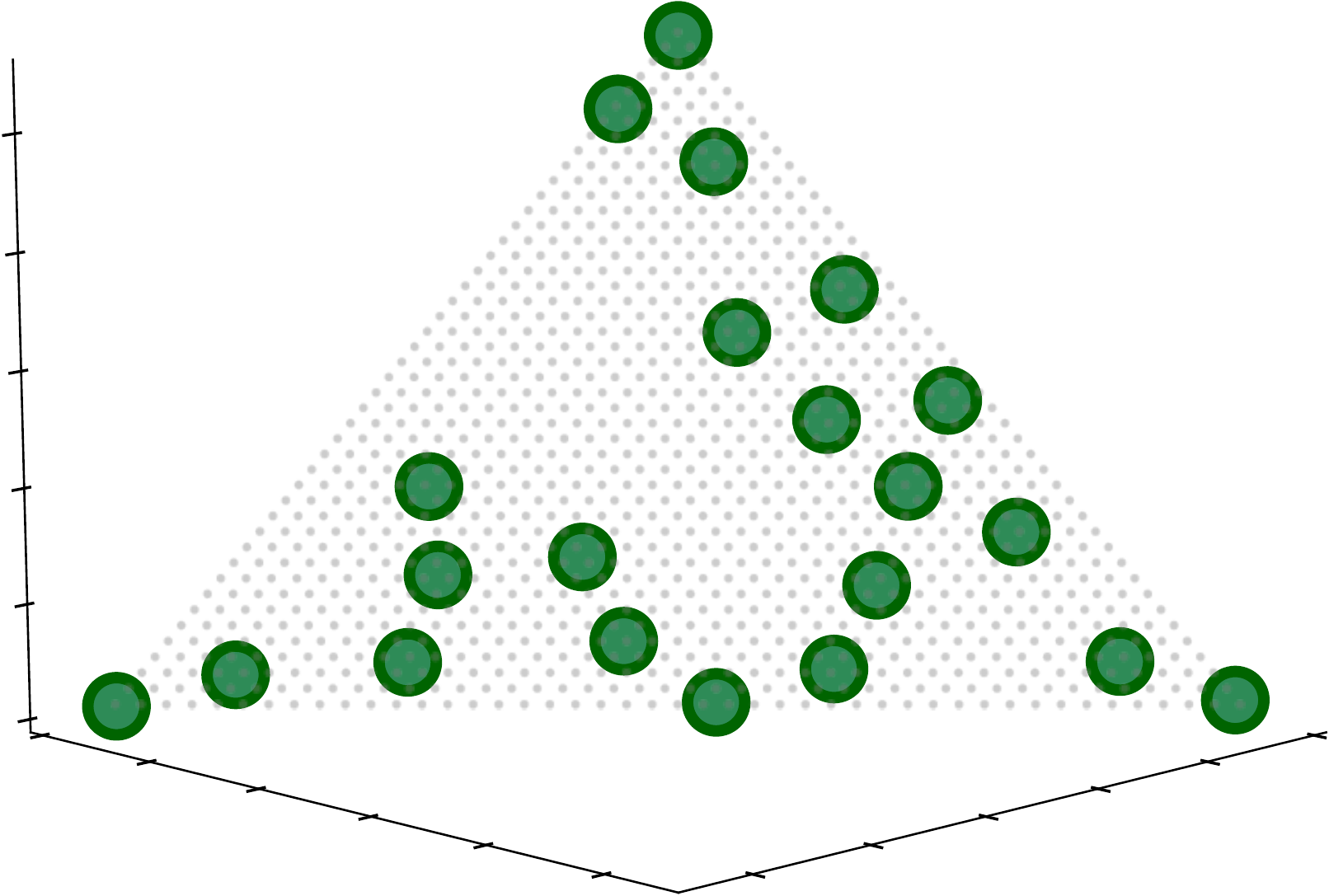}}
\\
\subfloat[$\vector{A}_{\rm PD}$]{\includegraphics[width=\widthvar\textwidth]{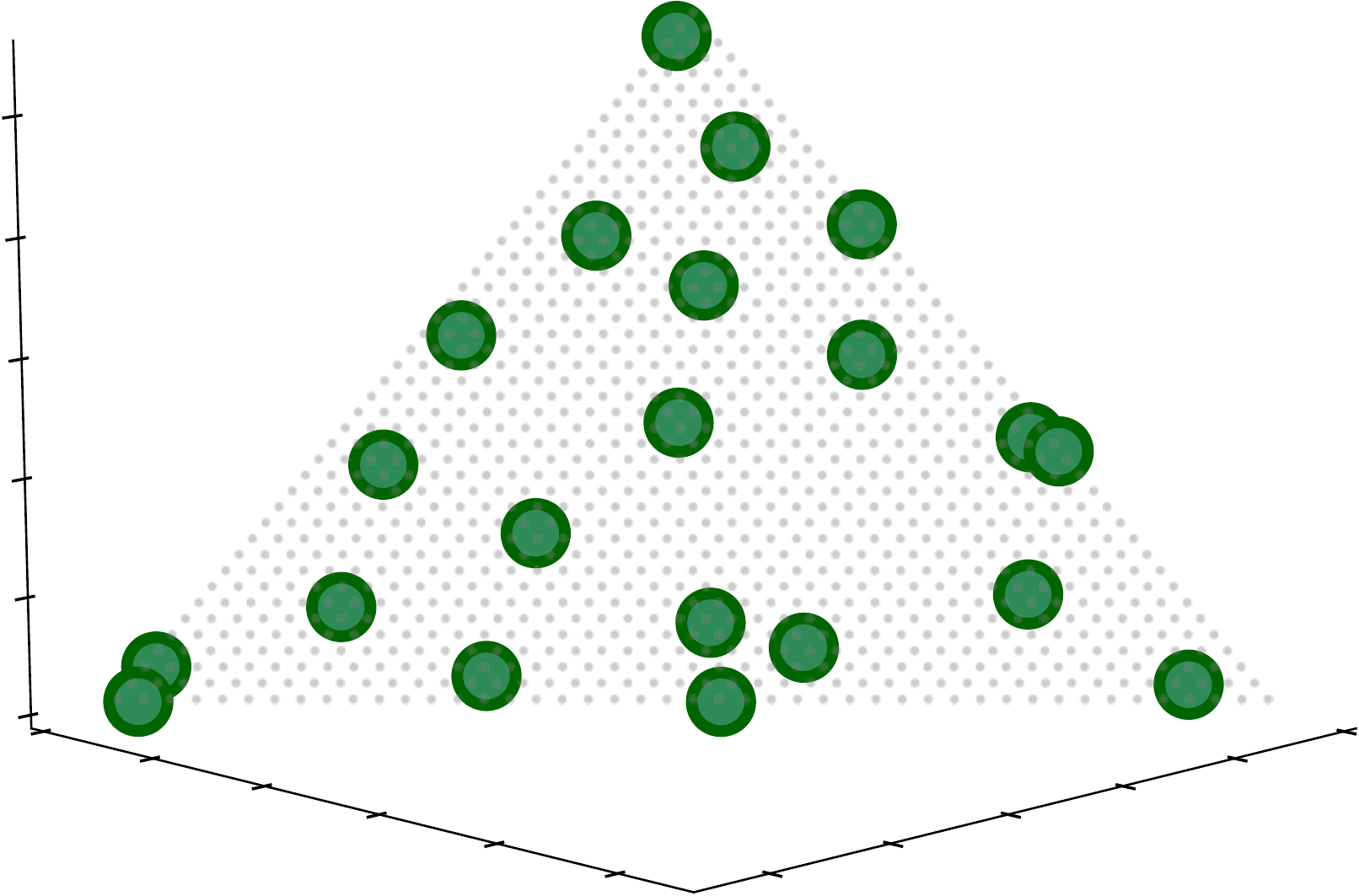}}
\subfloat[$\vector{A}_{\rm SLD}$]{\includegraphics[width=\widthvar\textwidth]{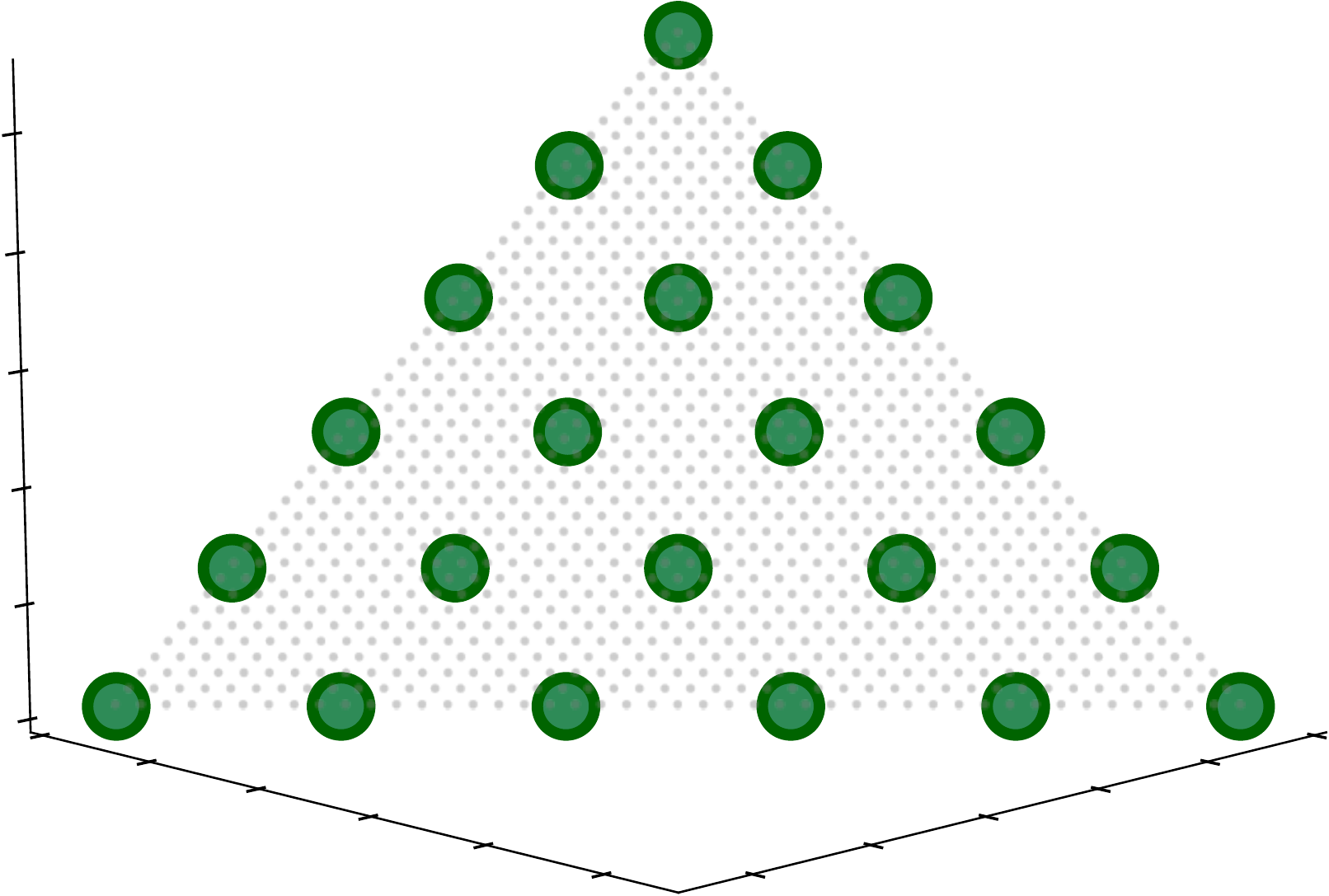}}
\caption{
\small
Results on $F_{\rm linear}$.
}
\label{fig:linear}
 \end{minipage}
 \begin{minipage}[t]{0.33\hsize}
   \centering
\subfloat[$\vector{A}_{\rm HV}$]{\includegraphics[width=\widthvar\textwidth]{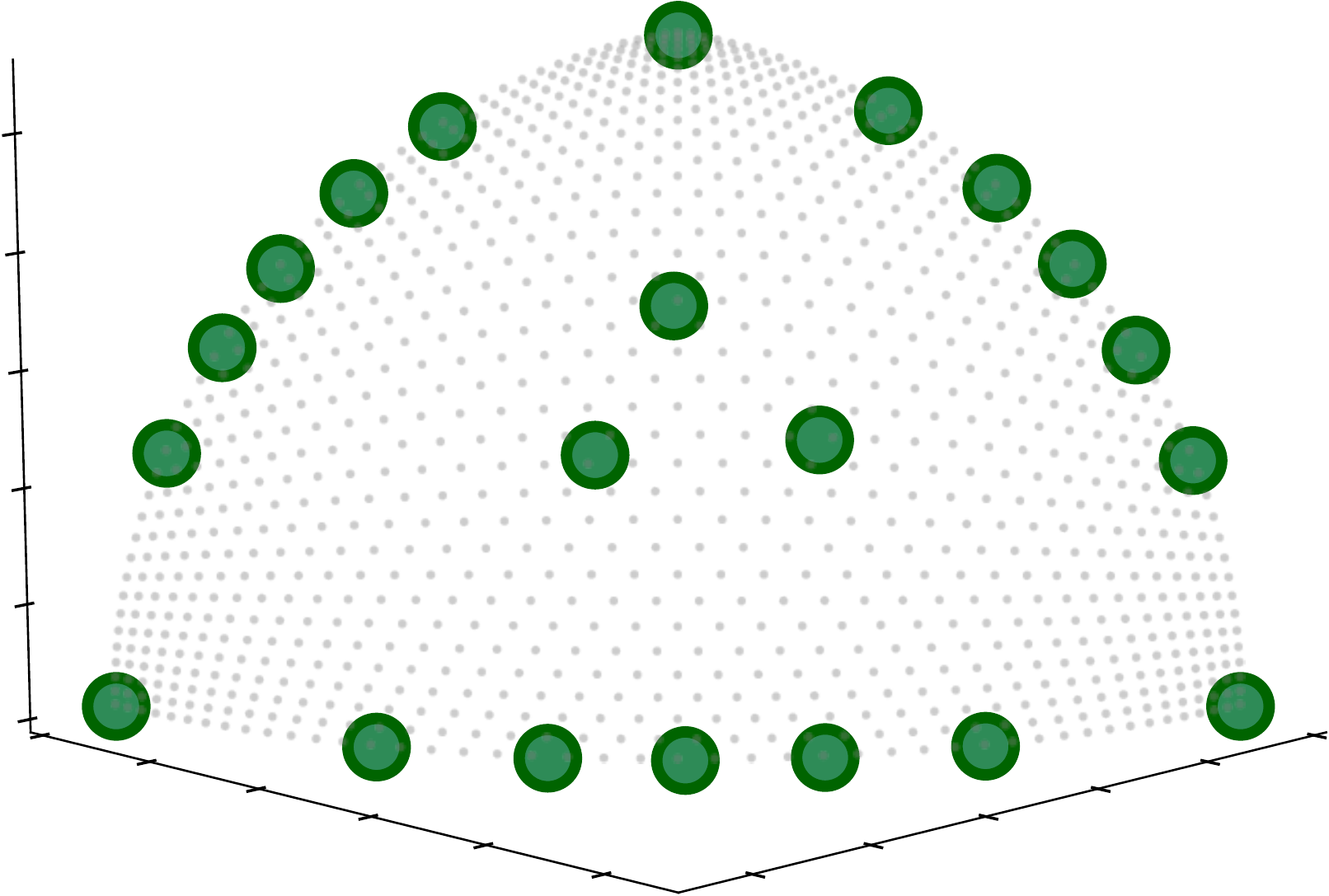}}
\subfloat[$\vector{A}_{\rm IGD}$]{\includegraphics[width=\widthvar\textwidth]{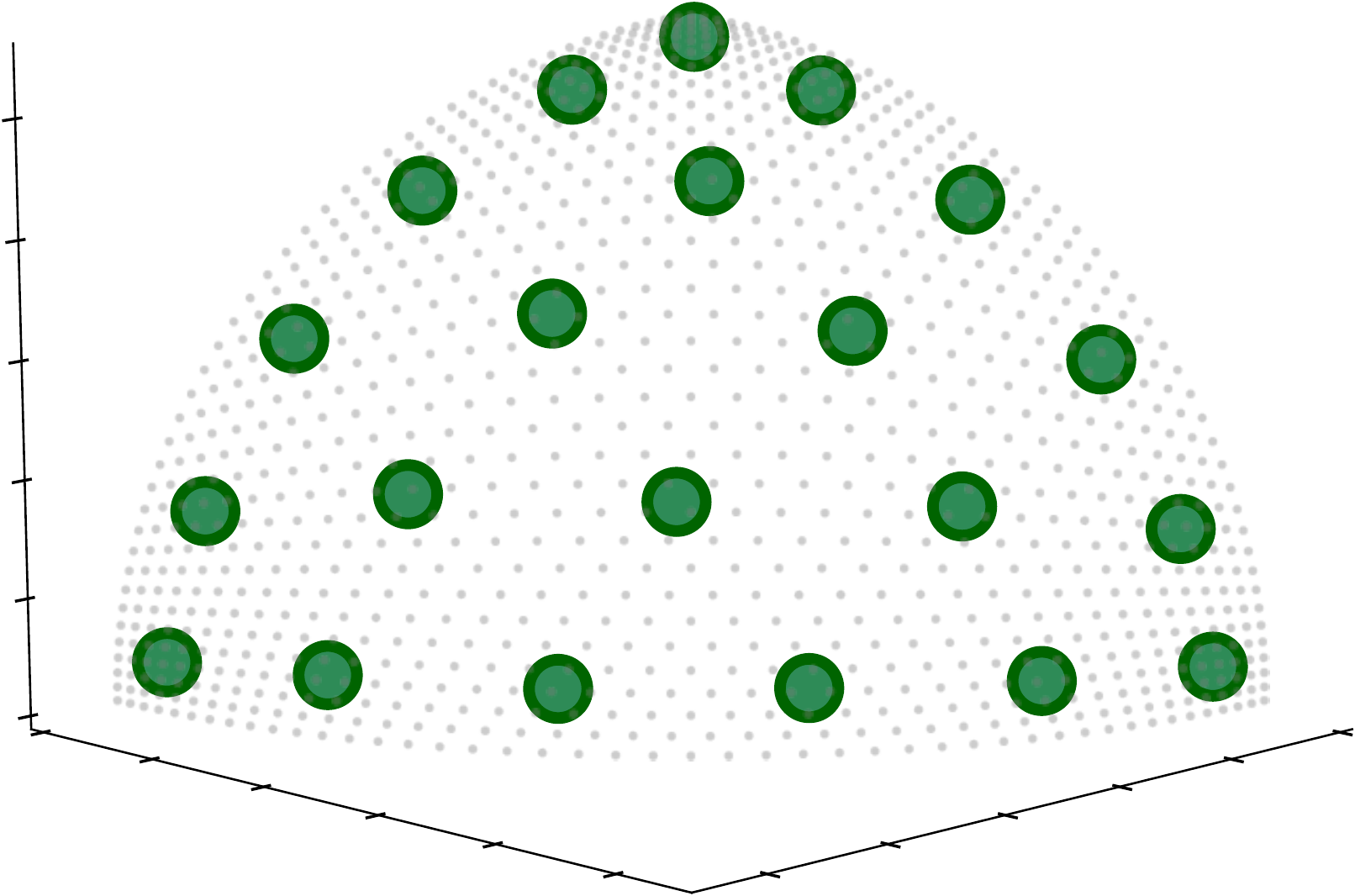}}
\\
\subfloat[$\vector{A}_{\rm IGD^+}$]{\includegraphics[width=\widthvar\textwidth]{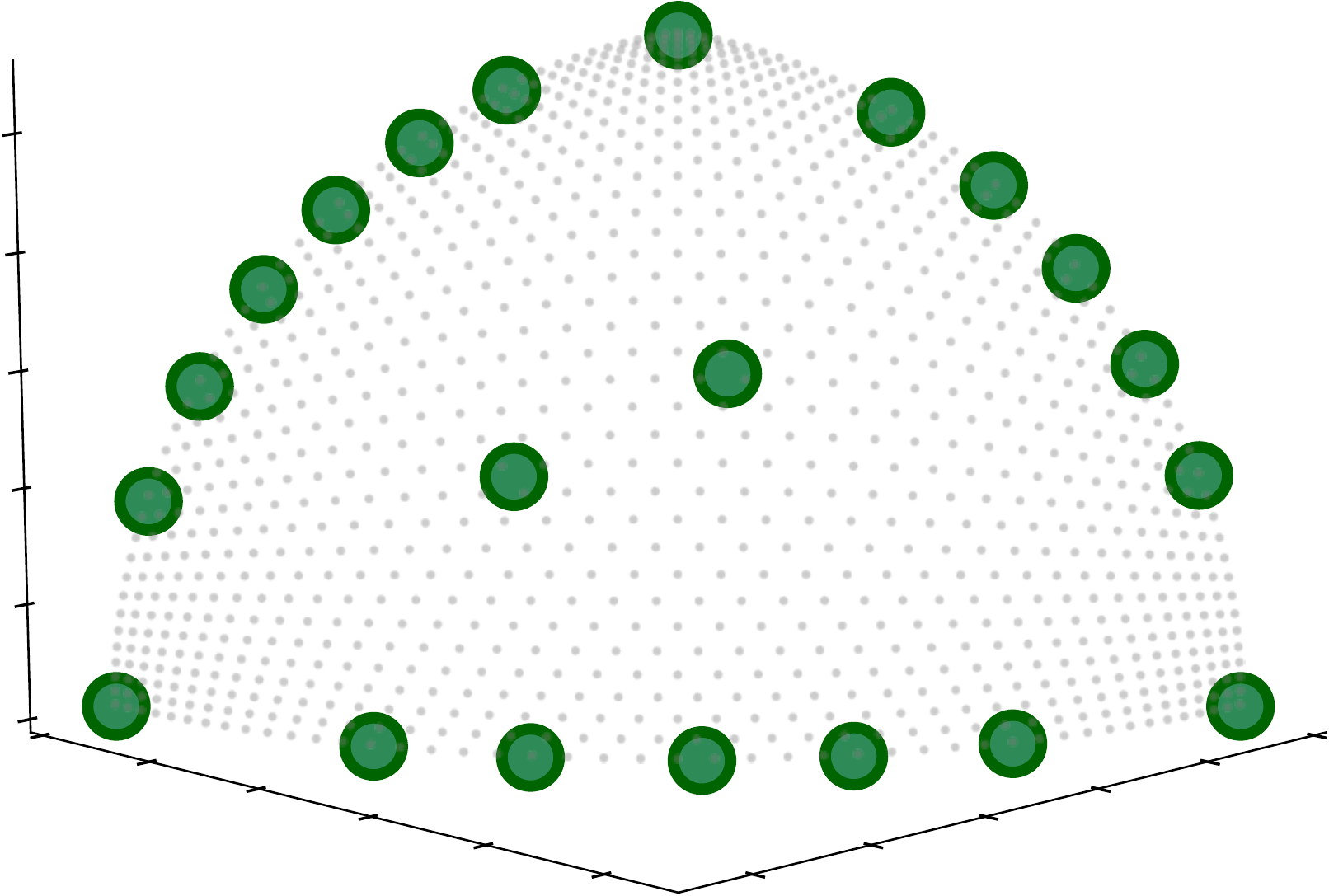}}
\subfloat[$\vector{A}_{\rm R2}$]{\includegraphics[width=\widthvar\textwidth]{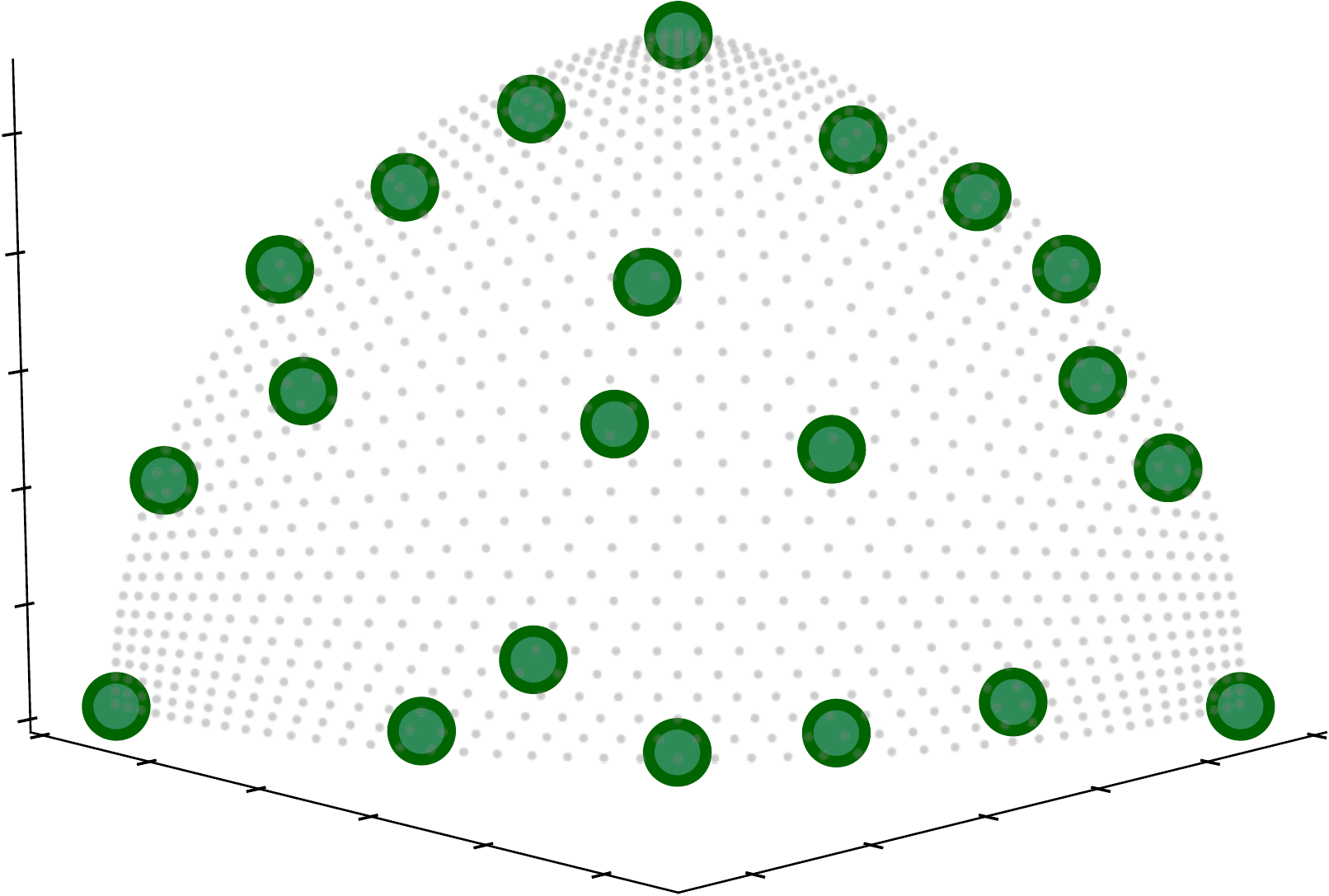}}
\\
\subfloat[$\vector{A}_{\rm NR2}$]{\includegraphics[width=\widthvar\textwidth]{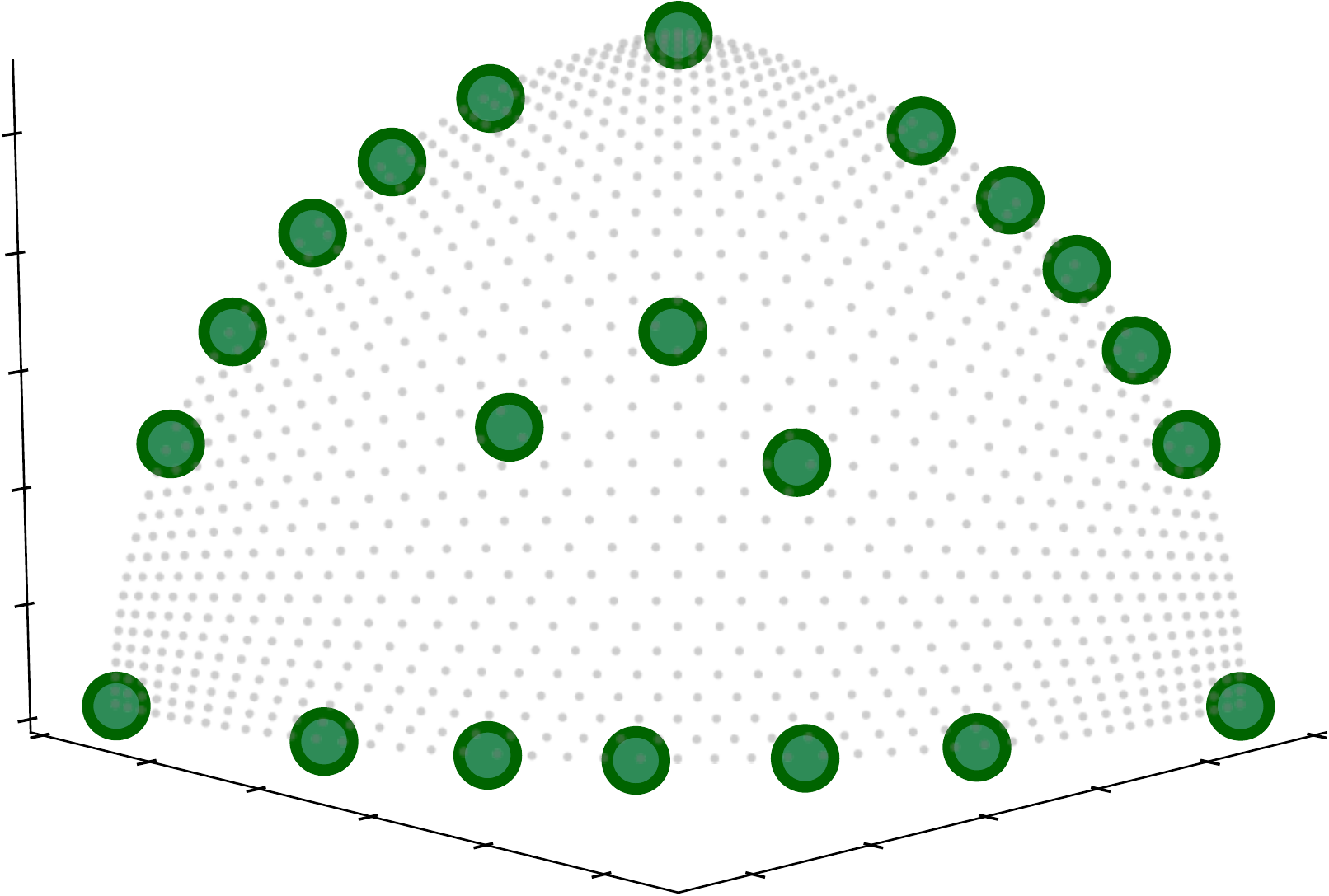}}
\subfloat[$\vector{A}_{I_{\epsilon+}}$]{\includegraphics[width=\widthvar\textwidth]{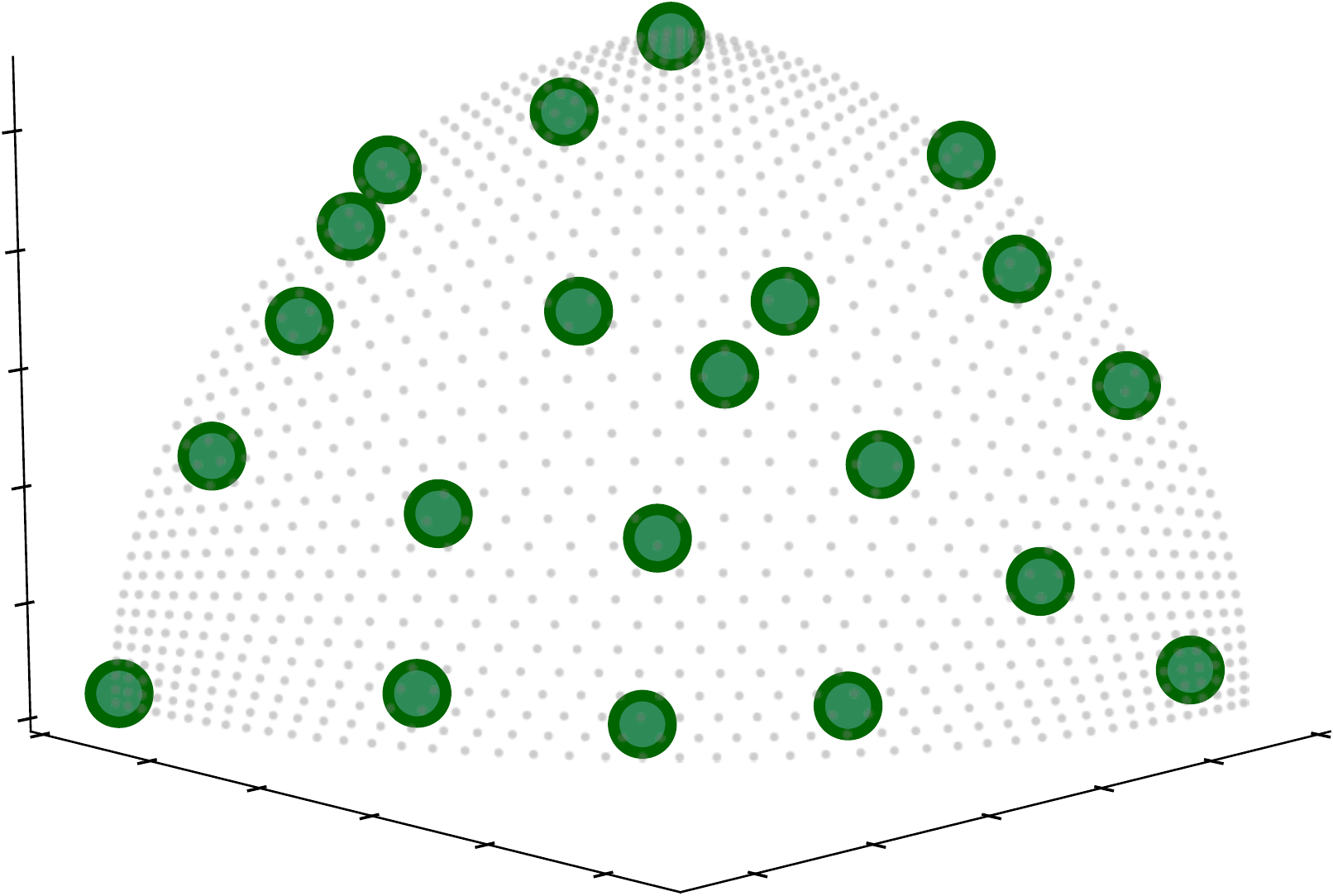}}
\\
\subfloat[$\vector{A}_{\rm SE}$]{\includegraphics[width=\widthvar\textwidth]{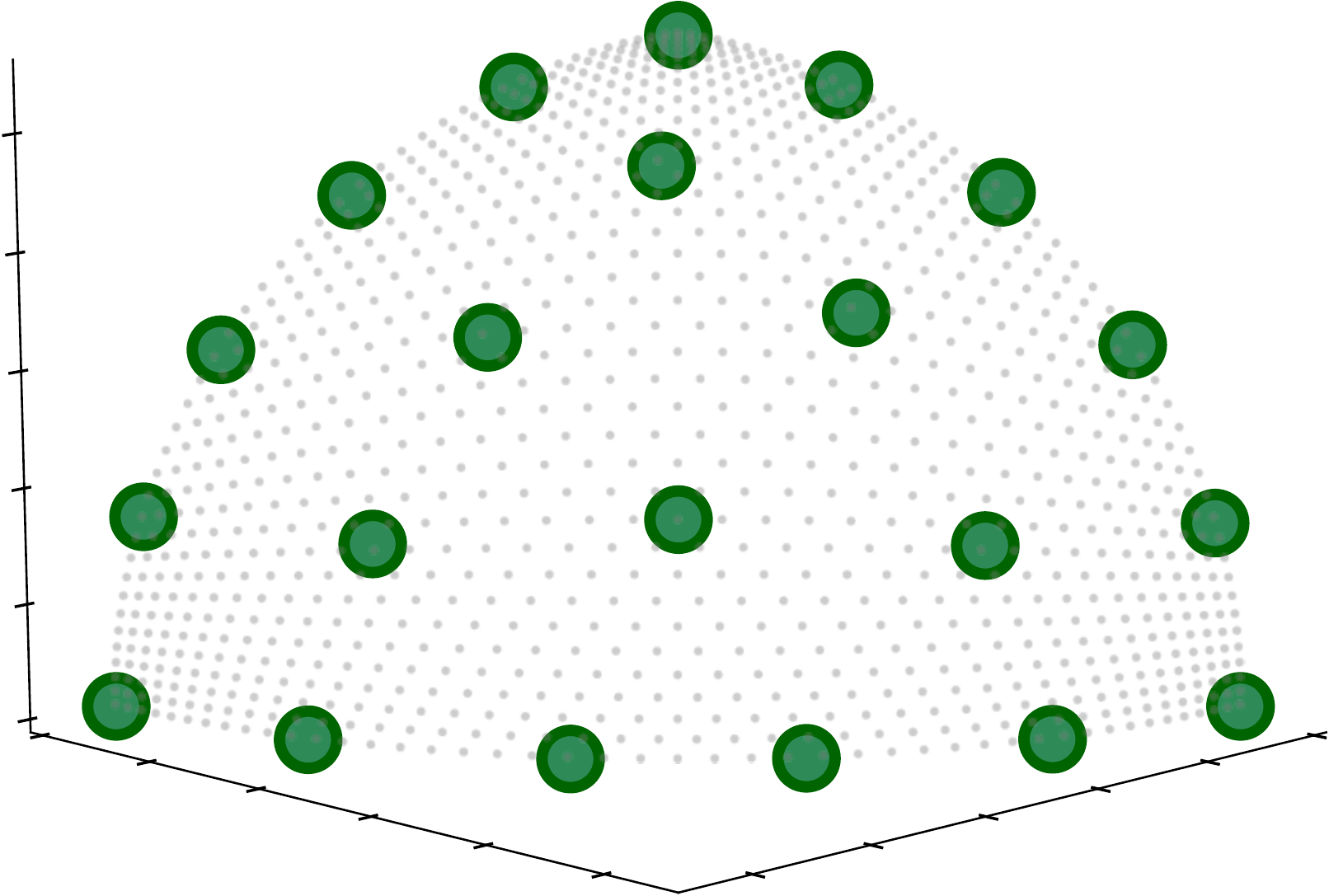}}
\subfloat[$\vector{A}_{\Delta}$]{\includegraphics[width=\widthvar\textwidth]{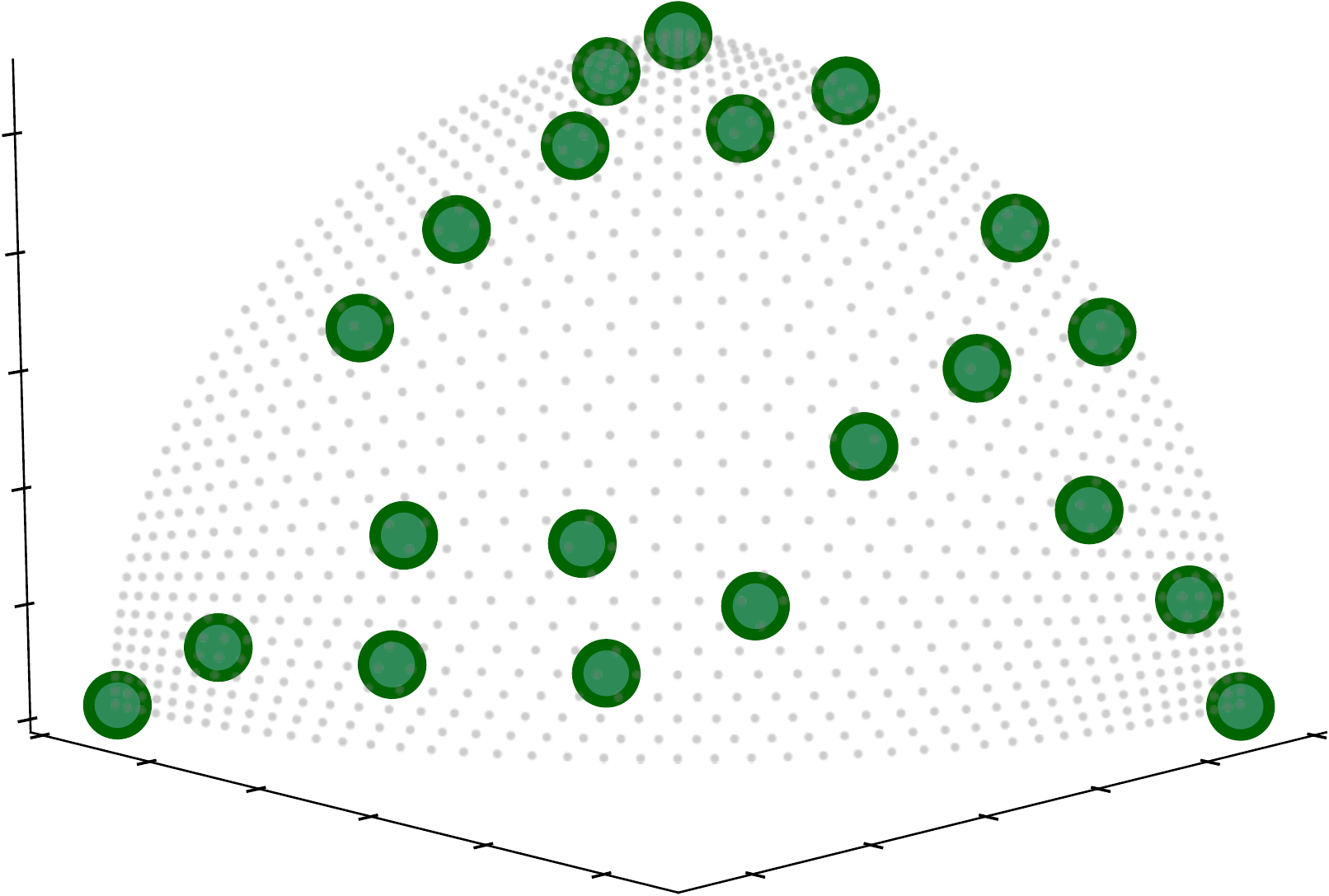}}
\\
\subfloat[$\vector{A}_{\rm PD}$]{\includegraphics[width=\widthvar\textwidth]{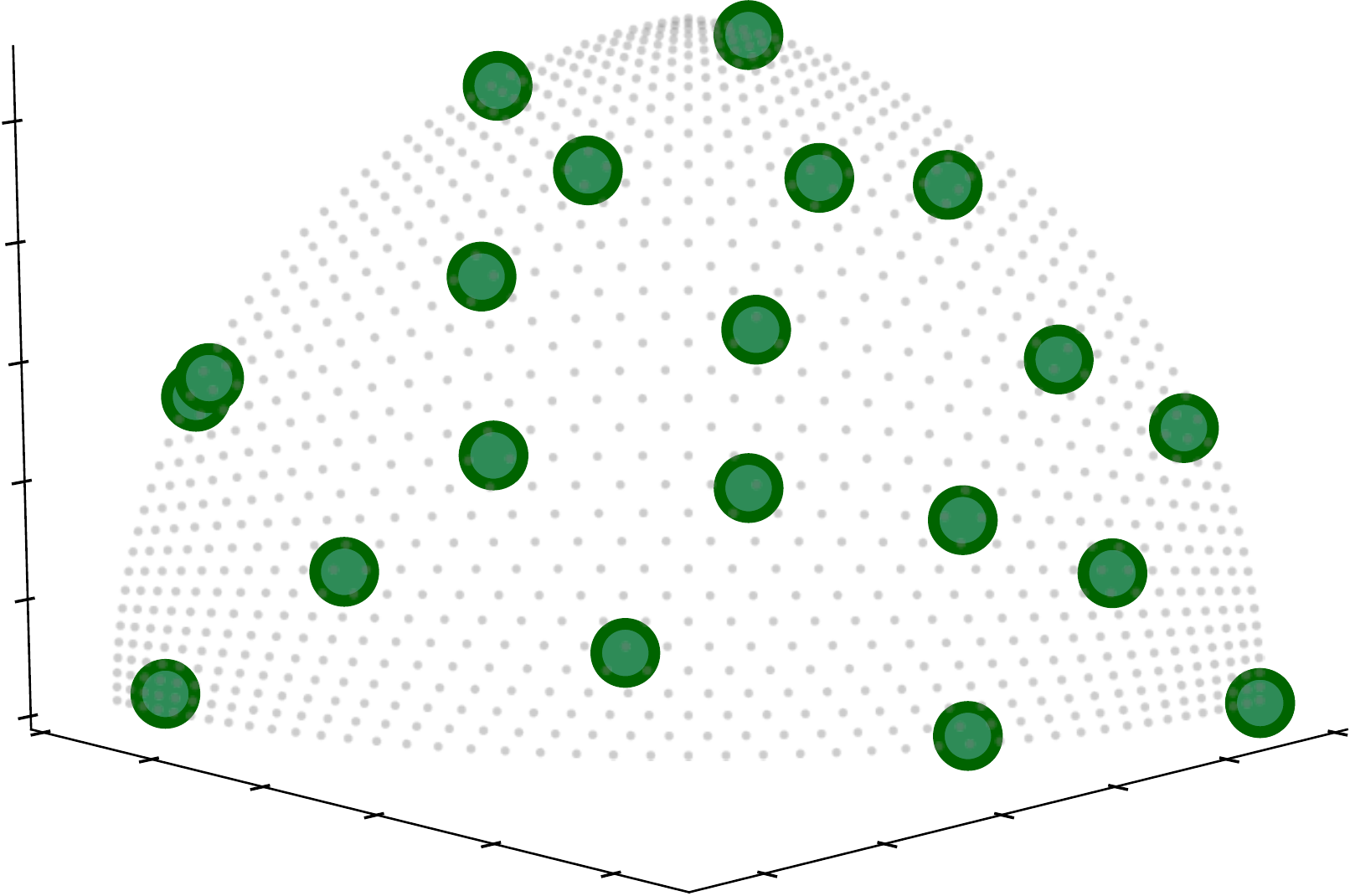}}
\subfloat[$\vector{A}_{\rm SLD}$]{\includegraphics[width=\widthvar\textwidth]{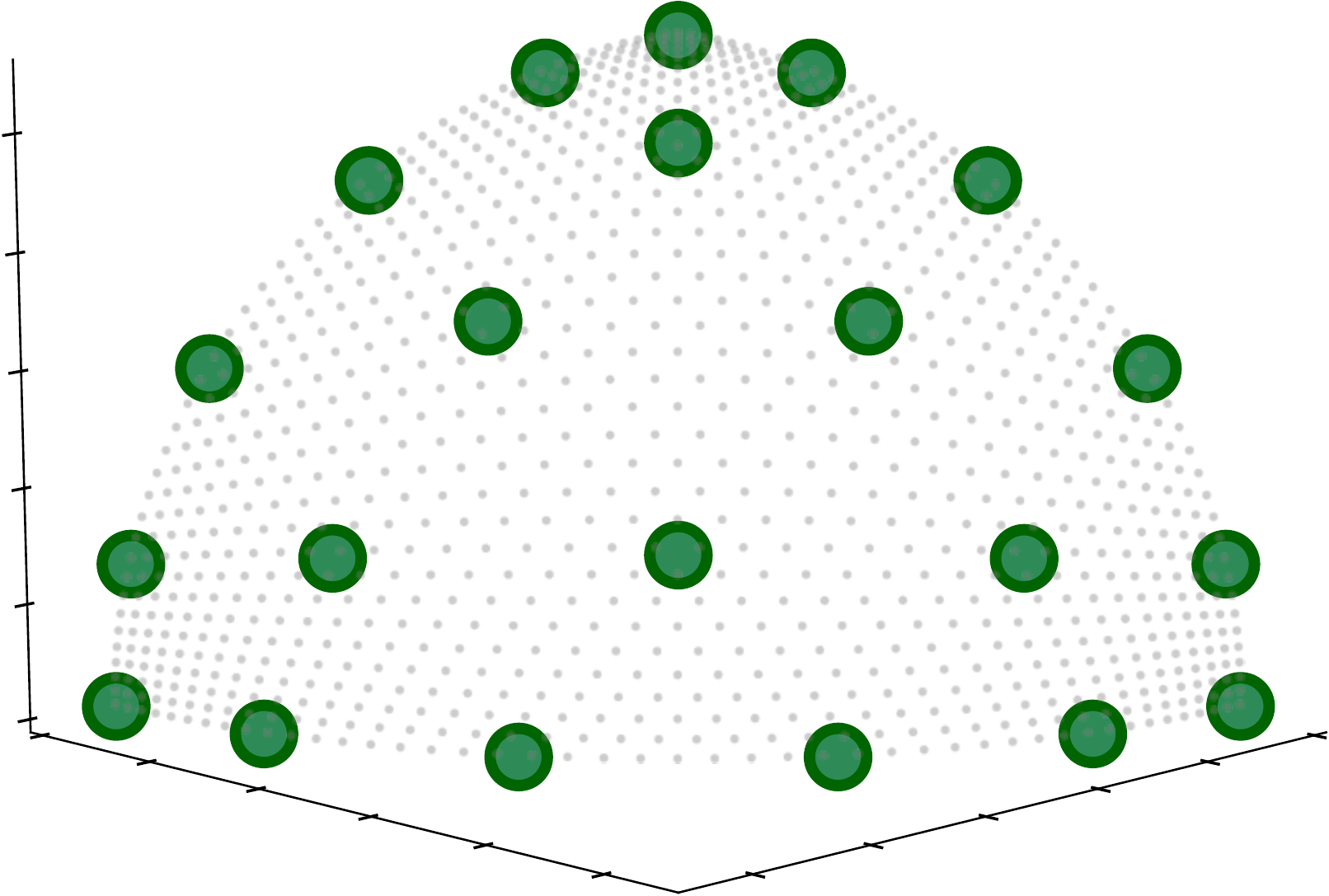}}
\caption{
\small
Results on $F_{\rm concave}$.
}
\label{fig:nonconvex}
 \end{minipage}
  \begin{minipage}[t]{0.33\hsize}
   \centering
\subfloat[$\vector{A}_{\rm HV}$]{\includegraphics[width=\widthvar\textwidth]{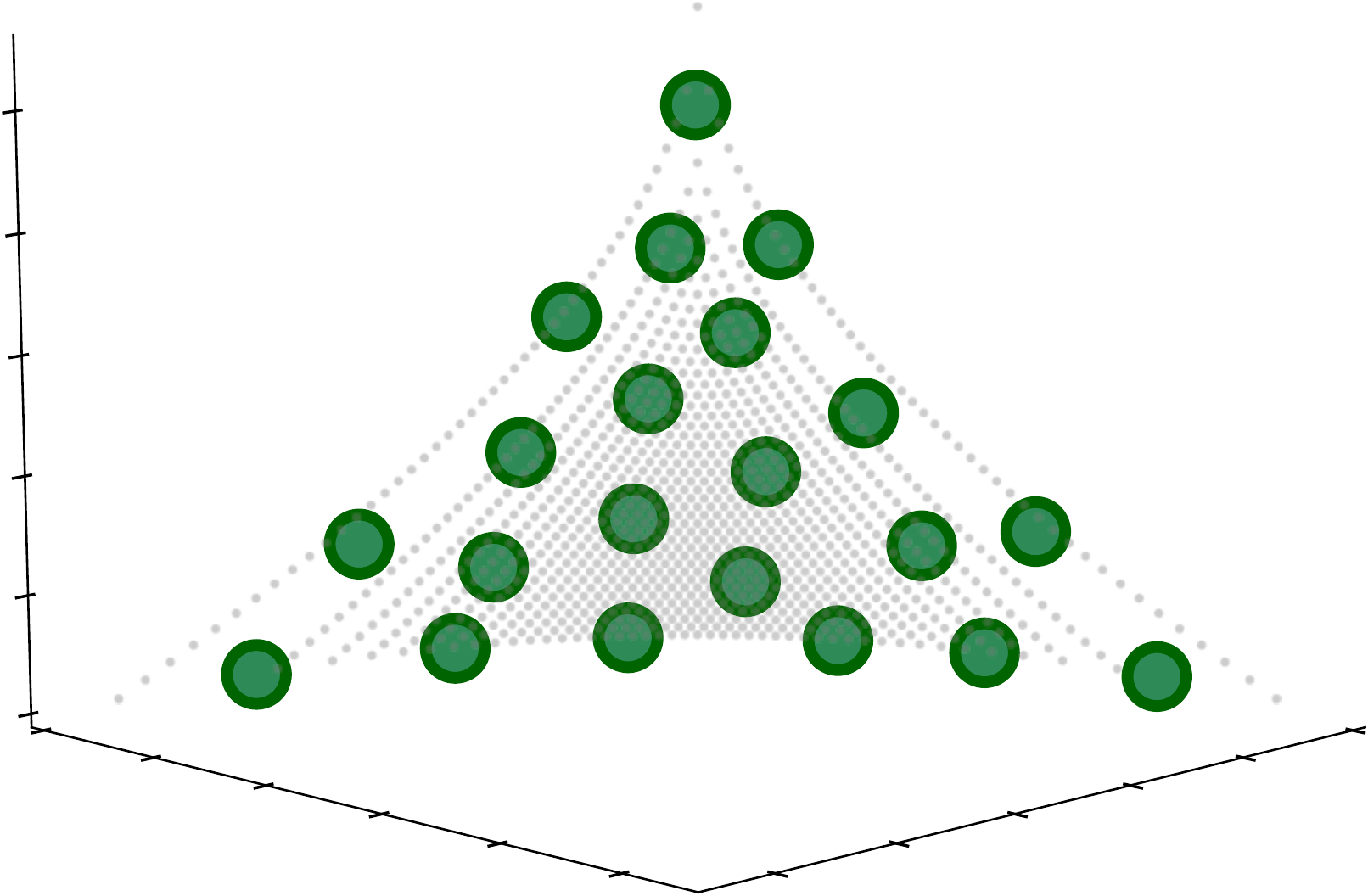}}
\subfloat[$\vector{A}_{\rm IGD}$]{\includegraphics[width=\widthvar\textwidth]{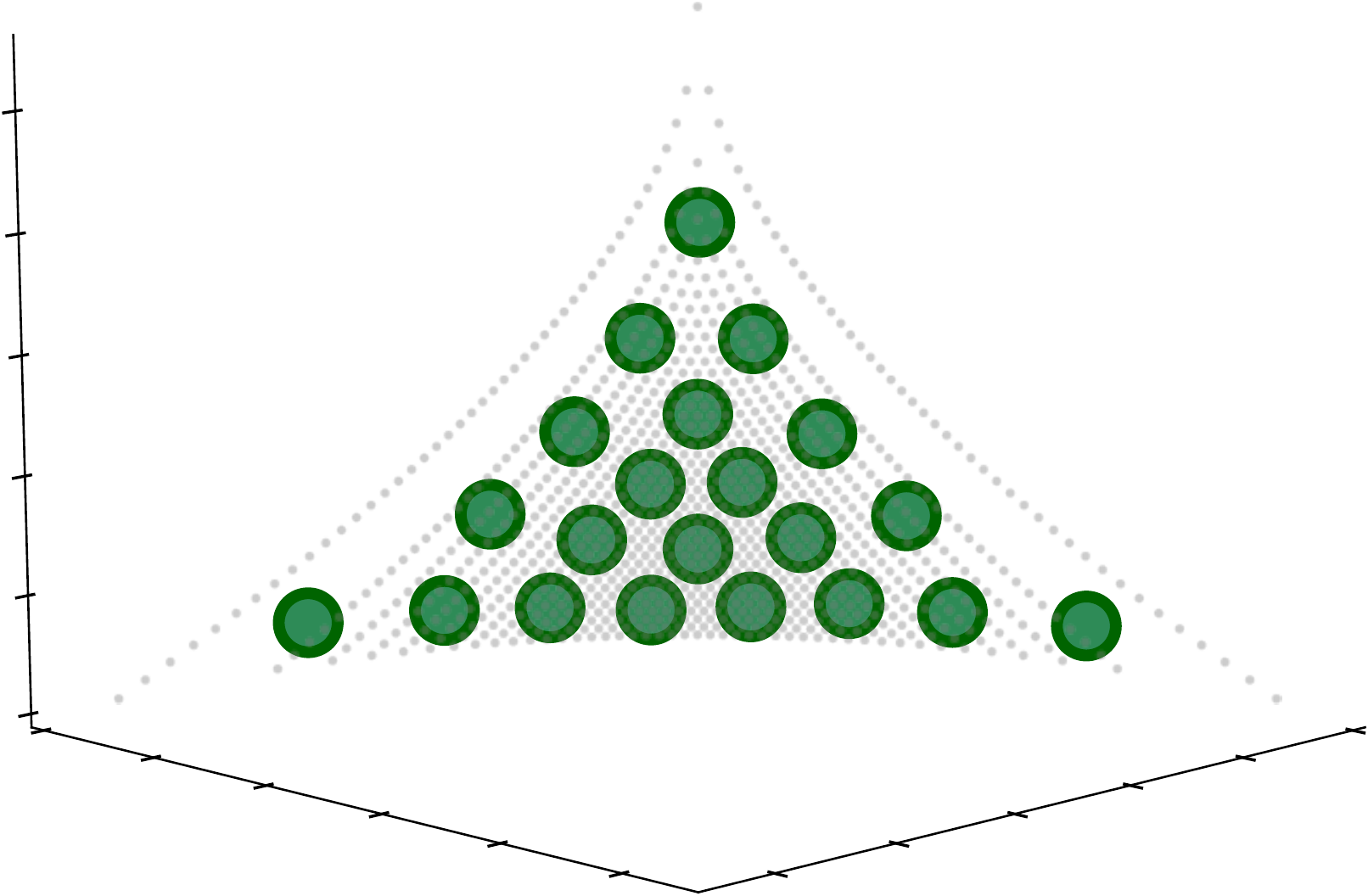}}
\\
\subfloat[$\vector{A}_{\rm IGD^+}$]{\includegraphics[width=\widthvar\textwidth]{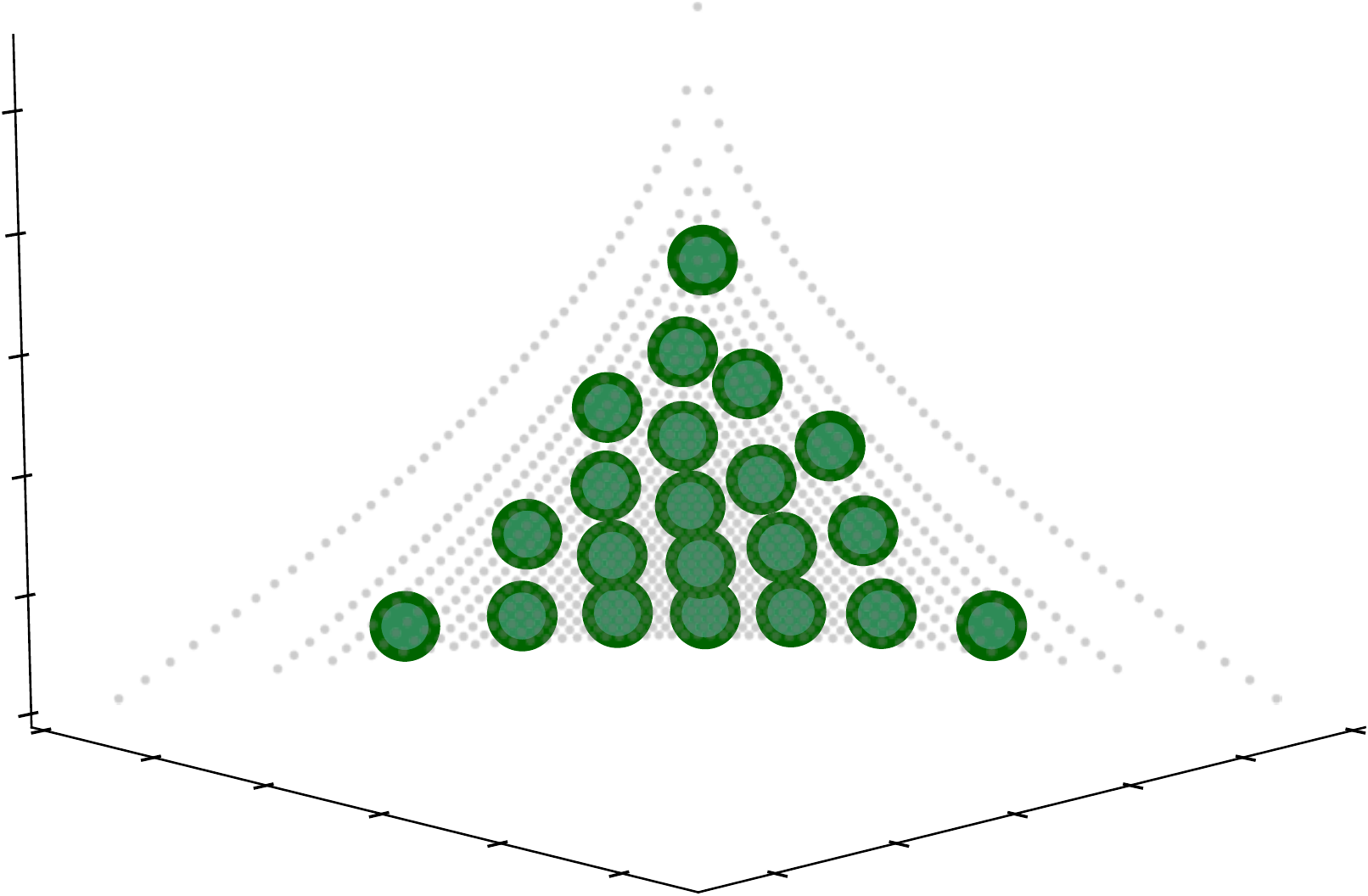}}
\subfloat[$\vector{A}_{\rm R2}$]{\includegraphics[width=\widthvar\textwidth]{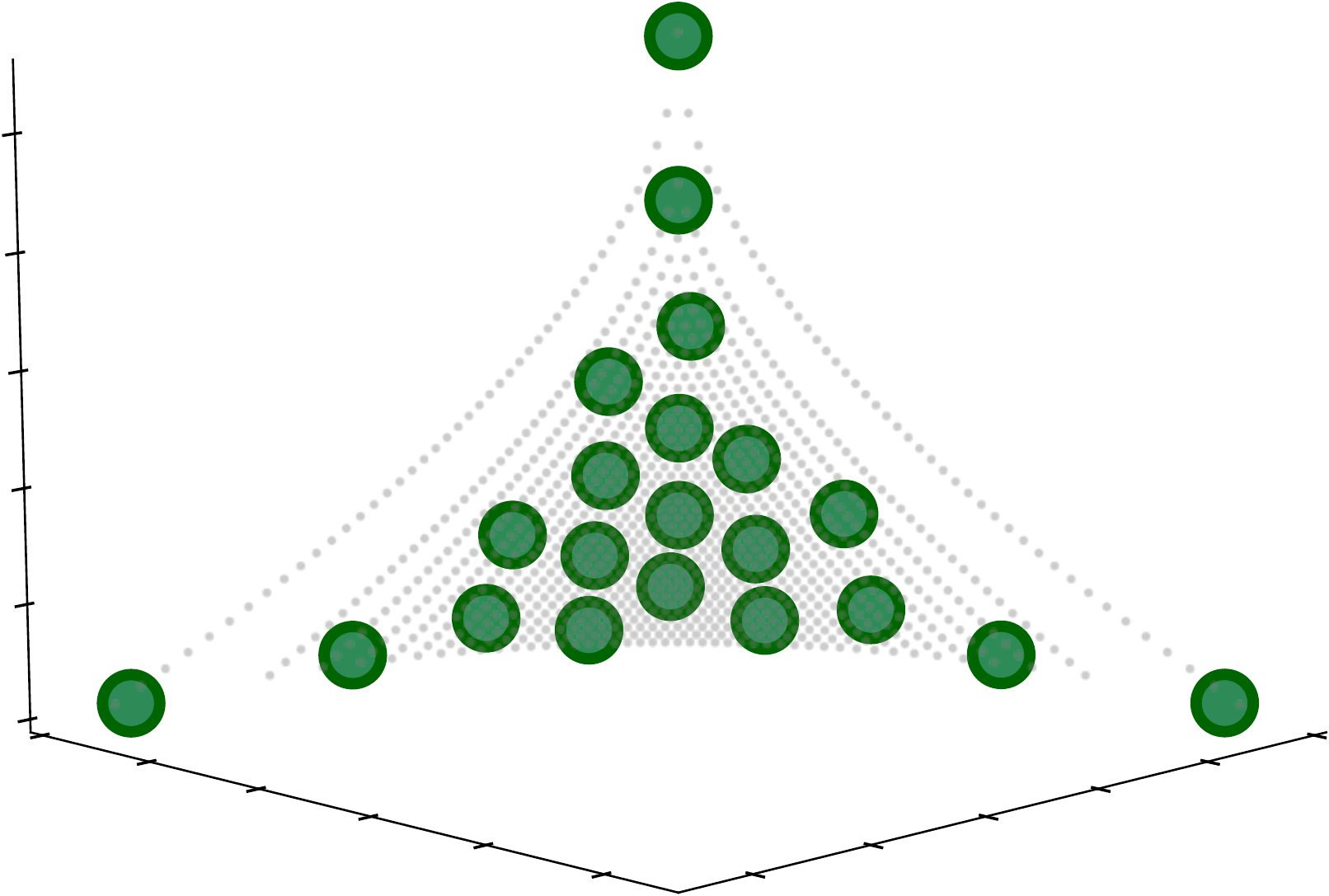}}
\\
\subfloat[$\vector{A}_{\rm NR2}$]{\includegraphics[width=\widthvar\textwidth]{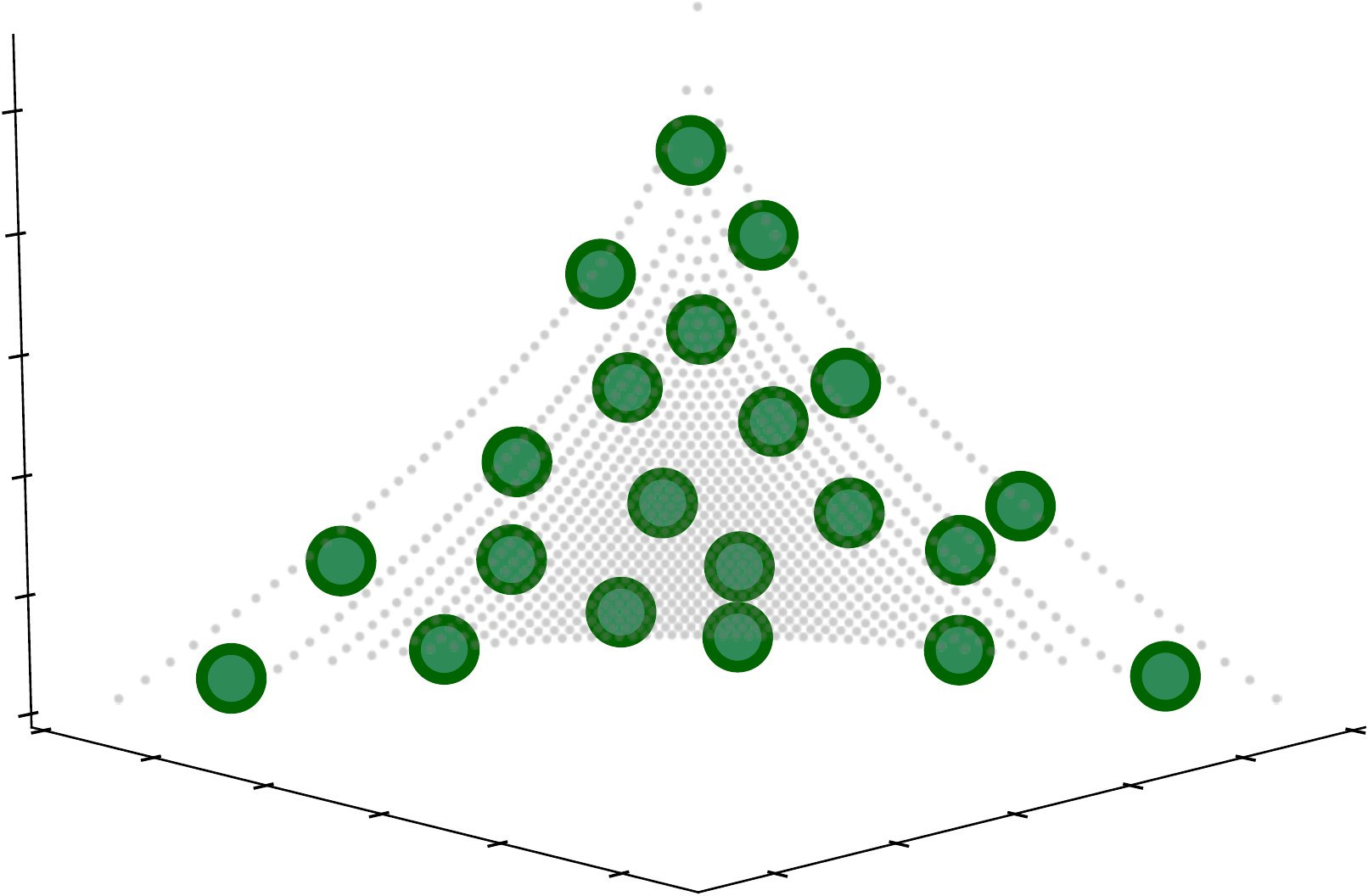}}
\subfloat[$\vector{A}_{I_{\epsilon+}}$]{\includegraphics[width=\widthvar\textwidth]{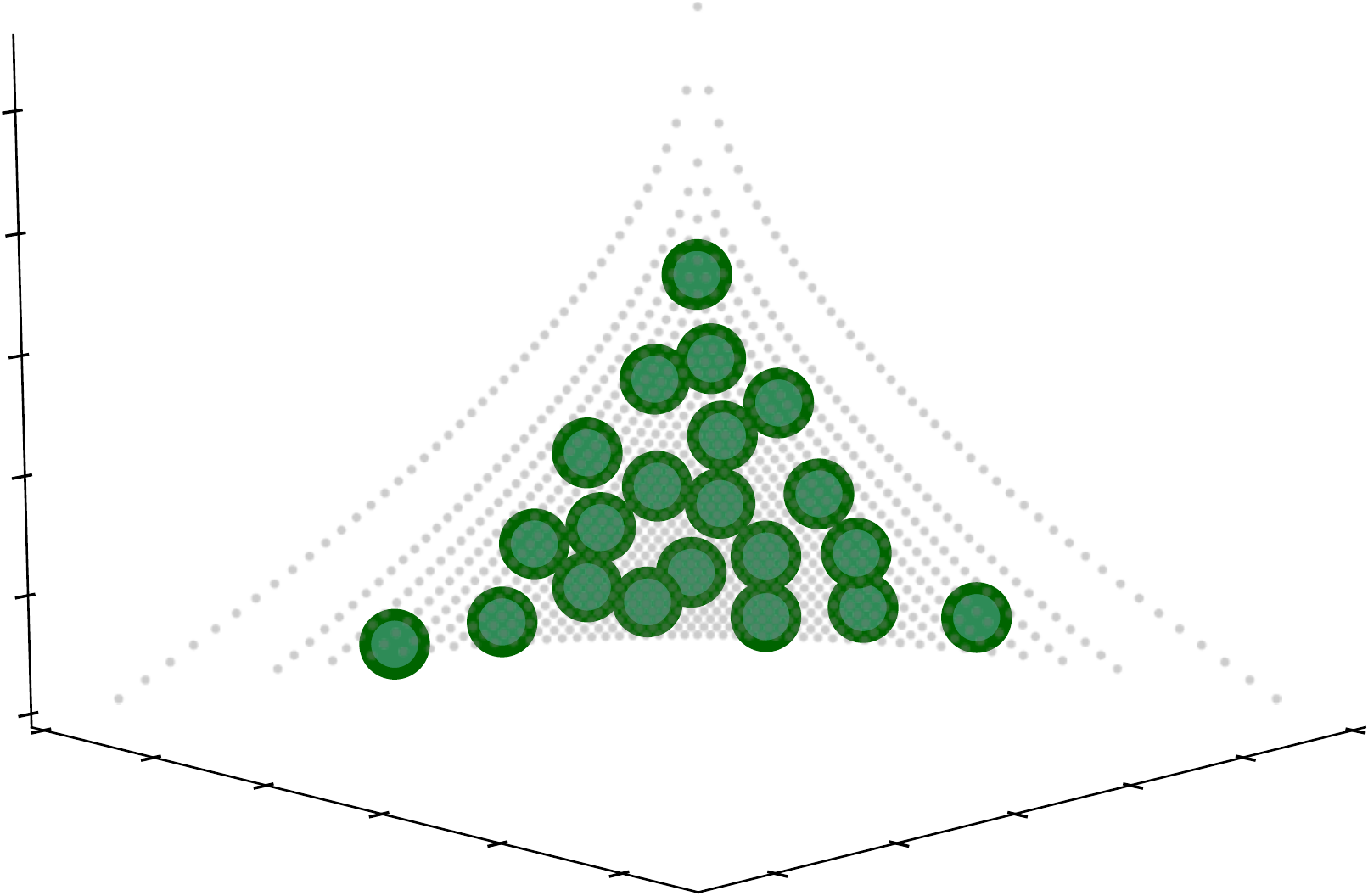}}
\\
\subfloat[$\vector{A}_{\rm SE}$]{\includegraphics[width=\widthvar\textwidth]{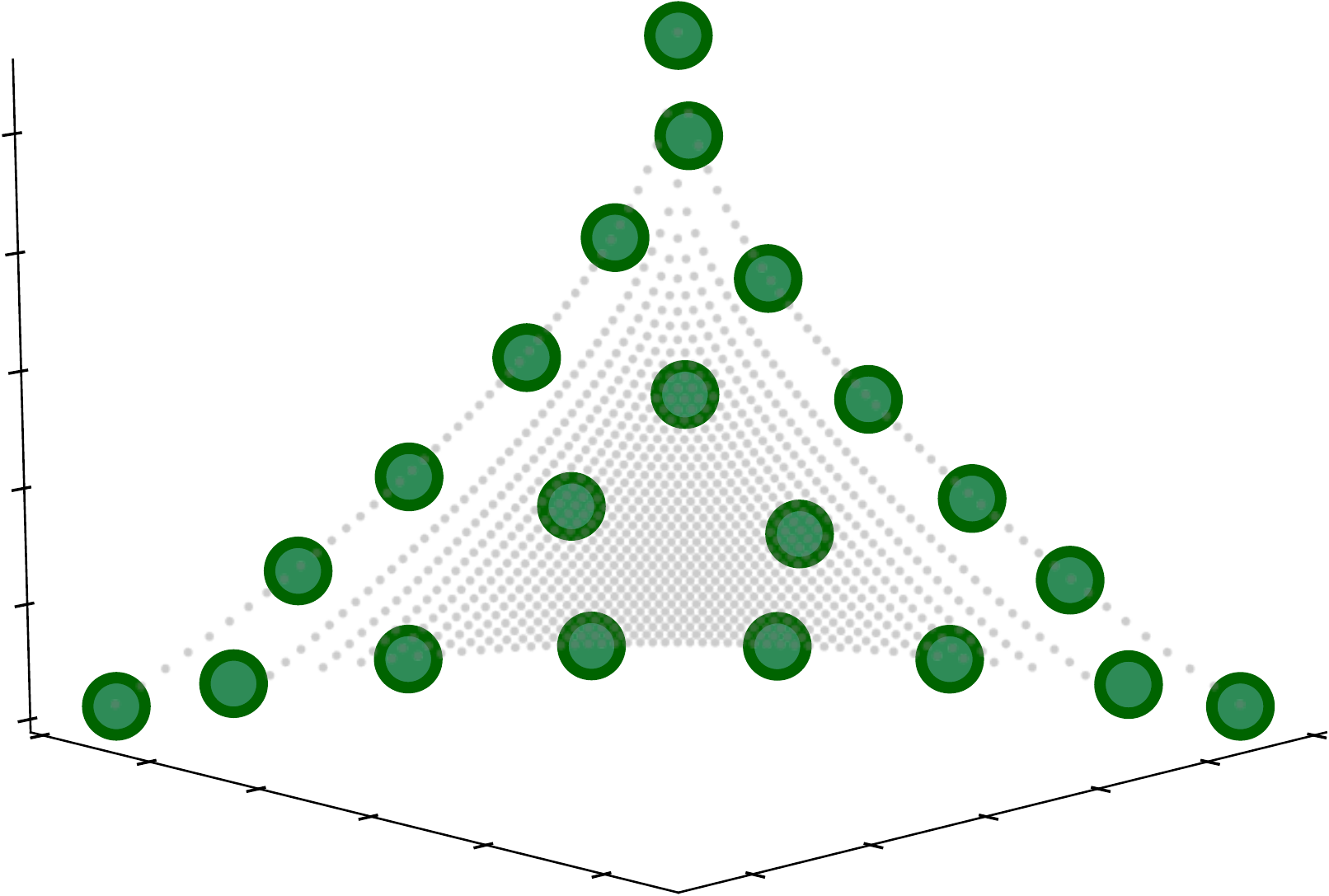}}
\subfloat[$\vector{A}_{\Delta}$]{\includegraphics[width=\widthvar\textwidth]{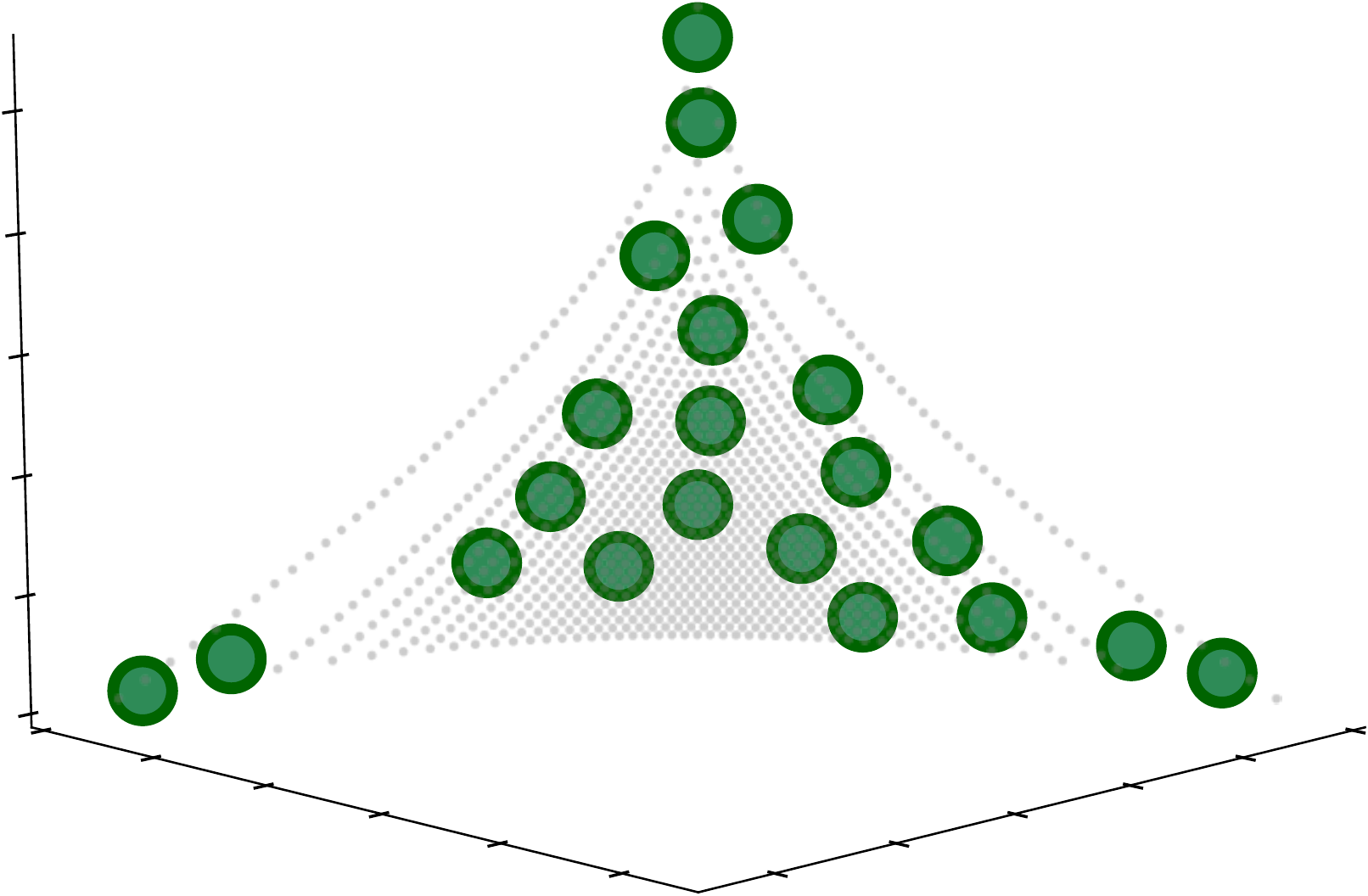}}
\\
\subfloat[$\vector{A}_{\rm PD}$]{\includegraphics[width=\widthvar\textwidth]{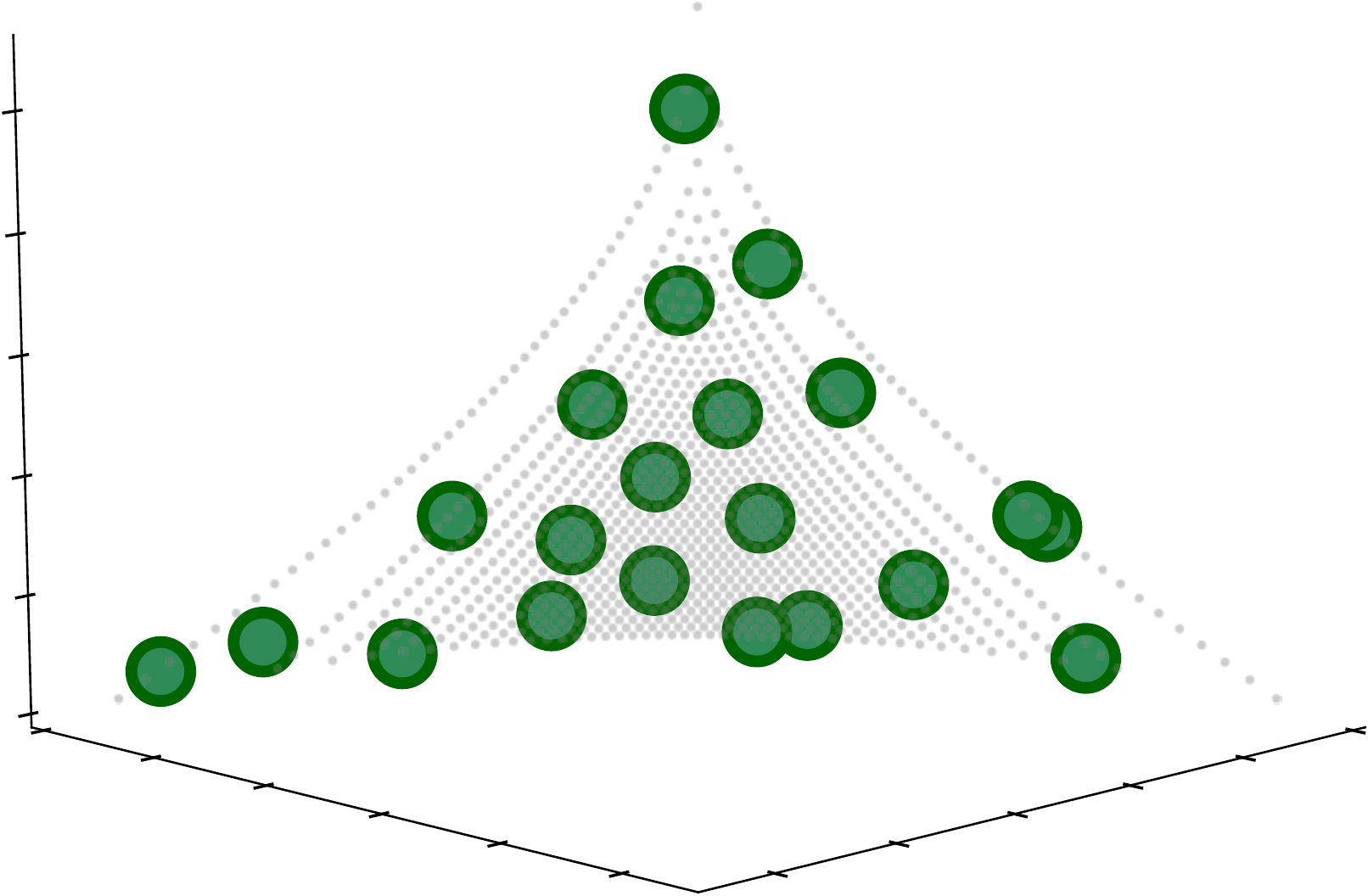}}
\subfloat[$\vector{A}_{\rm SLD}$]{\includegraphics[width=\widthvar\textwidth]{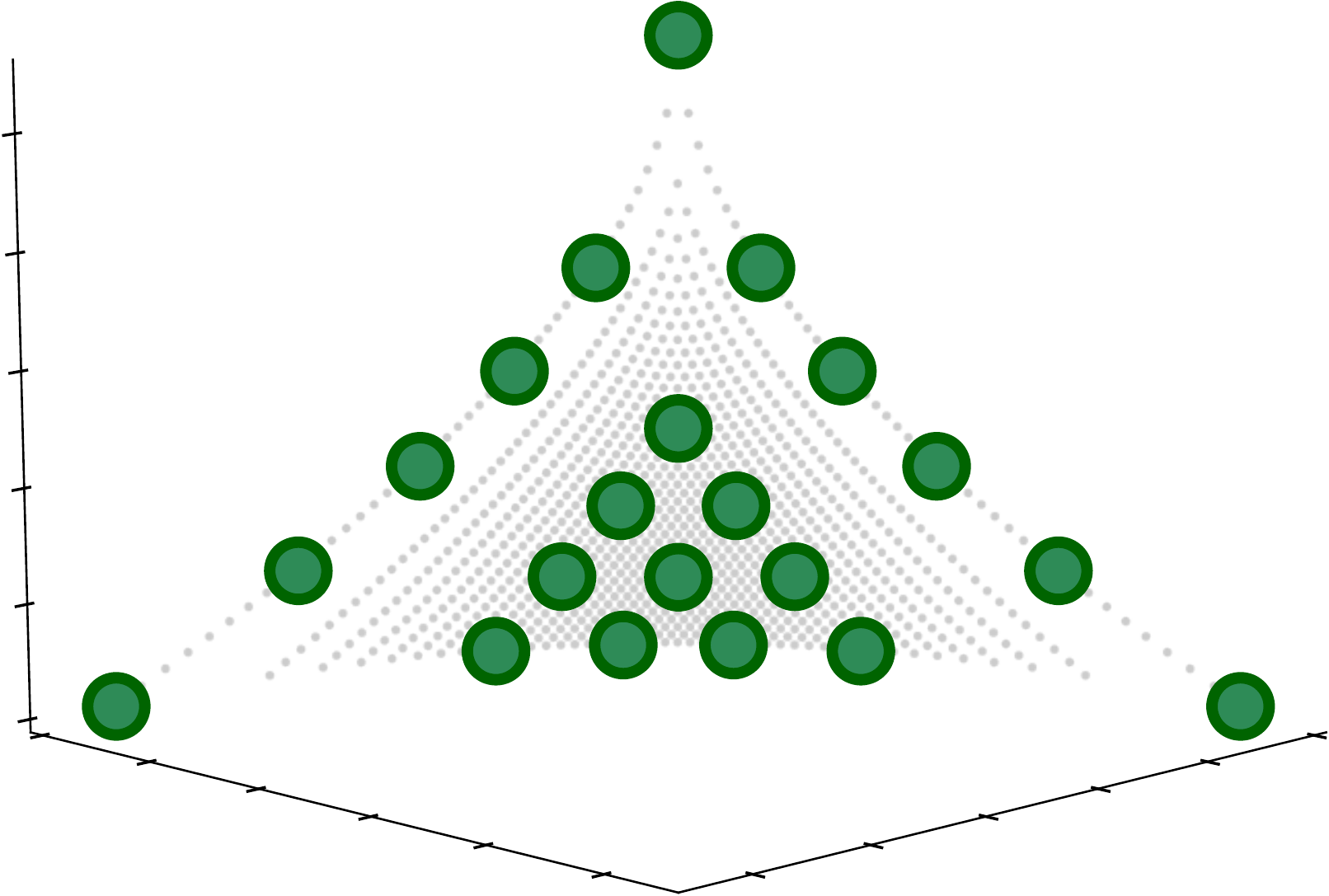}}
\caption{
\small
Results on $F_{\rm convex}$.
}
\label{fig:convex}
 \end{minipage}
\end{tabular}
\end{figure*}

\begin{figure*}[htp]
\begin{tabular}{ccc}
  \newcommand{\widthvar}{0.41}    
 \begin{minipage}[t]{0.33\hsize}
   \centering
\subfloat[$\vector{A}_{\rm HV}$]{\includegraphics[width=\widthvar\textwidth]{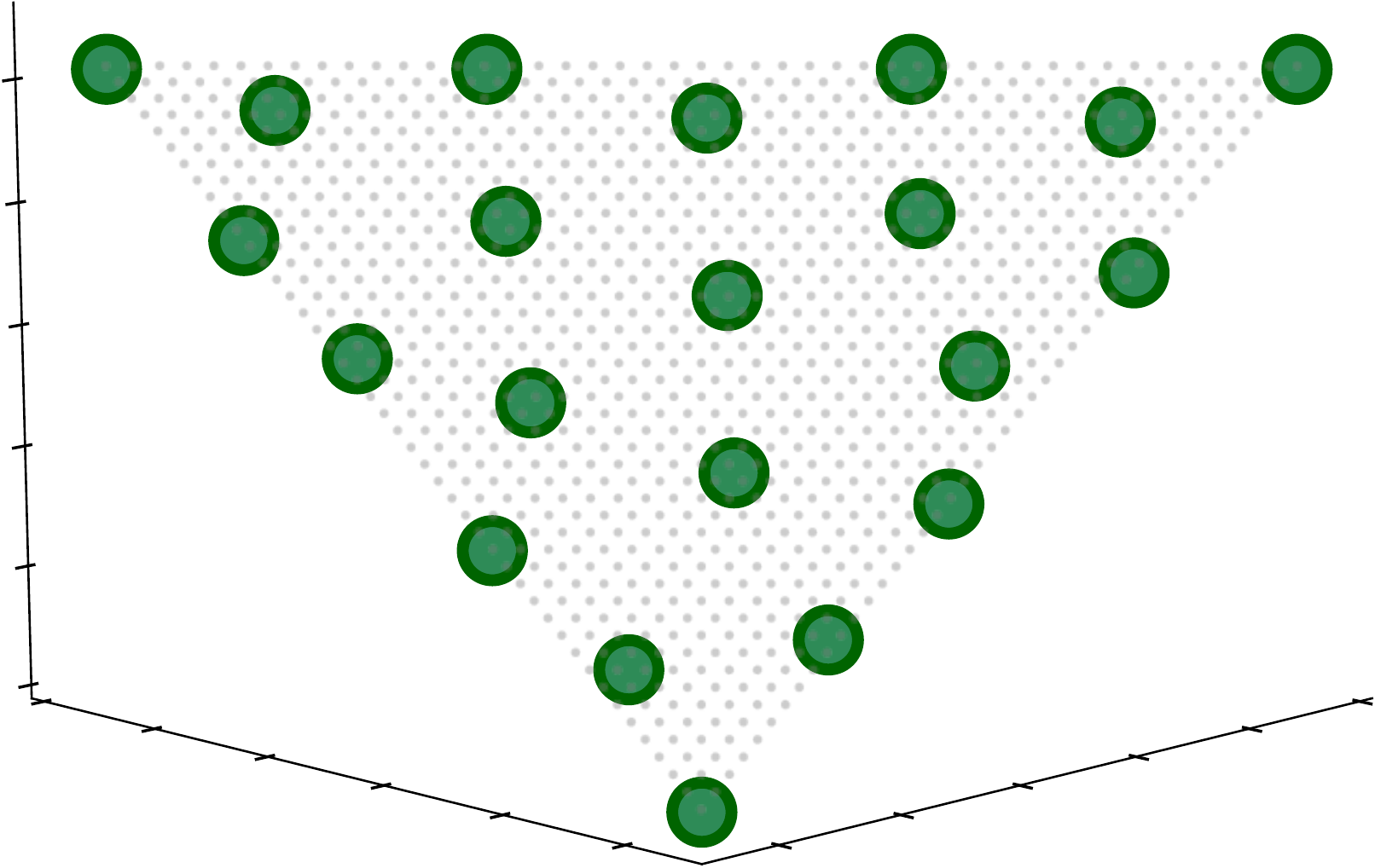}}
\subfloat[$\vector{A}_{\rm IGD}$]{\includegraphics[width=\widthvar\textwidth]{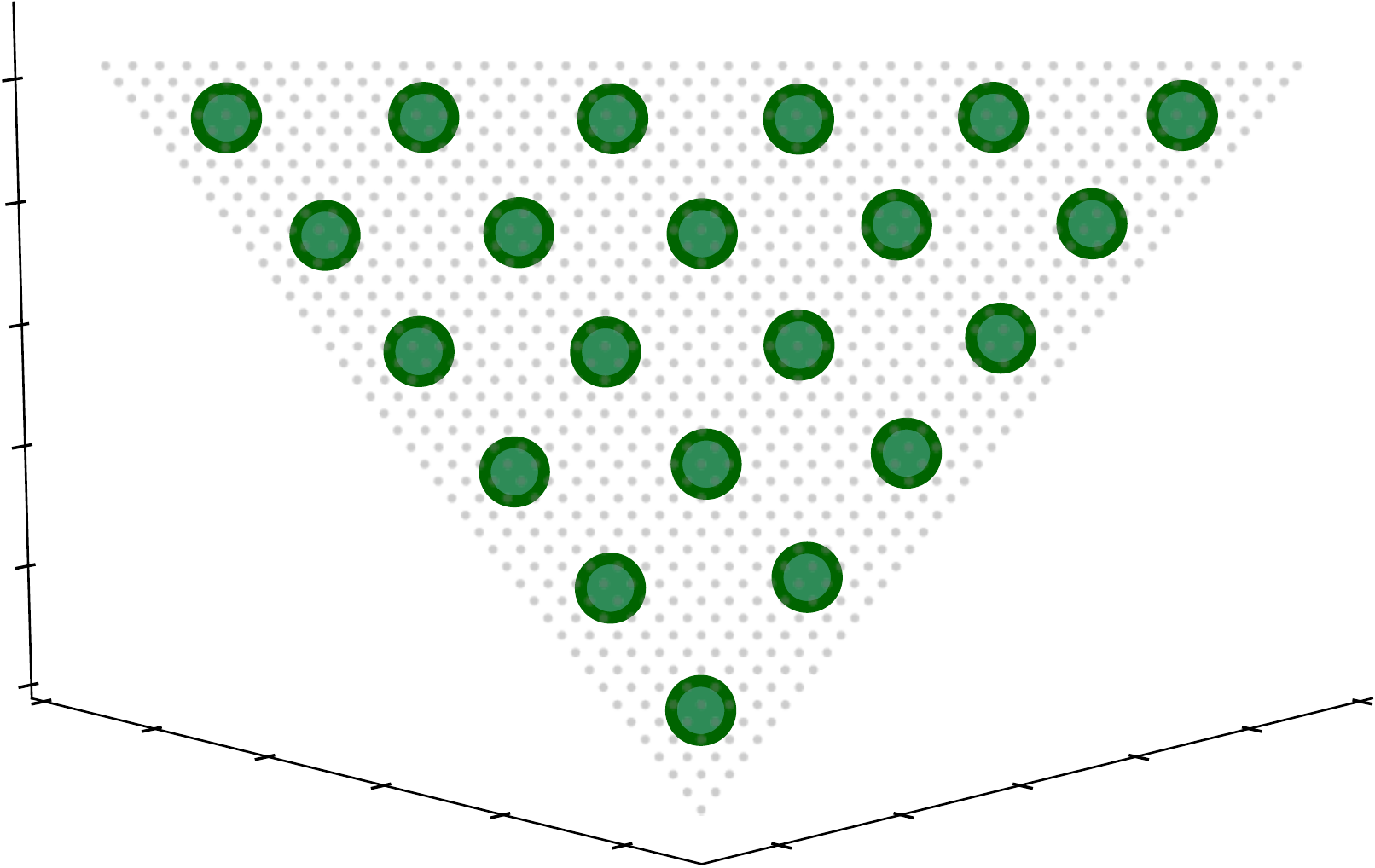}}
\\
\subfloat[$\vector{A}_{\rm IGD^+}$]{\includegraphics[width=\widthvar\textwidth]{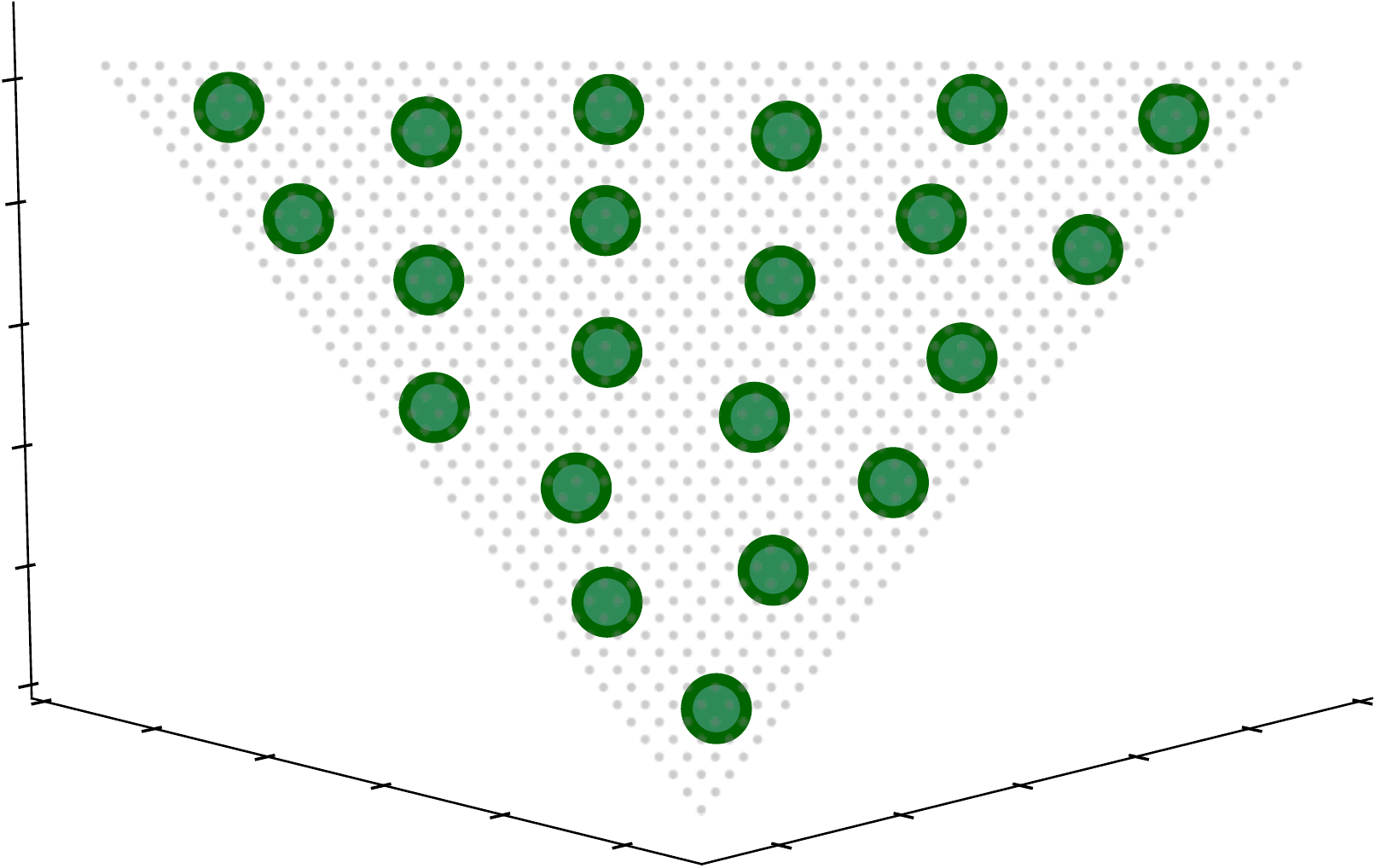}}
\subfloat[$\vector{A}_{\rm R2}$]{\includegraphics[width=\widthvar\textwidth]{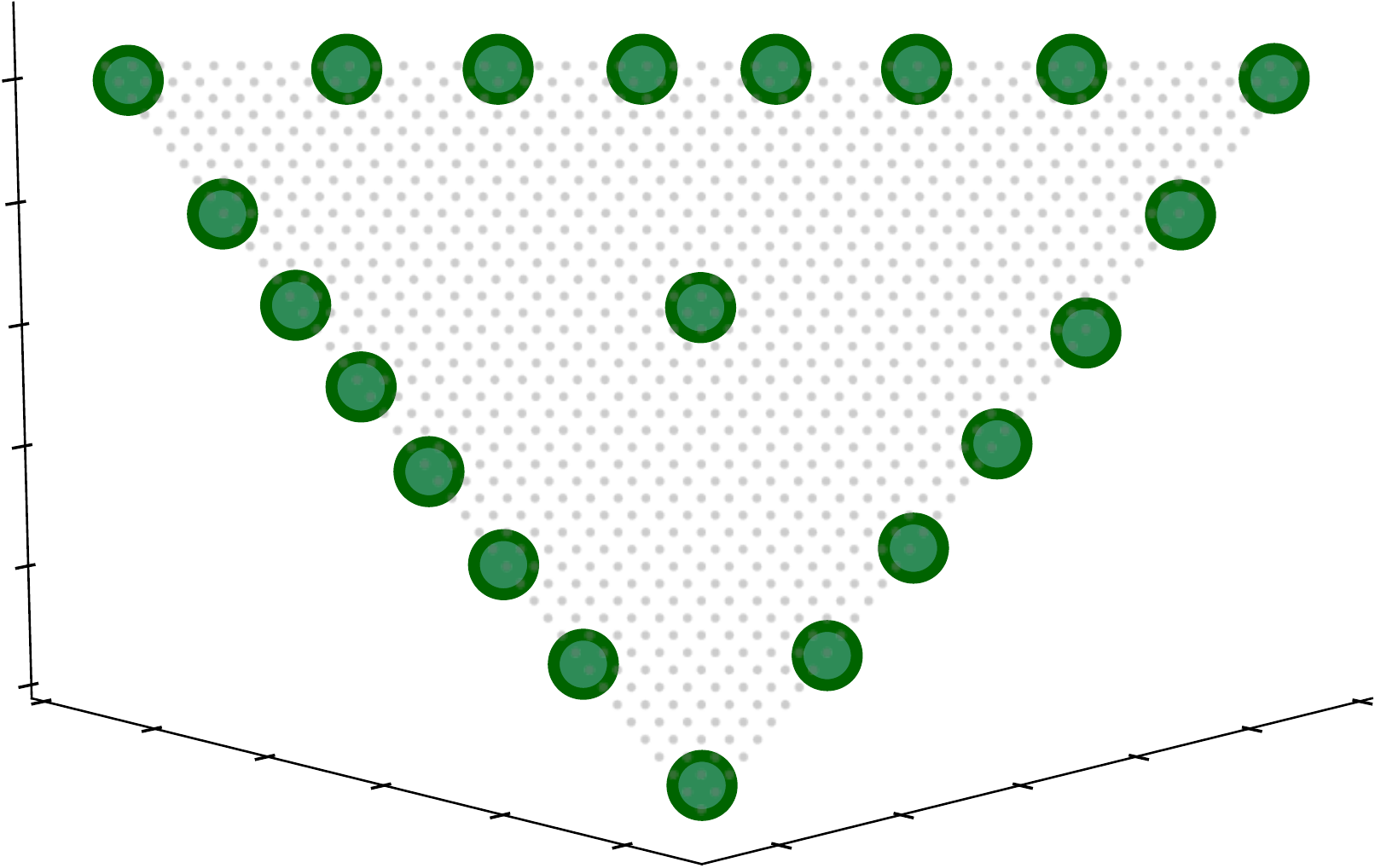}}
\\
\subfloat[$\vector{A}_{\rm NR2}$]{\includegraphics[width=\widthvar\textwidth]{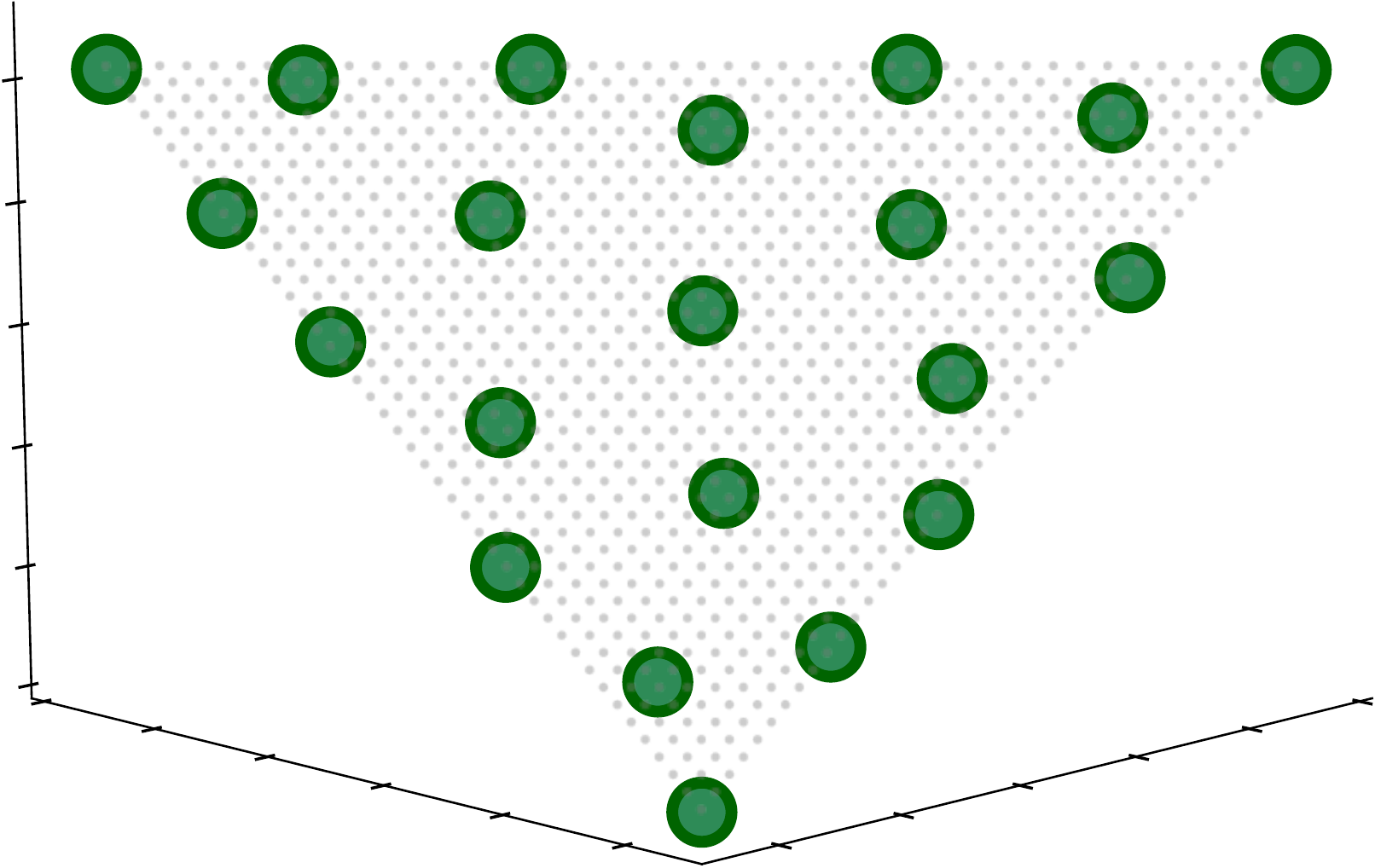}}
\subfloat[$\vector{A}_{I_{\epsilon+}}$]{\includegraphics[width=\widthvar\textwidth]{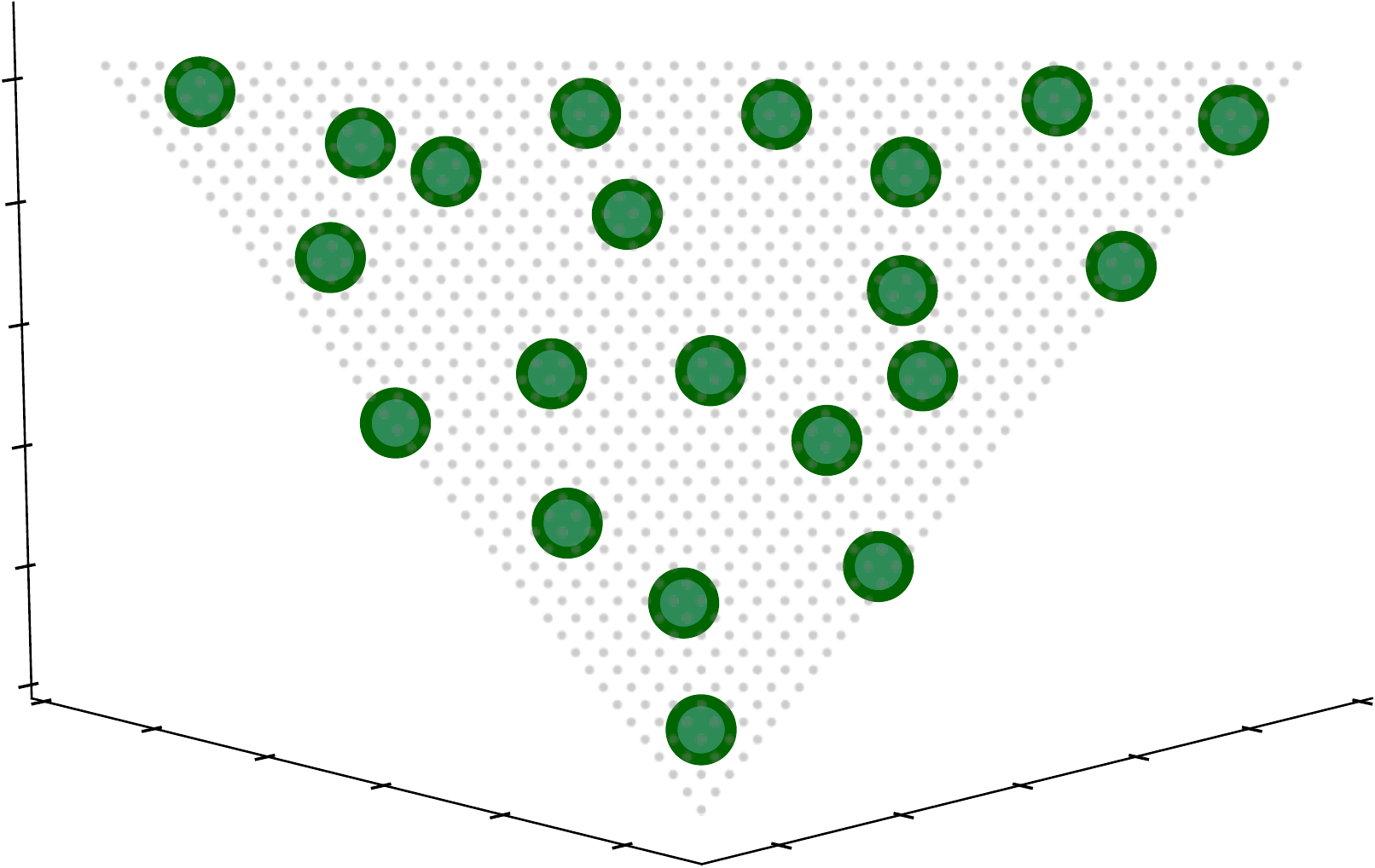}}
\\
\subfloat[$\vector{A}_{\rm SE}$]{\includegraphics[width=\widthvar\textwidth]{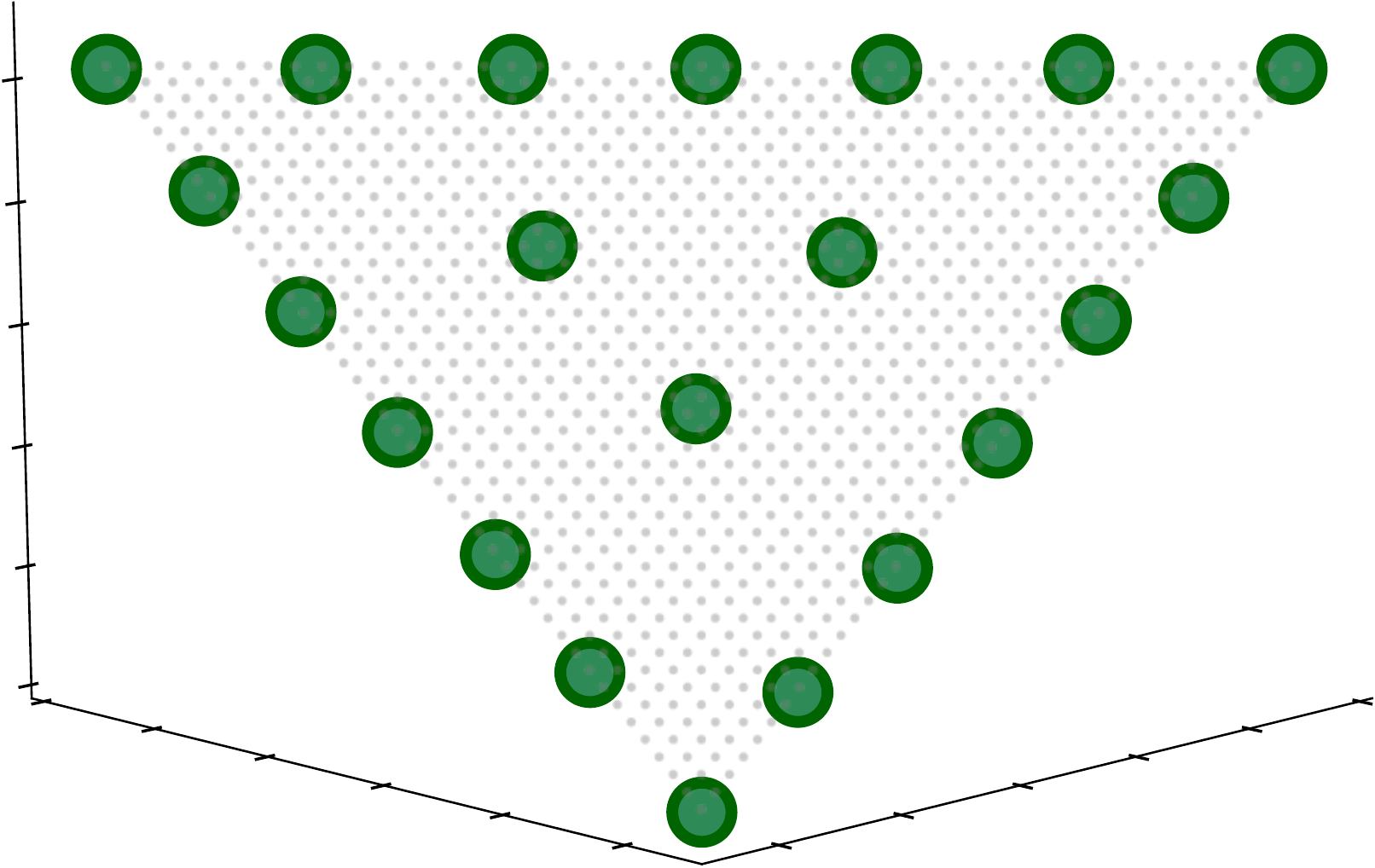}}
\subfloat[$\vector{A}_{\Delta}$]{\includegraphics[width=\widthvar\textwidth]{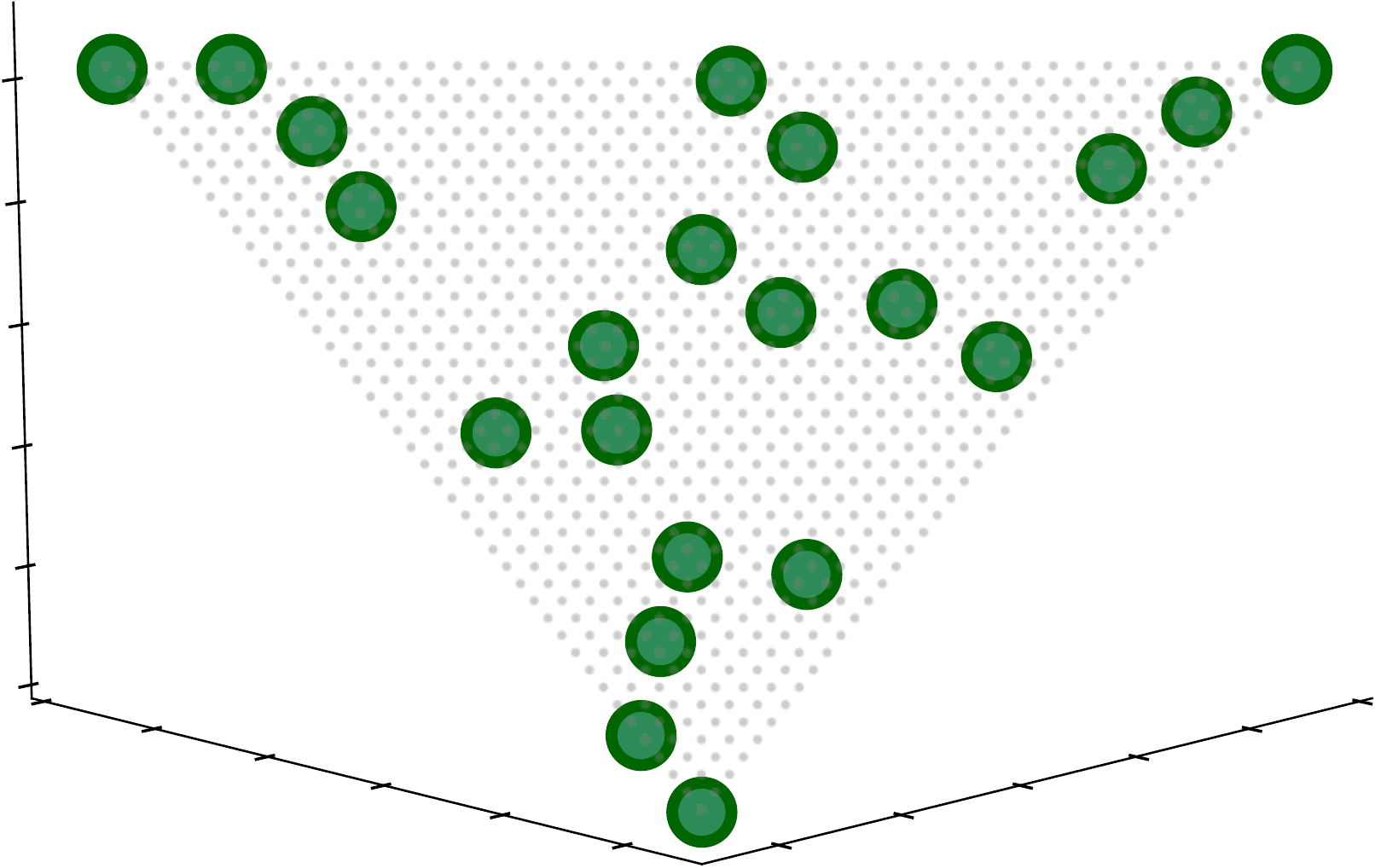}}
\\
\subfloat[$\vector{A}_{\rm PD}$]{\includegraphics[width=\widthvar\textwidth]{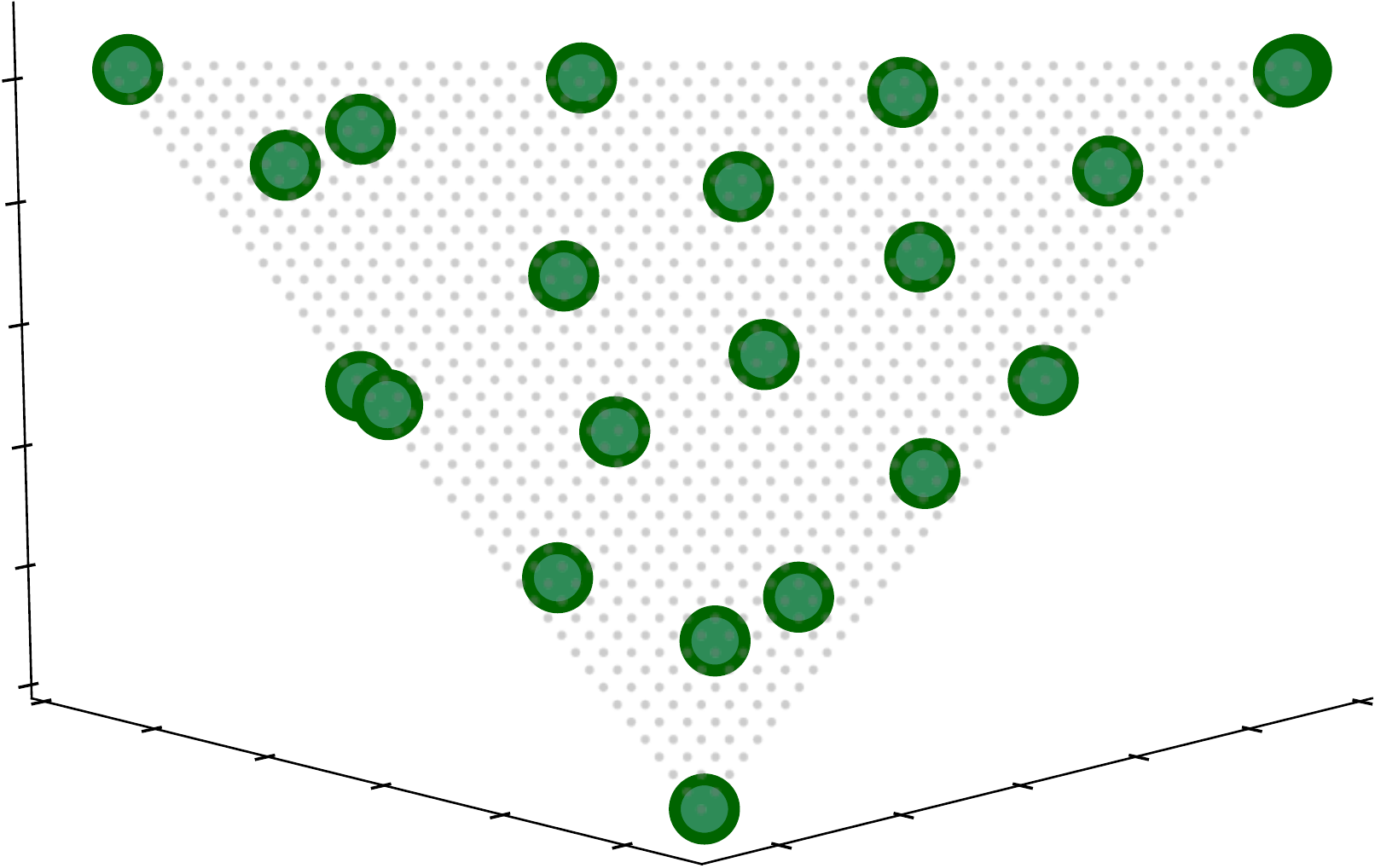}}
\subfloat[$\vector{A}_{\rm SLD}$]{\includegraphics[width=\widthvar\textwidth]{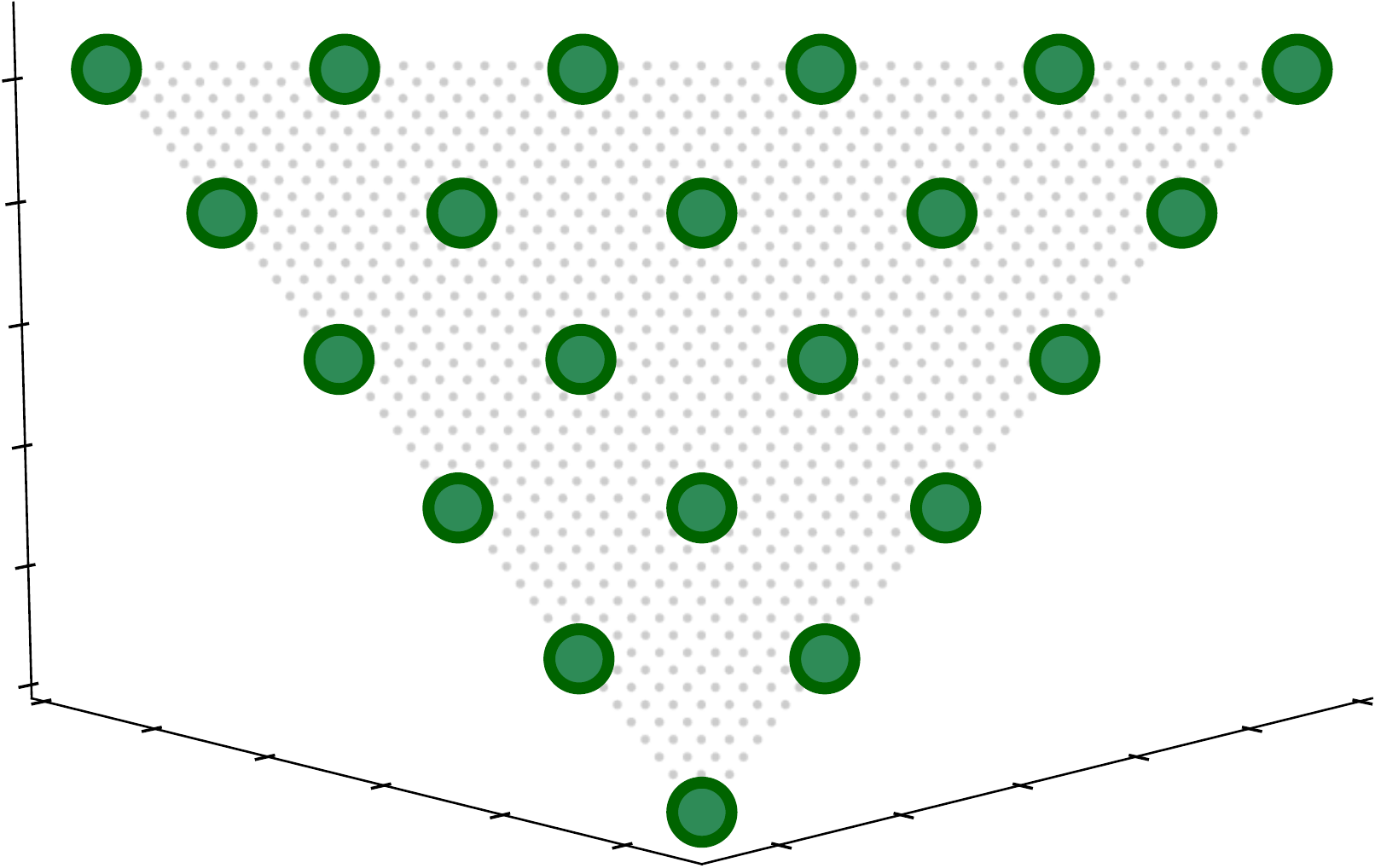}}
\caption{
\small
Results on $F_{\rm i\shyp linear}$.
}
\label{fig:ilinear}
 \end{minipage}

 \begin{minipage}[t]{0.33\hsize}
   \centering
\subfloat[$\vector{A}_{\rm HV}$]{\includegraphics[width=\widthvar\textwidth]{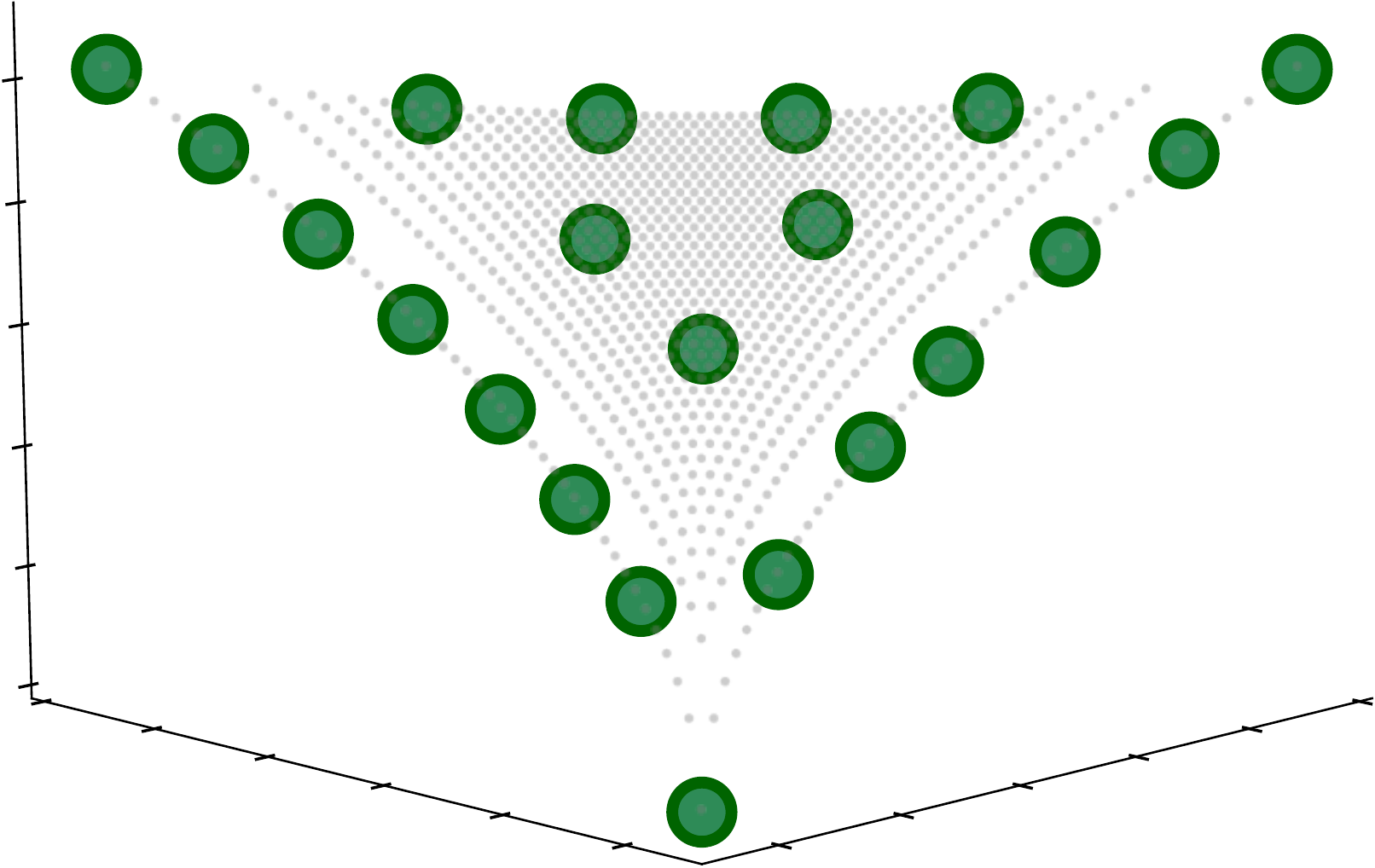}}
\subfloat[$\vector{A}_{\rm IGD}$]{\includegraphics[width=\widthvar\textwidth]{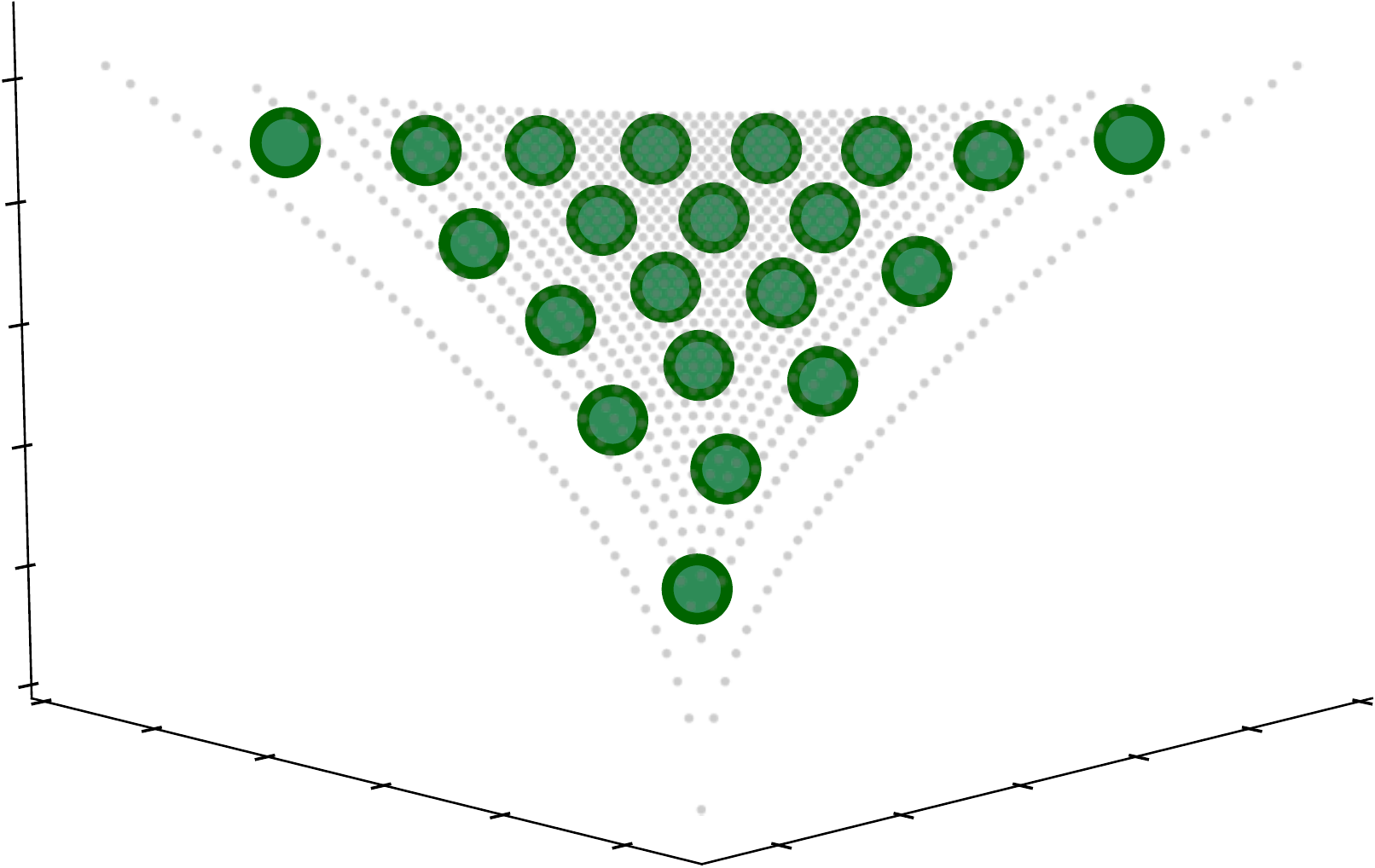}}
\\
\subfloat[$\vector{A}_{\rm IGD^+}$]{\includegraphics[width=\widthvar\textwidth]{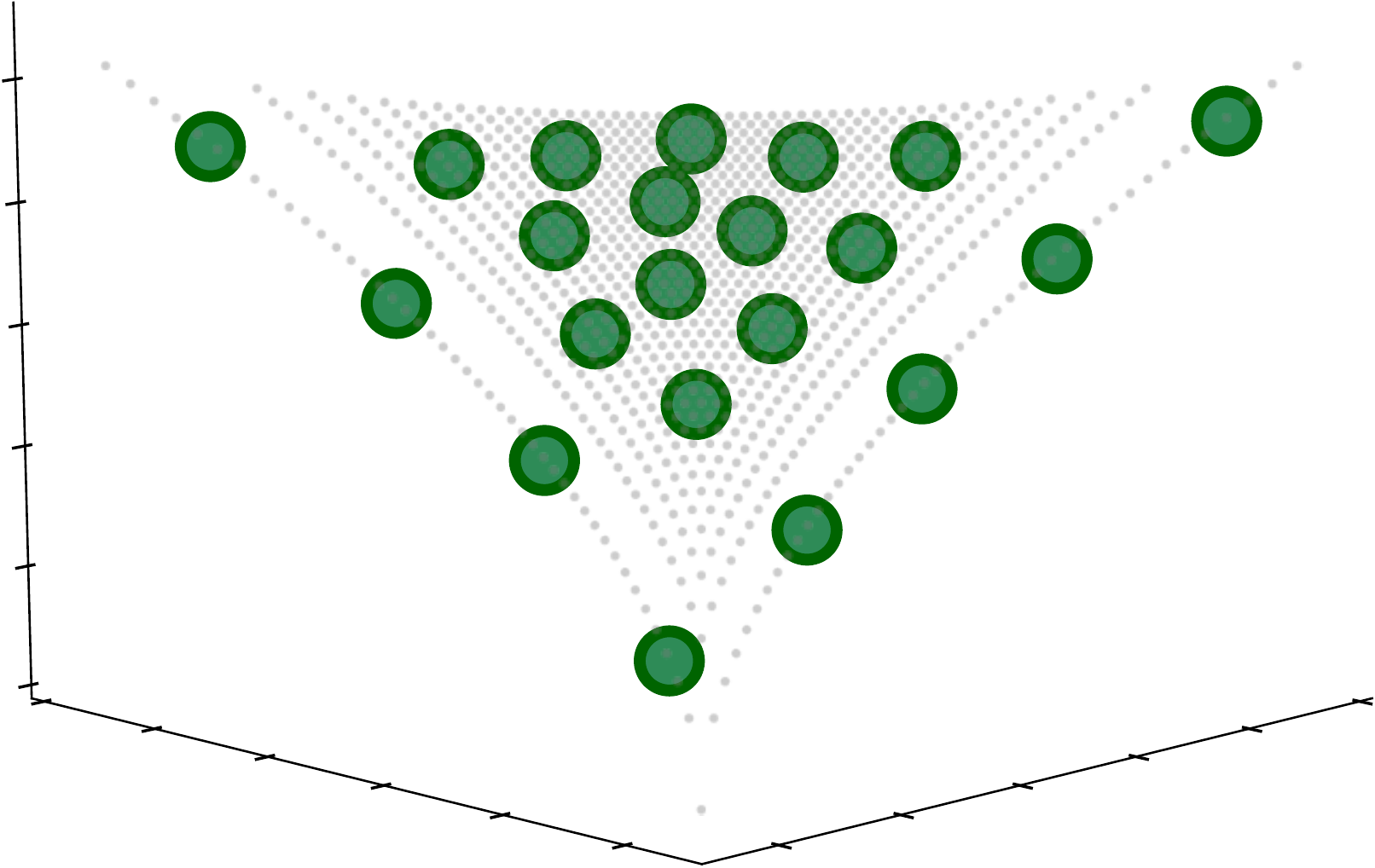}}
\subfloat[$\vector{A}_{\rm R2}$]{\includegraphics[width=\widthvar\textwidth]{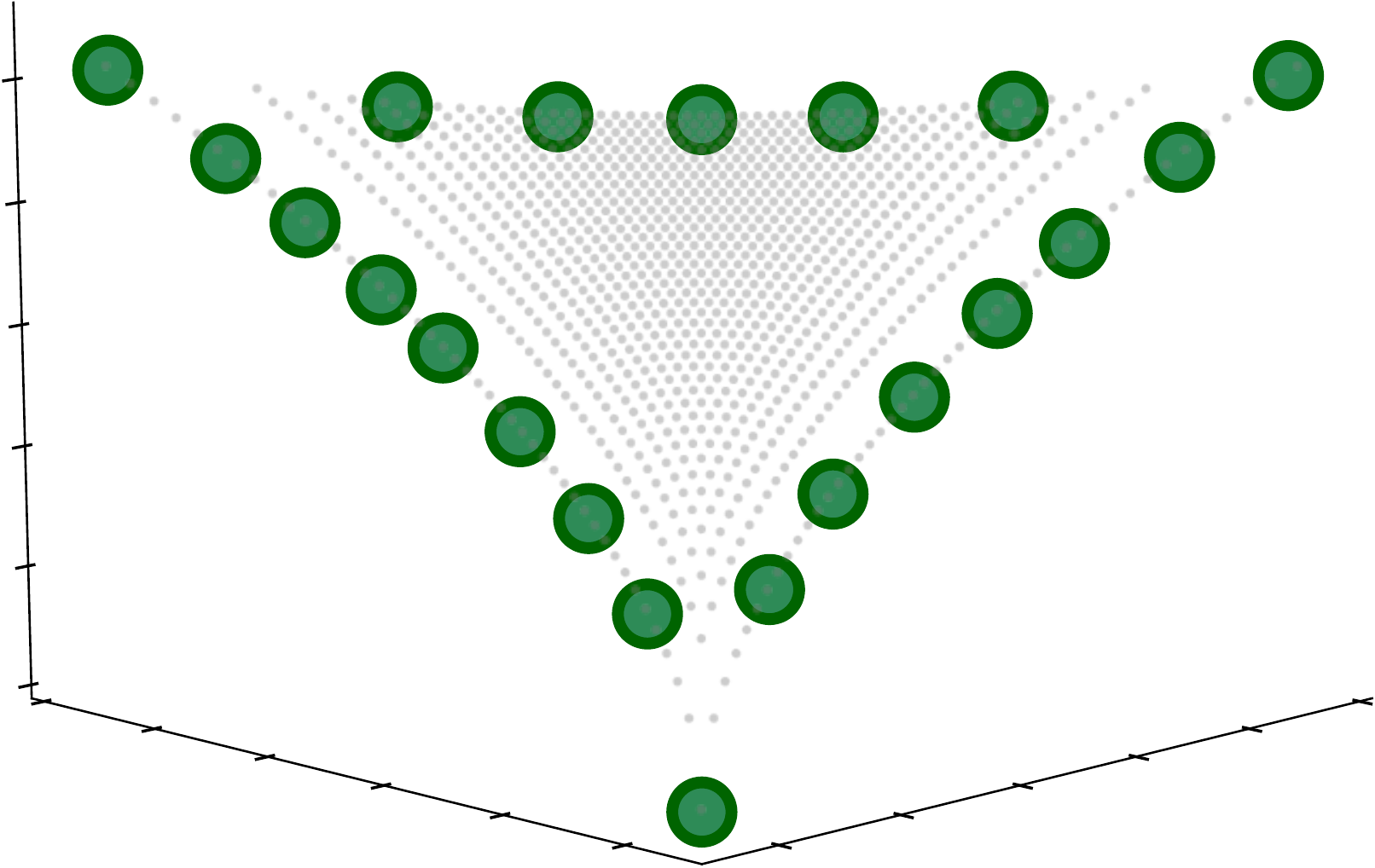}}
\\
\subfloat[$\vector{A}_{\rm NR2}$]{\includegraphics[width=\widthvar\textwidth]{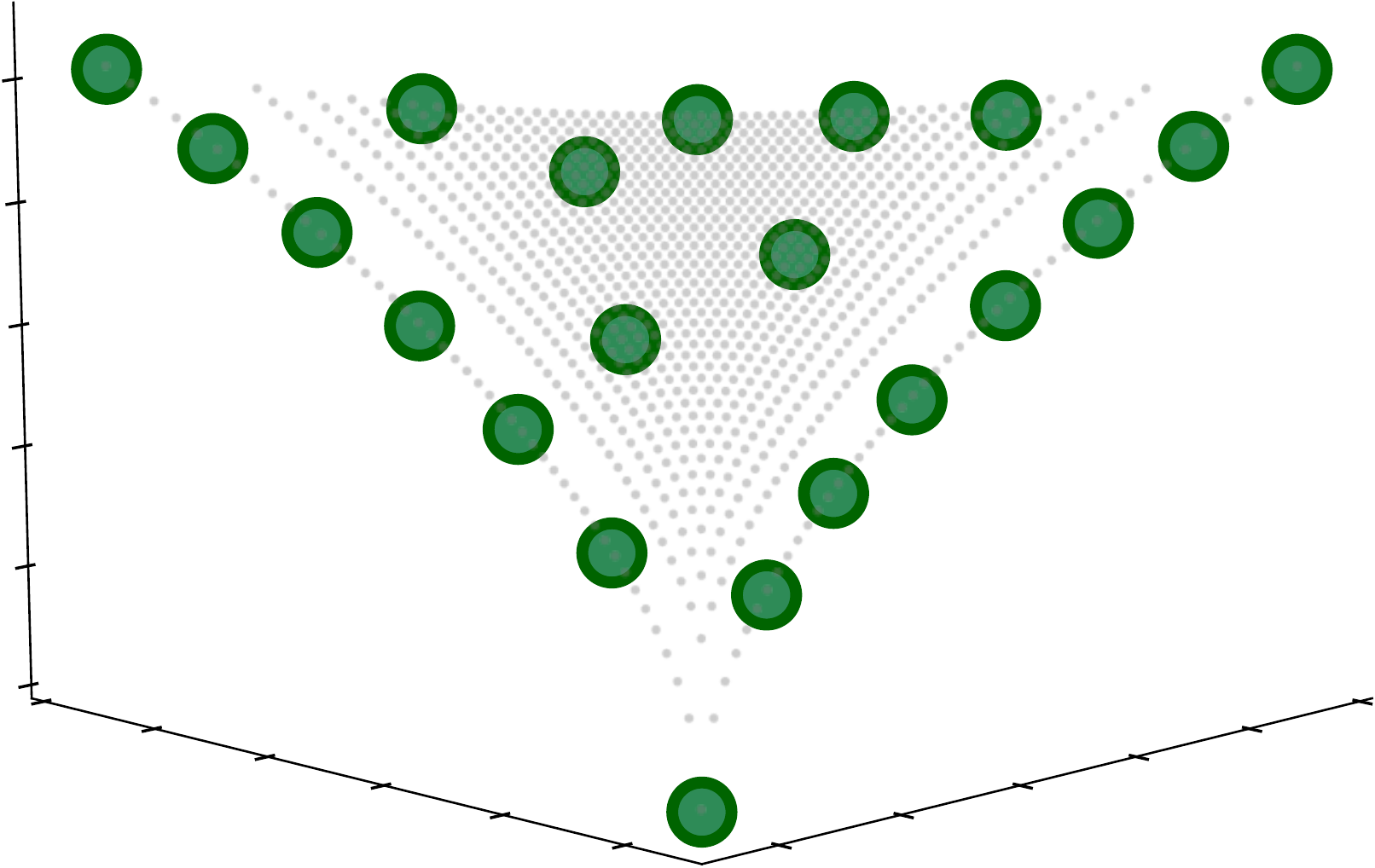}}
\subfloat[$\vector{A}_{I_{\epsilon+}}$]{\includegraphics[width=\widthvar\textwidth]{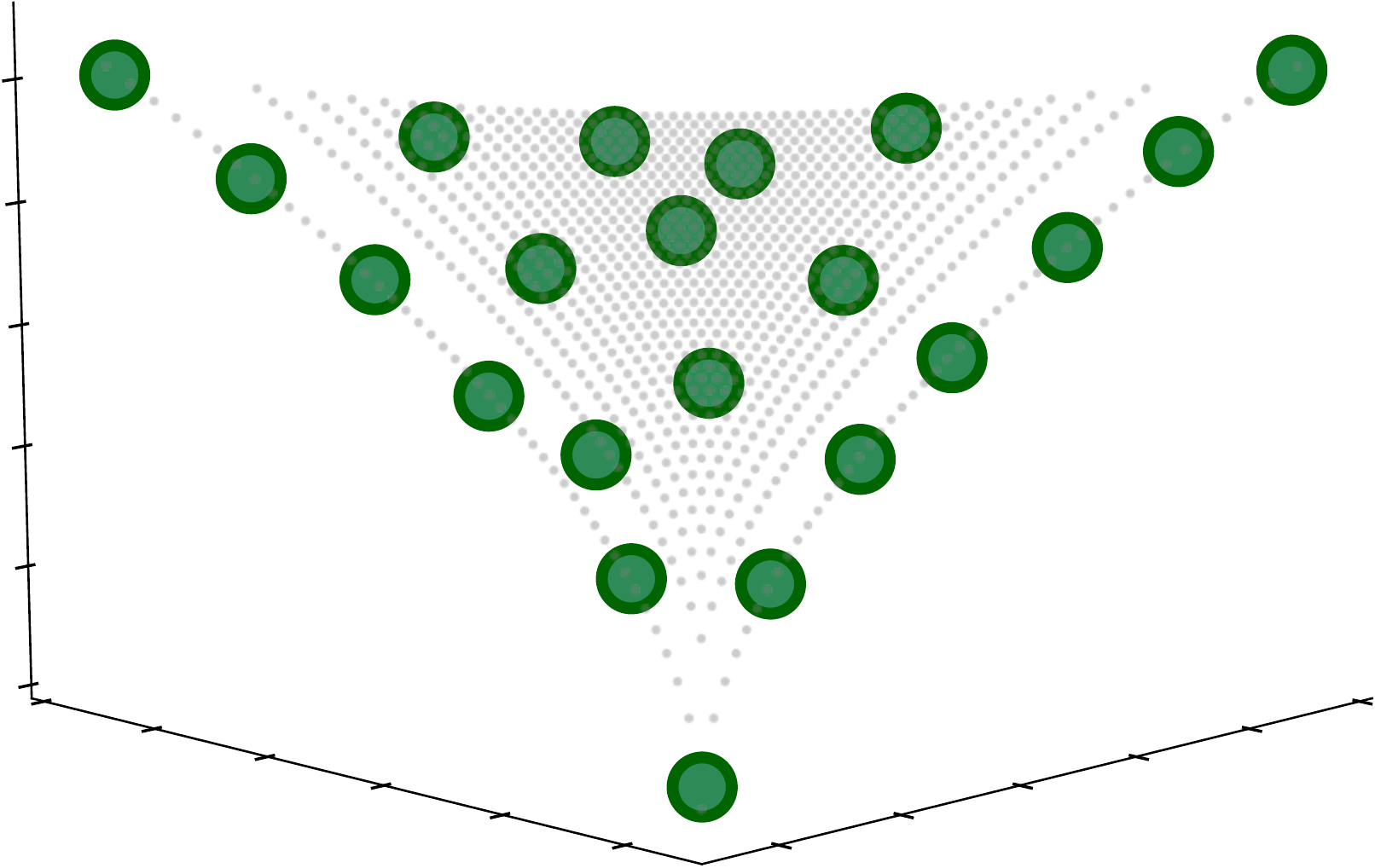}}
\\
\subfloat[$\vector{A}_{\rm SE}$]{\includegraphics[width=\widthvar\textwidth]{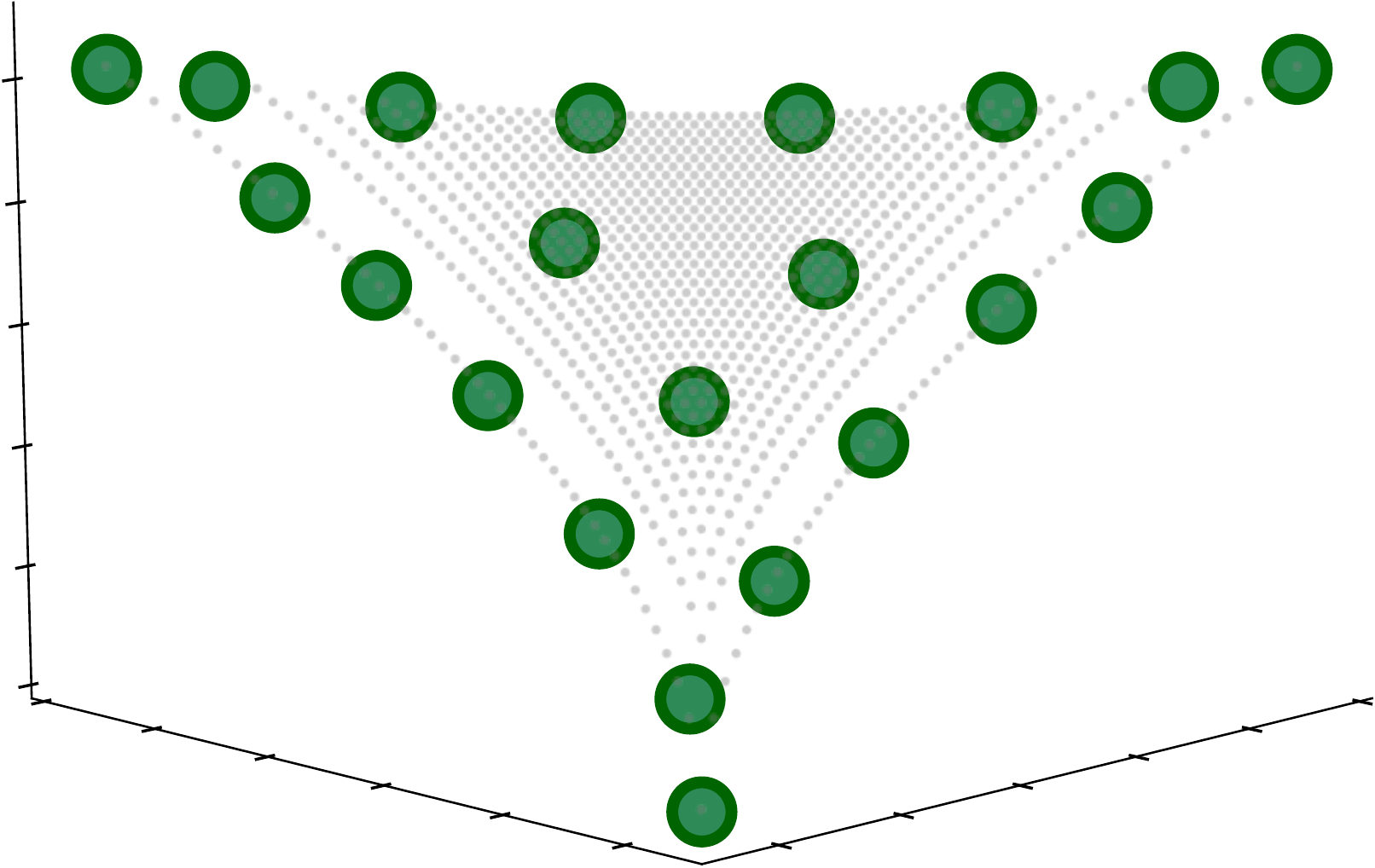}}
\subfloat[$\vector{A}_{\Delta}$]{\includegraphics[width=\widthvar\textwidth]{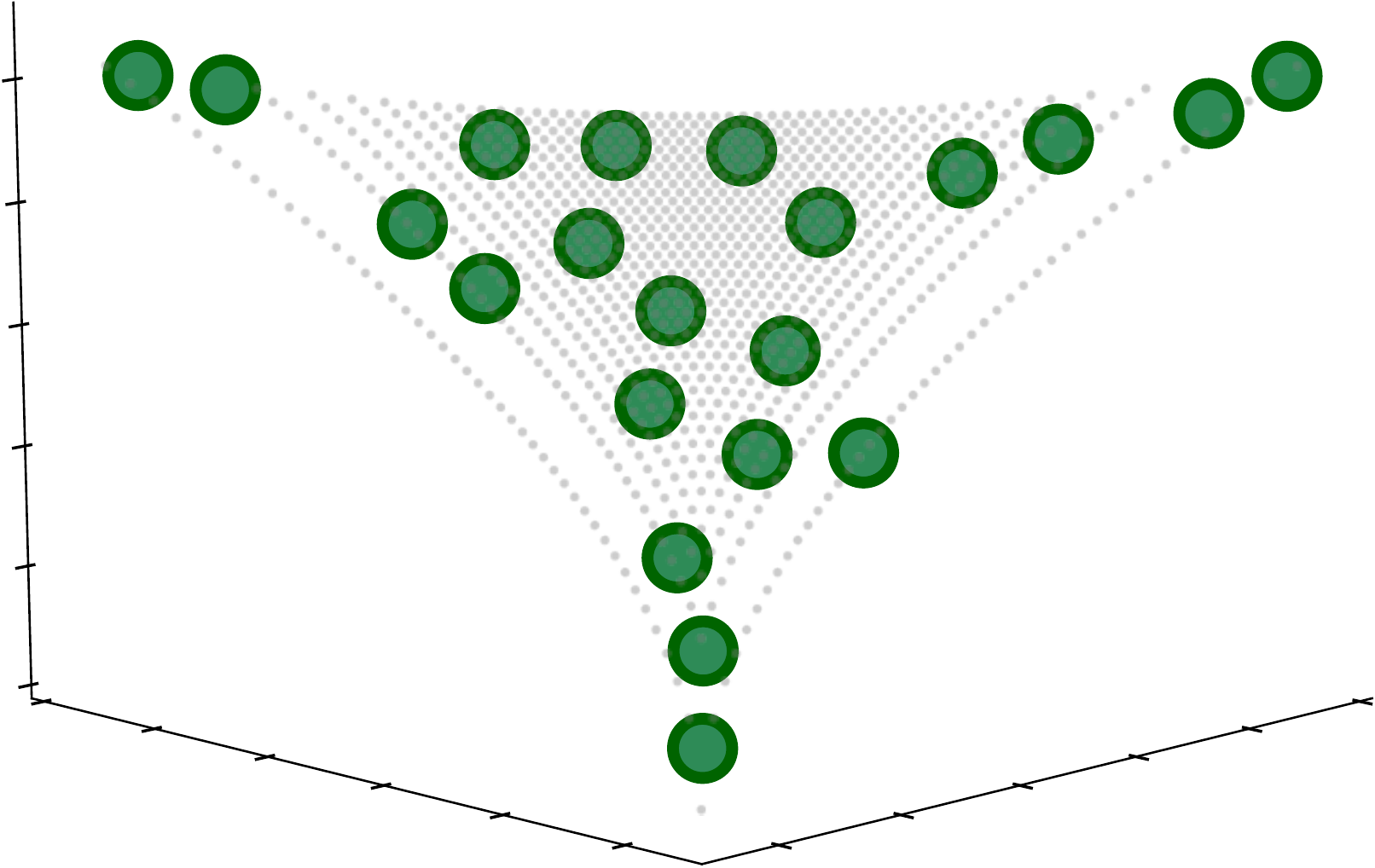}}
\\
\subfloat[$\vector{A}_{\rm PD}$]{\includegraphics[width=\widthvar\textwidth]{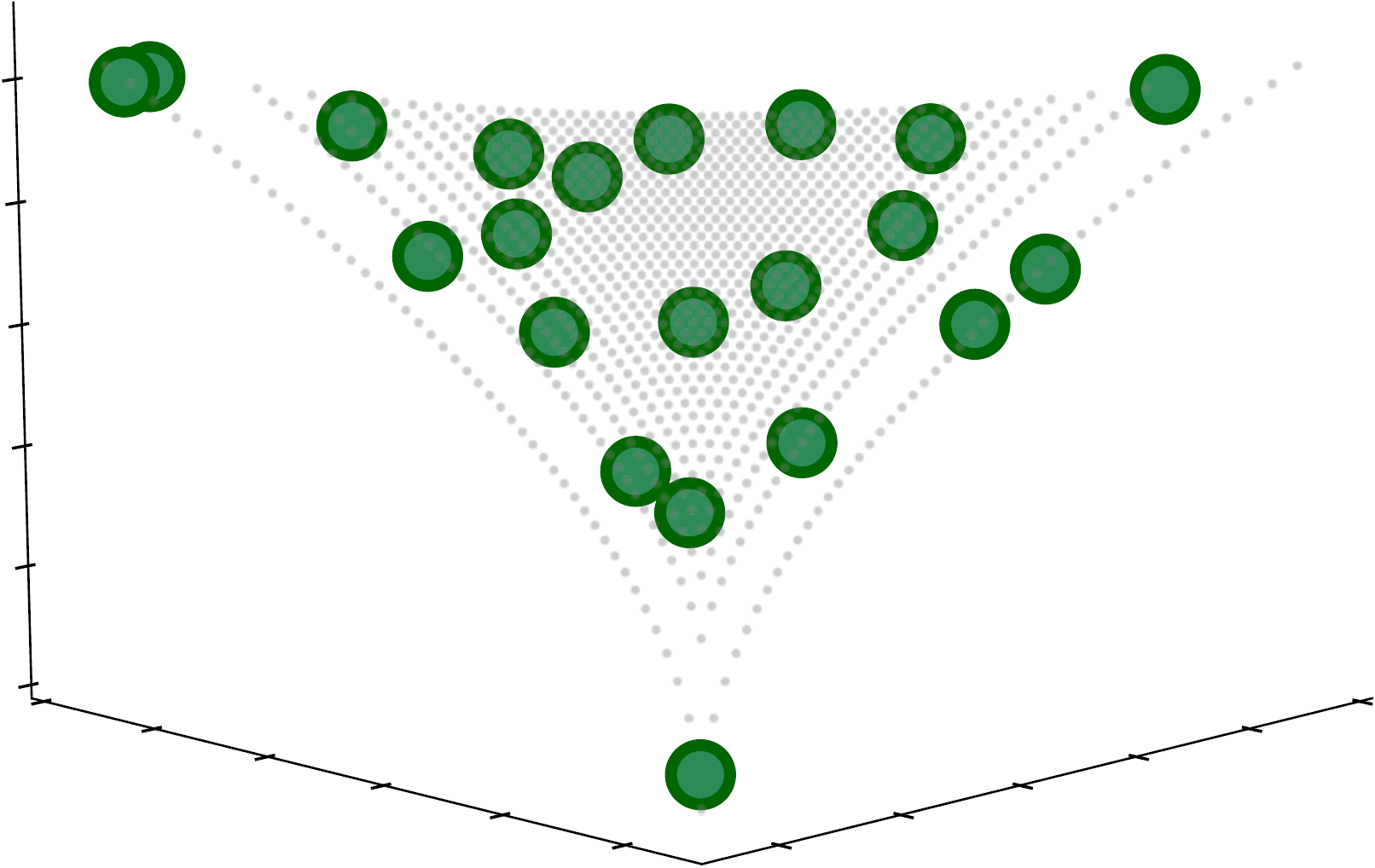}}
\subfloat[$\vector{A}_{\rm SLD}$]{\includegraphics[width=\widthvar\textwidth]{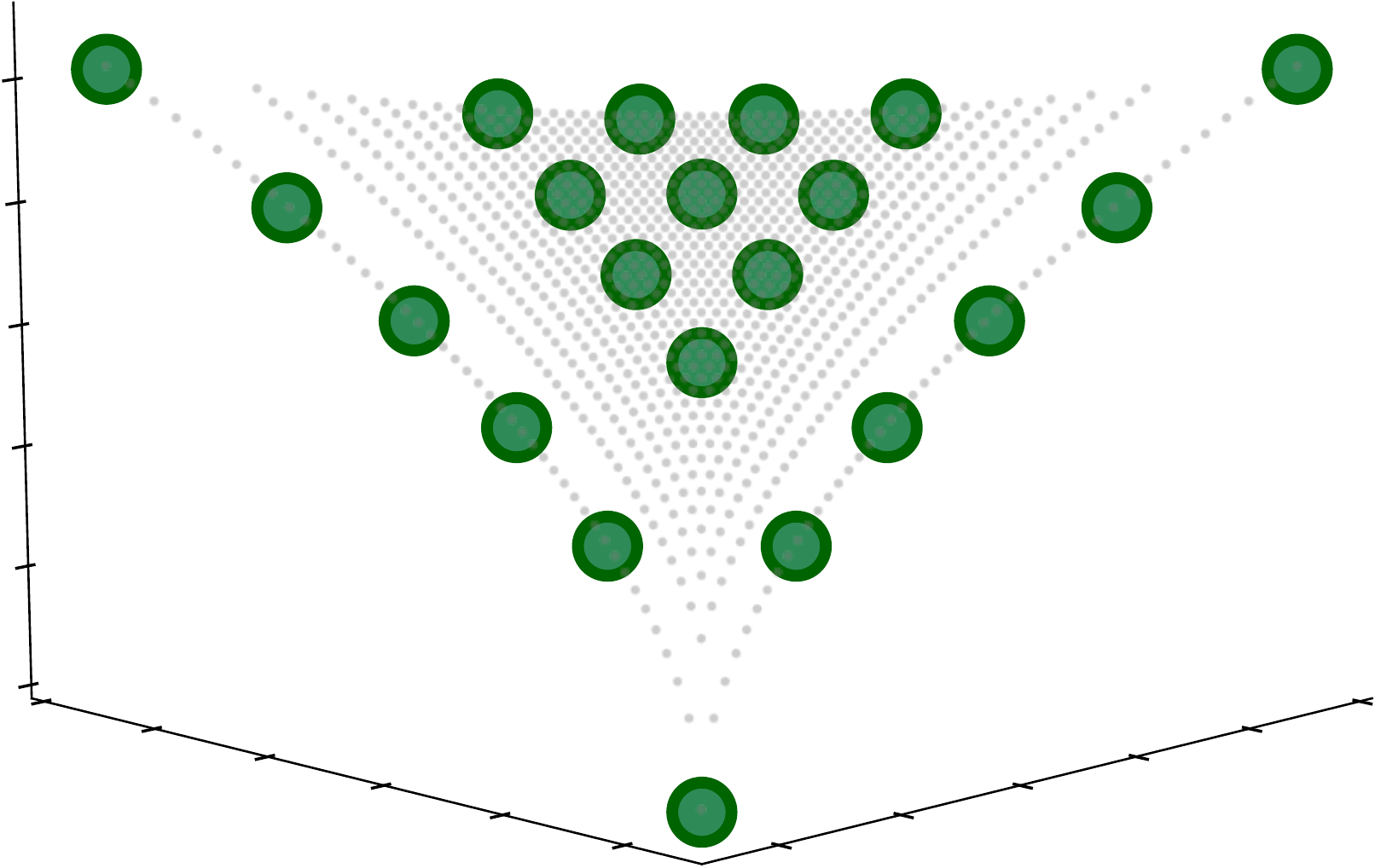}}
\caption{
\small
Results on $F_{\rm i\shyp concave}$.
}
\label{fig:iconcave}
 \end{minipage}
  \begin{minipage}[t]{0.33\hsize}
   \centering
\subfloat[$\vector{A}_{\rm HV}$]{\includegraphics[width=\widthvar\textwidth]{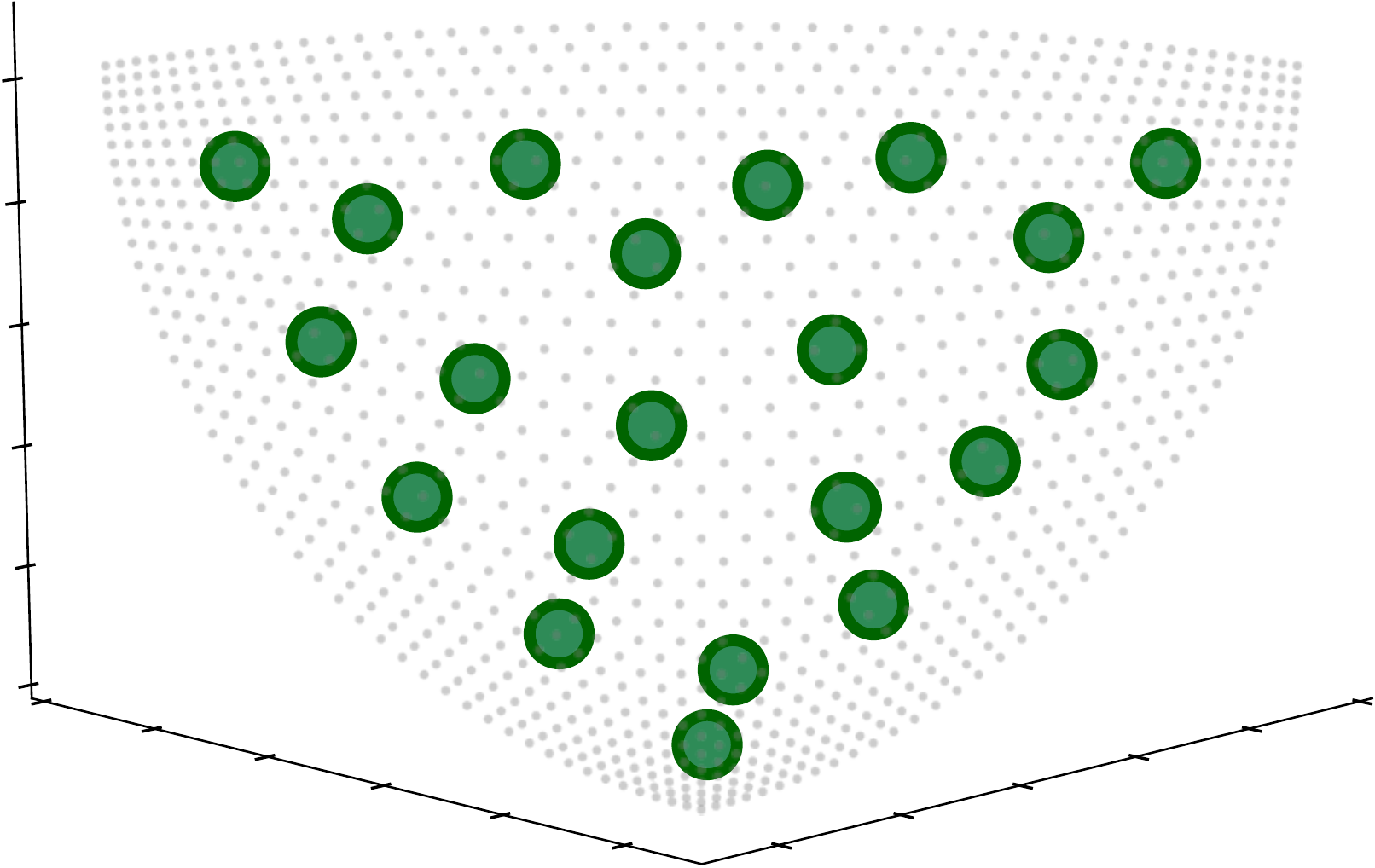}}
\subfloat[$\vector{A}_{\rm IGD}$]{\includegraphics[width=\widthvar\textwidth]{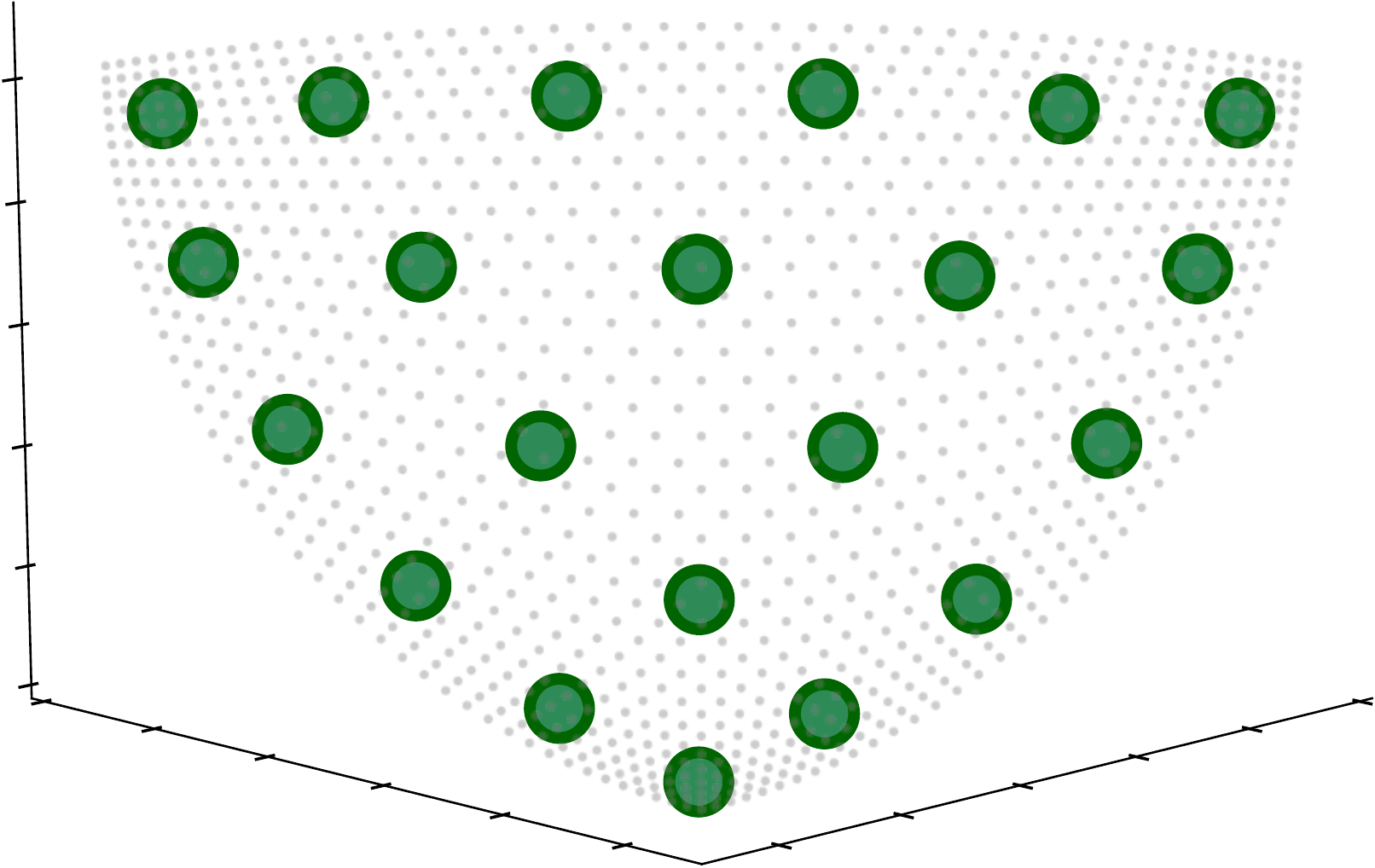}}
\\
\subfloat[$\vector{A}_{\rm IGD^+}$]{\includegraphics[width=\widthvar\textwidth]{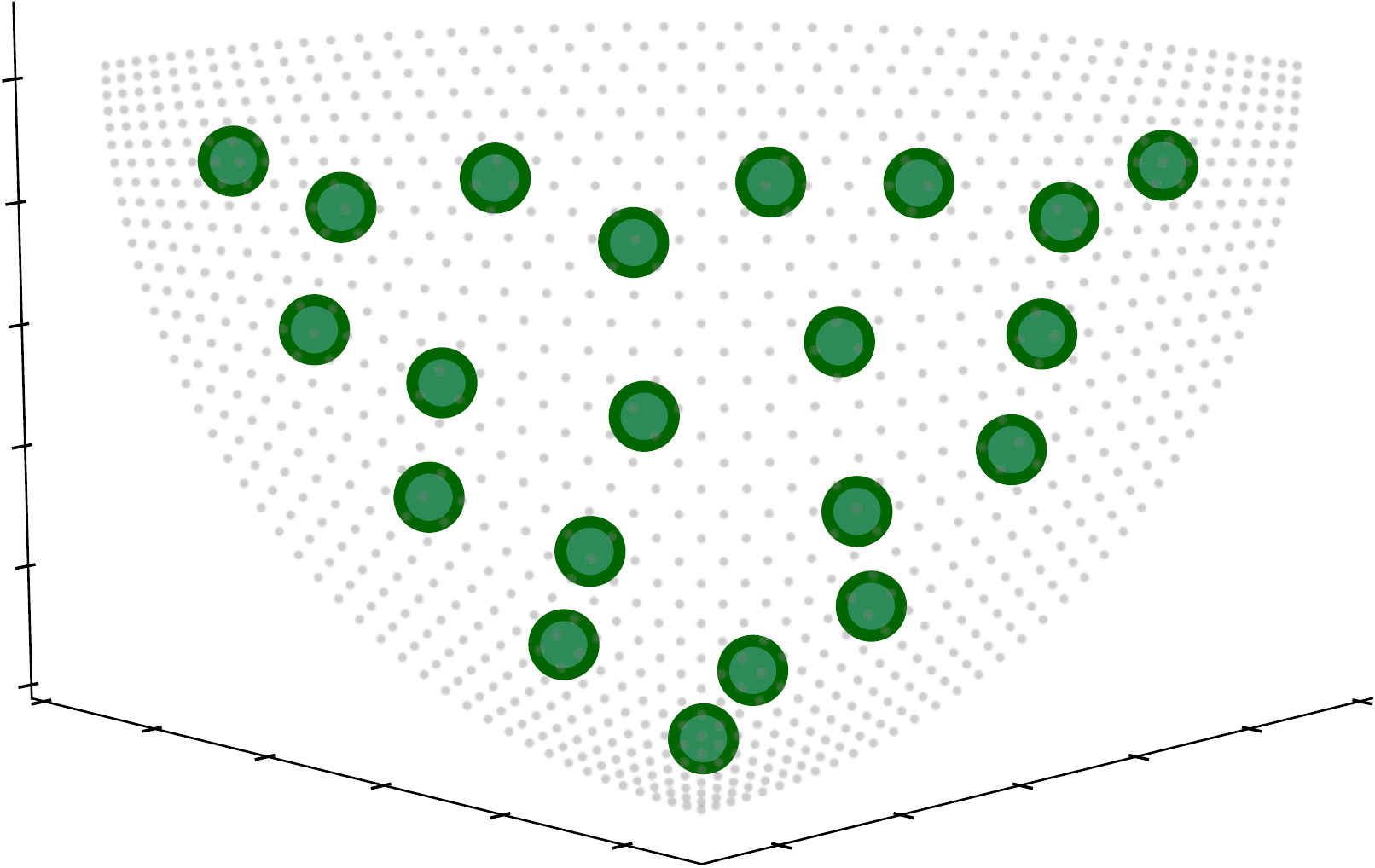}}
\subfloat[$\vector{A}_{\rm R2}$]{\includegraphics[width=\widthvar\textwidth]{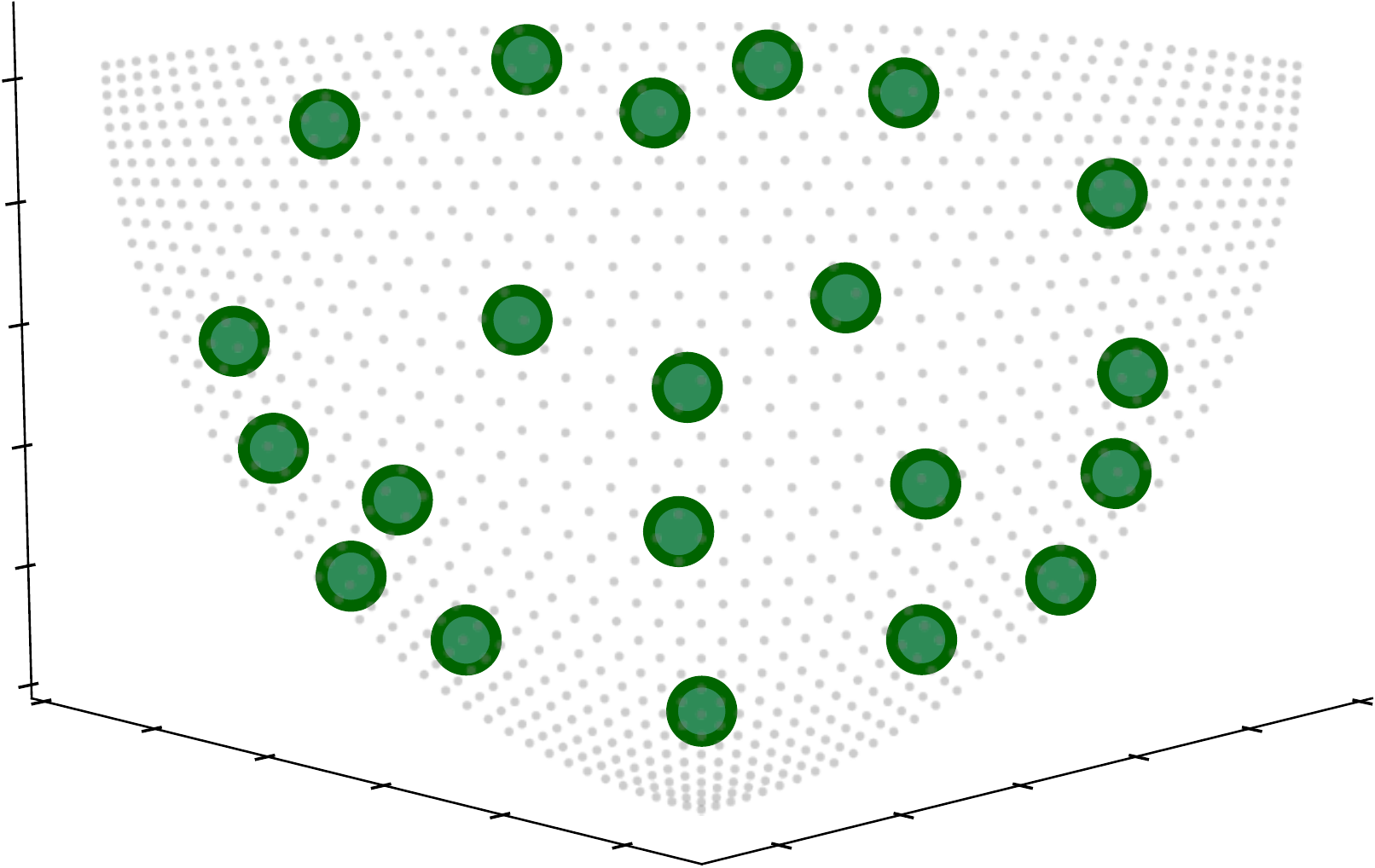}}
\\
\subfloat[$\vector{A}_{\rm NR2}$]{\includegraphics[width=\widthvar\textwidth]{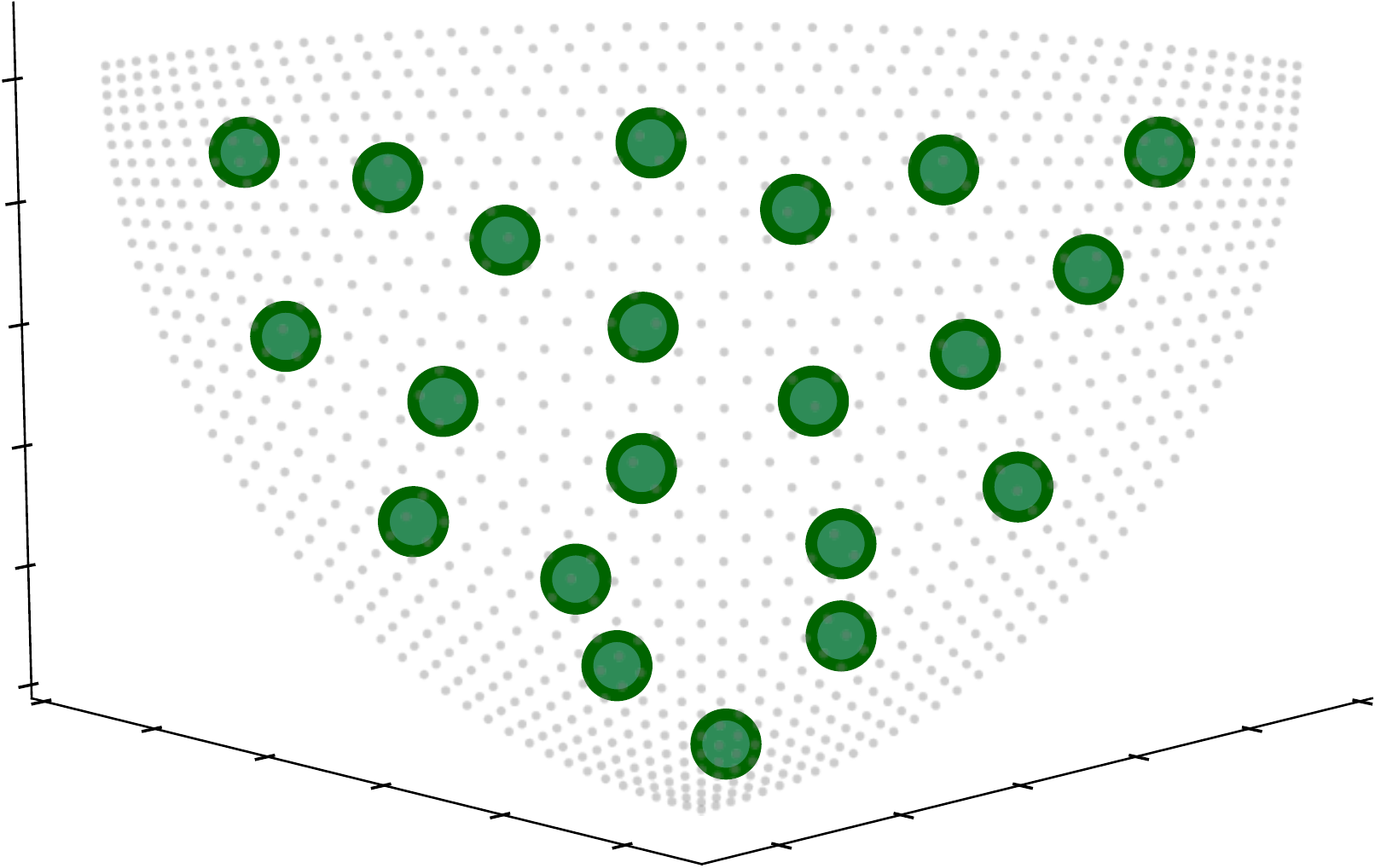}}
\subfloat[$\vector{A}_{I_{\epsilon+}}$]{\includegraphics[width=\widthvar\textwidth]{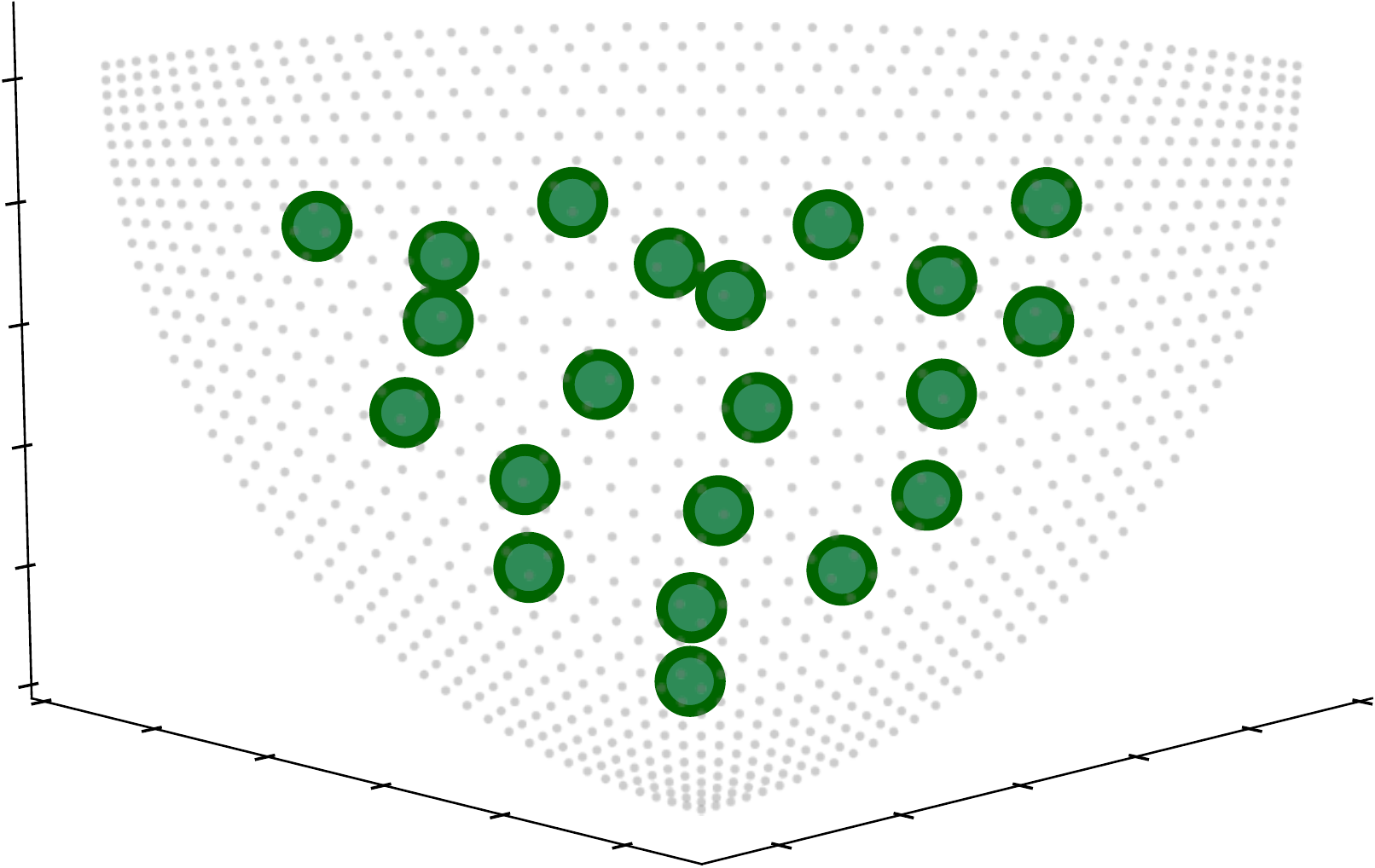}}
\\
\subfloat[$\vector{A}_{\rm SE}$]{\includegraphics[width=\widthvar\textwidth]{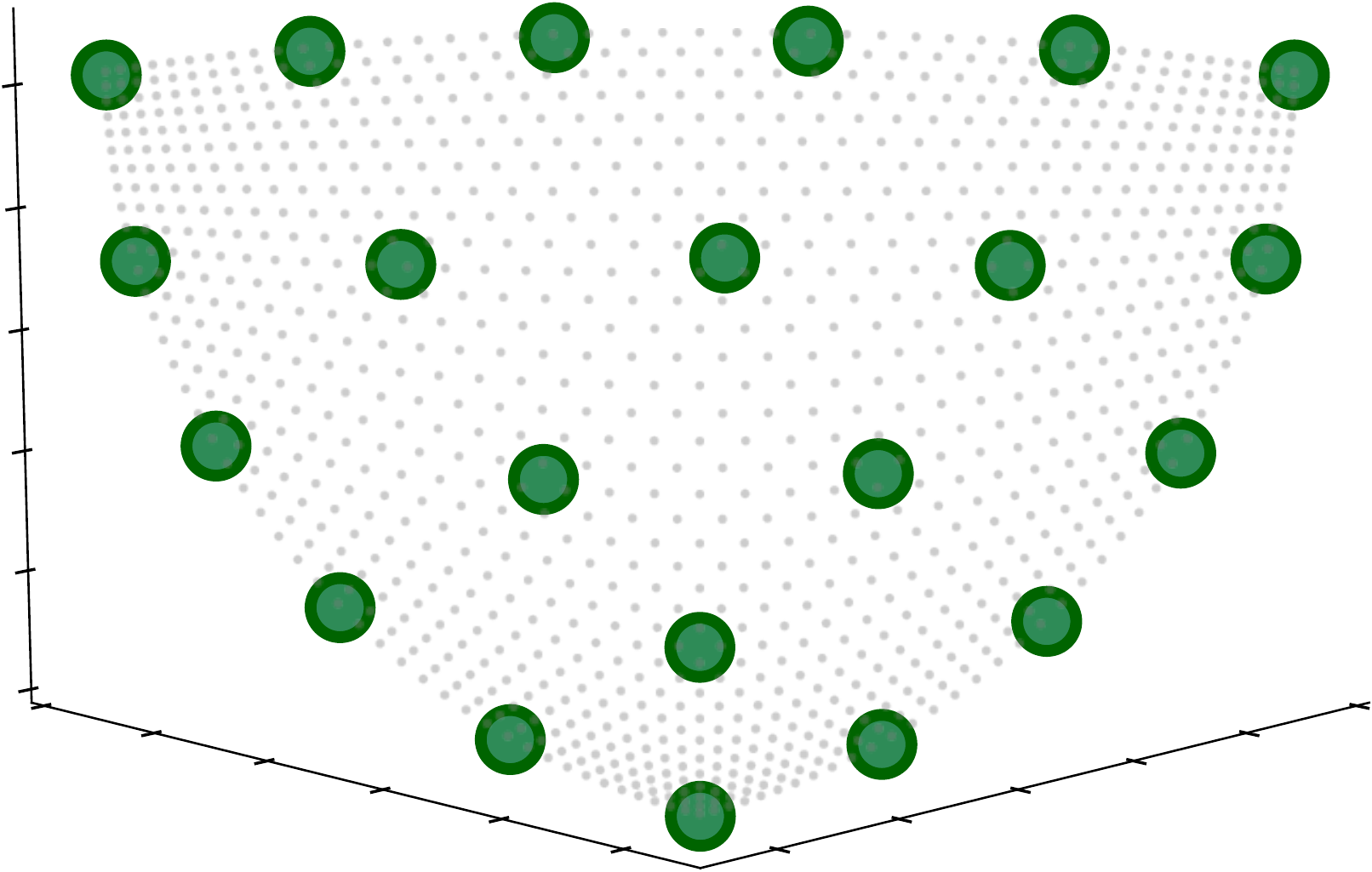}}
\subfloat[$\vector{A}_{\Delta}$]{\includegraphics[width=\widthvar\textwidth]{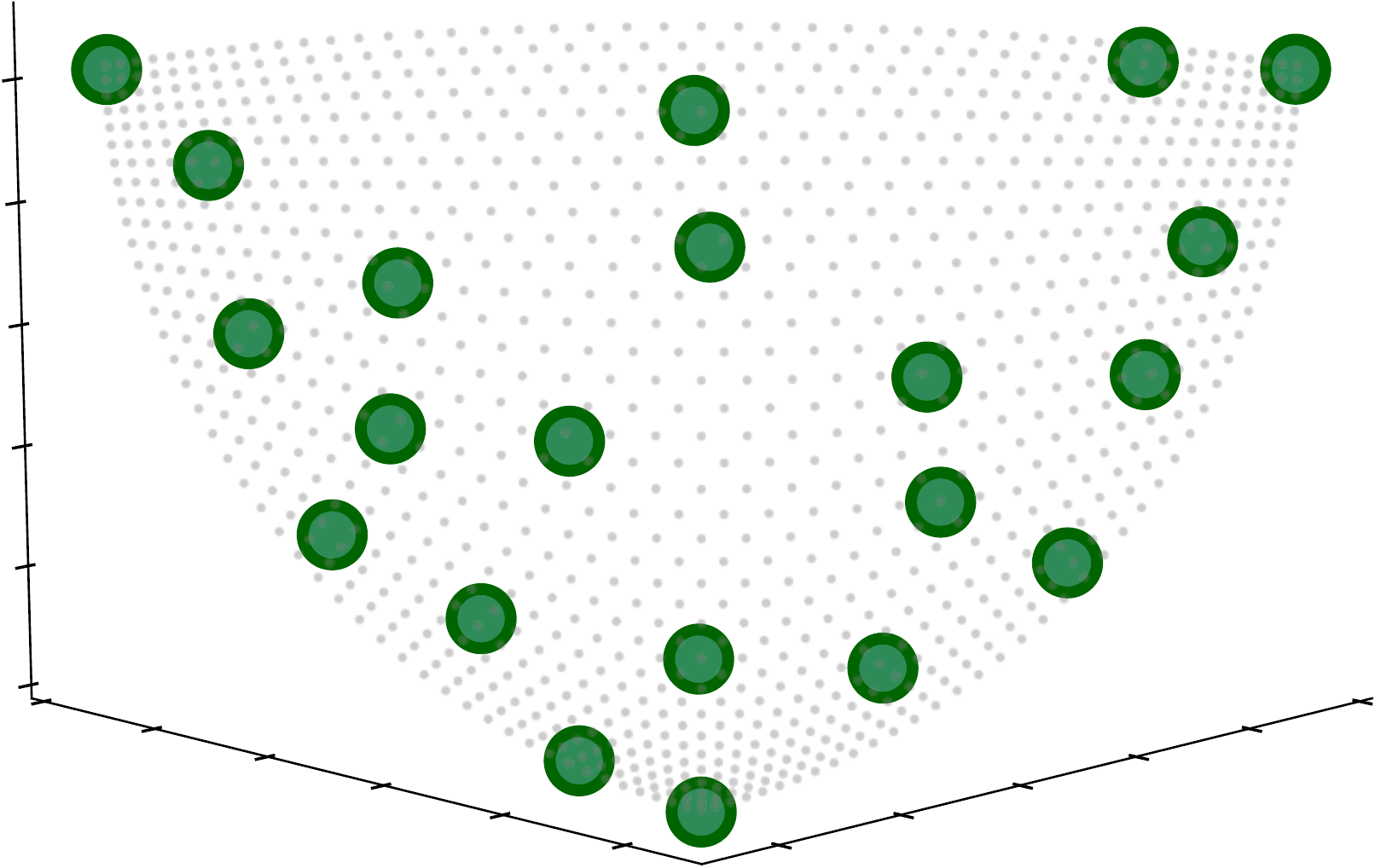}}
\\
\subfloat[$\vector{A}_{\rm PD}$]{\includegraphics[width=\widthvar\textwidth]{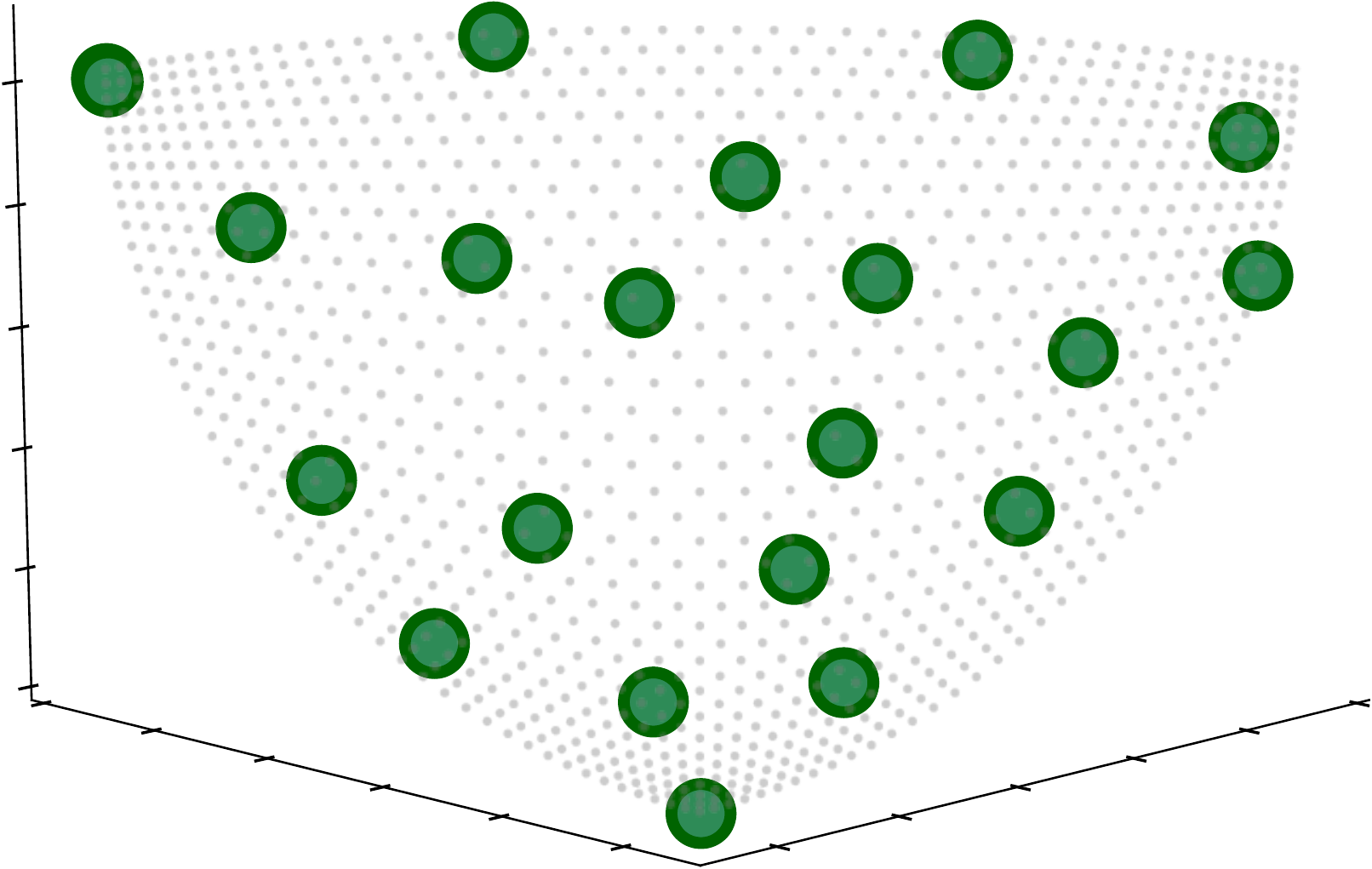}}
\subfloat[$\vector{A}_{\rm SLD}$]{\includegraphics[width=\widthvar\textwidth]{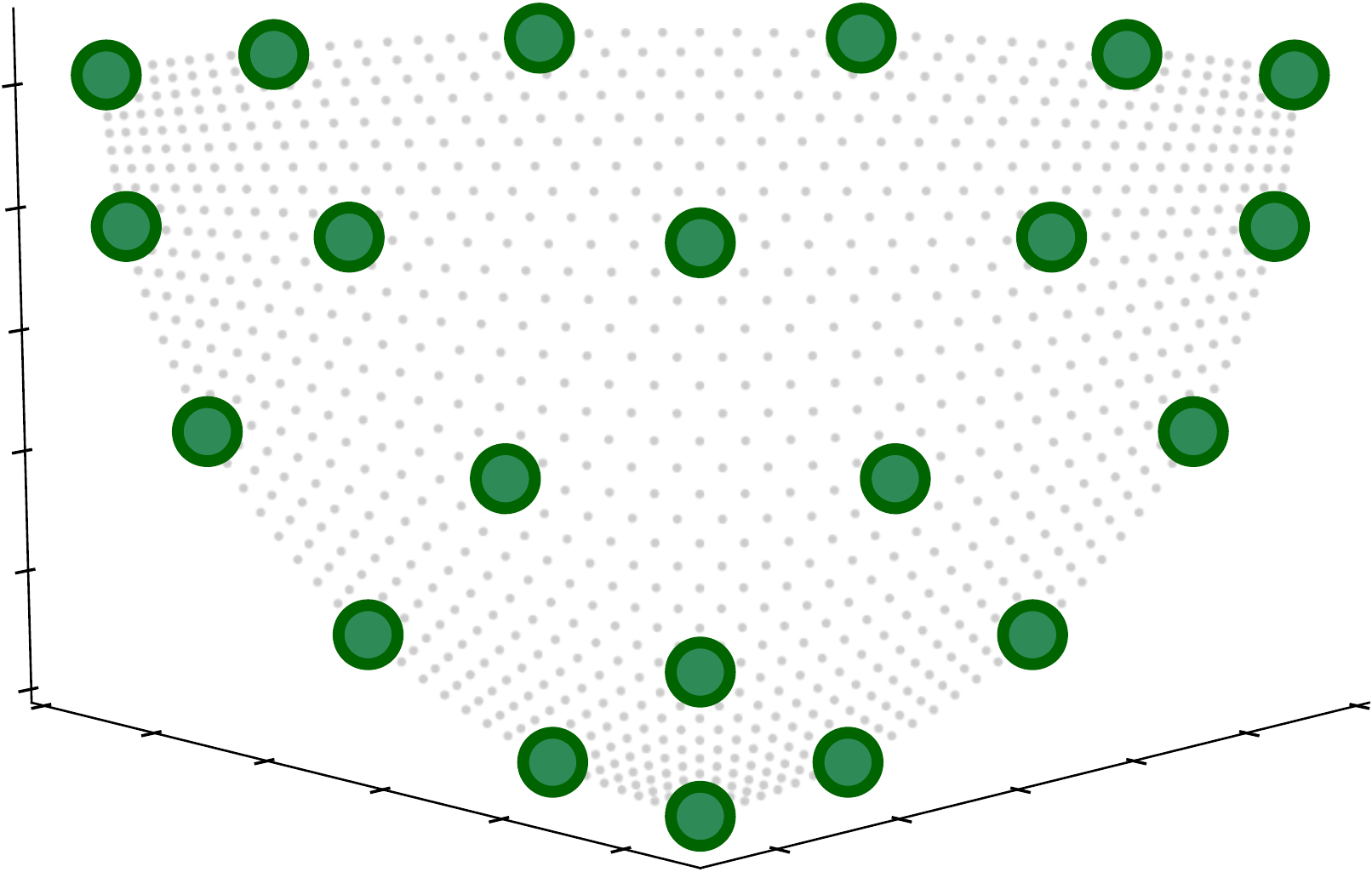}}
\caption{
\small
Results on $F_{\rm i\shyp convex}$.
}
\label{fig:iconvex}
  \end{minipage}
\end{tabular}
\end{figure*}

\begin{figure}[t]
\newcommand{\widthvar}{0.14}
\centering
\subfloat[$\vector{A}_{\rm HV}$]{\includegraphics[width=\widthvar\textwidth]{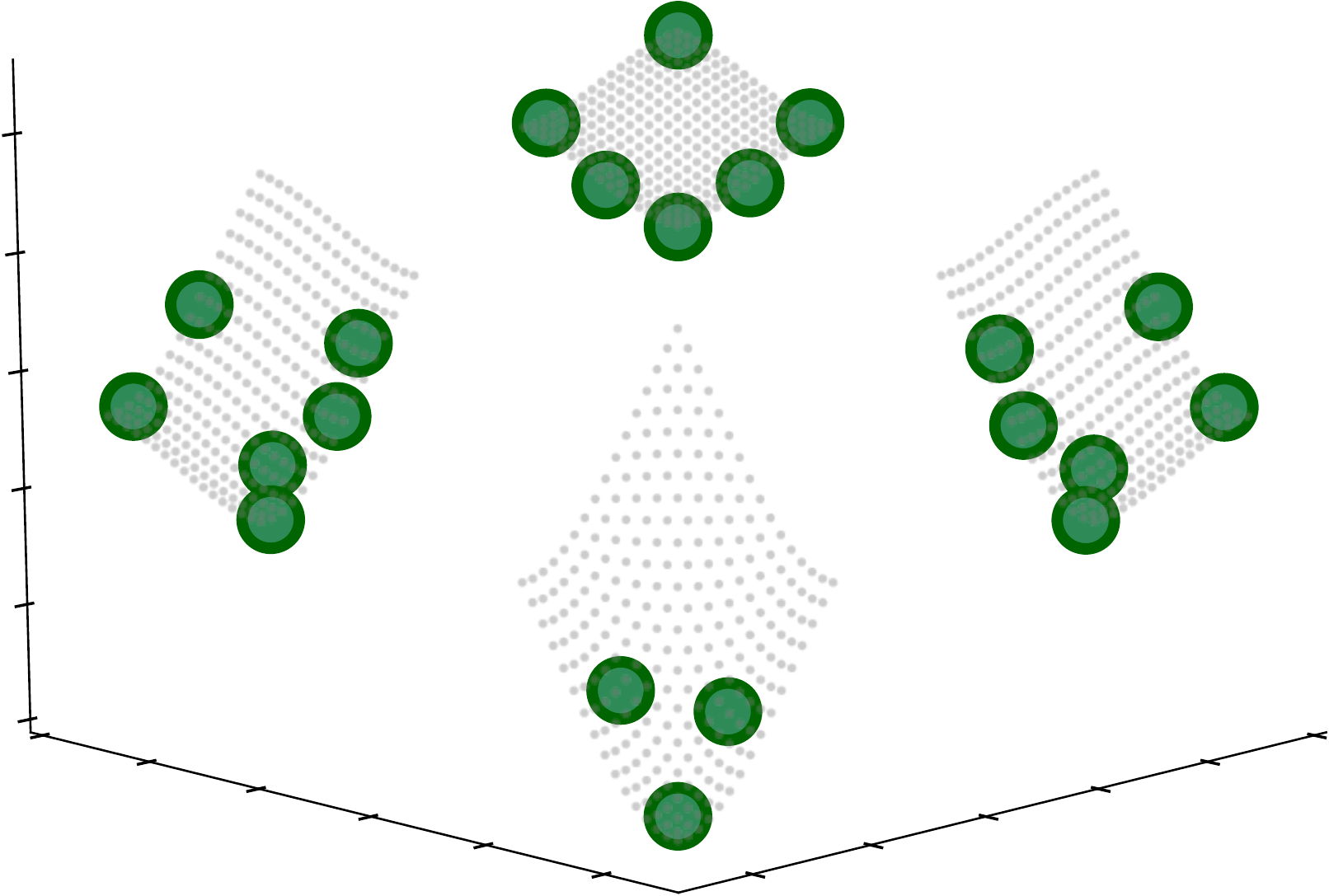}}
\subfloat[$\vector{A}_{\rm IGD}$]{\includegraphics[width=\widthvar\textwidth]{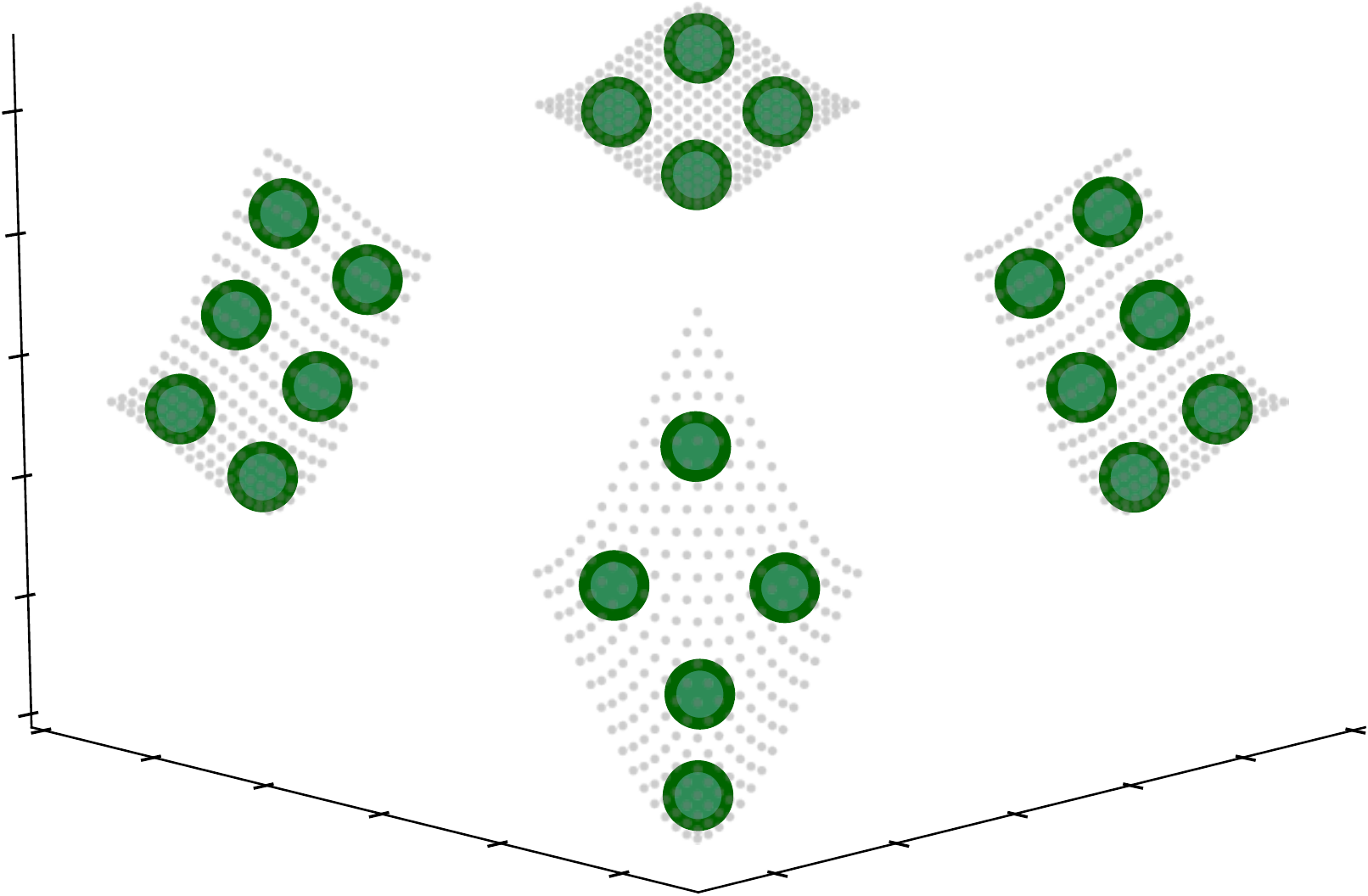}}
\subfloat[$\vector{A}_{\rm IGD^+}$]{\includegraphics[width=\widthvar\textwidth]{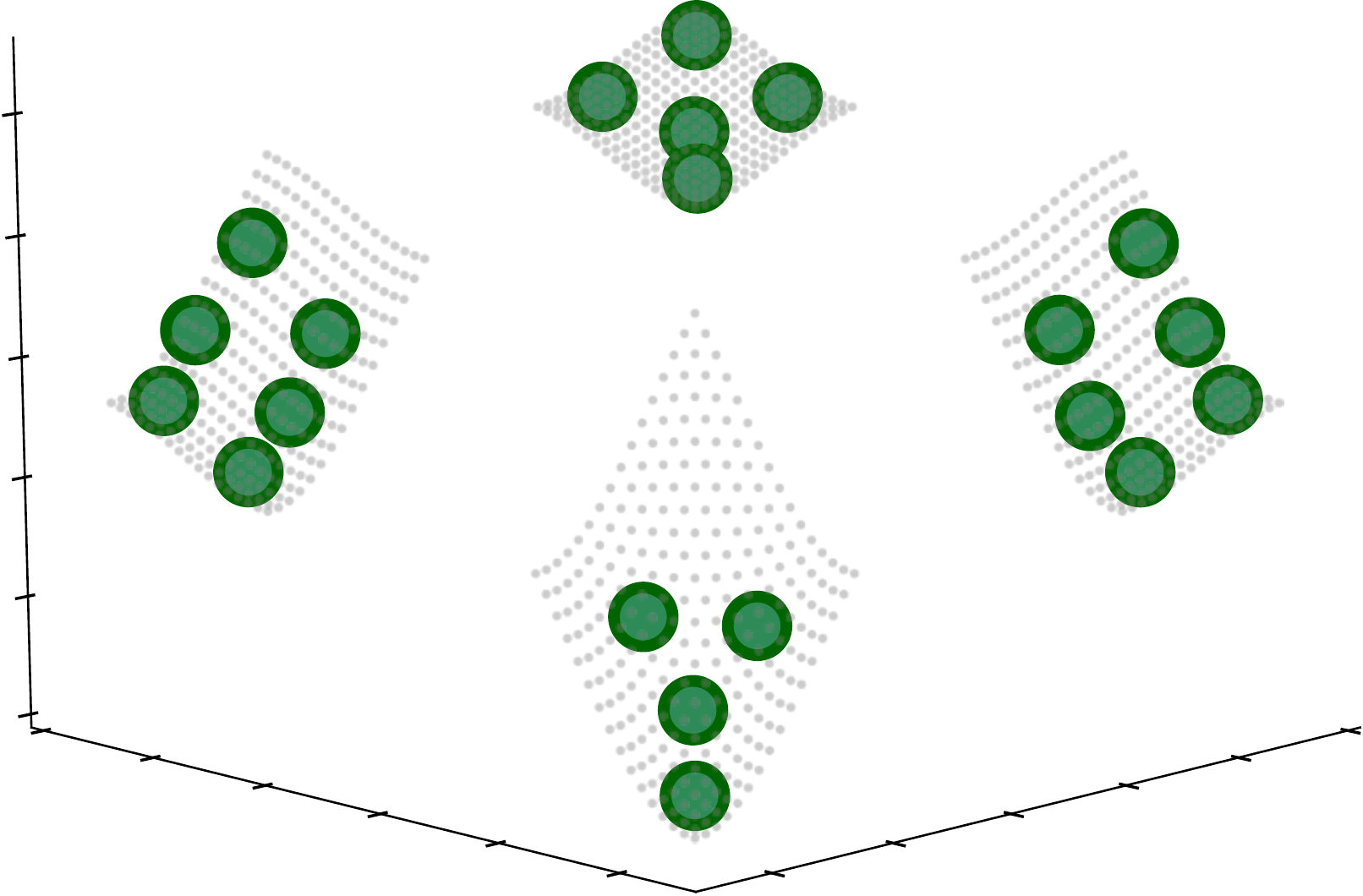}}
\\
\subfloat[$\vector{A}_{\rm R2}$]{\includegraphics[width=\widthvar\textwidth]{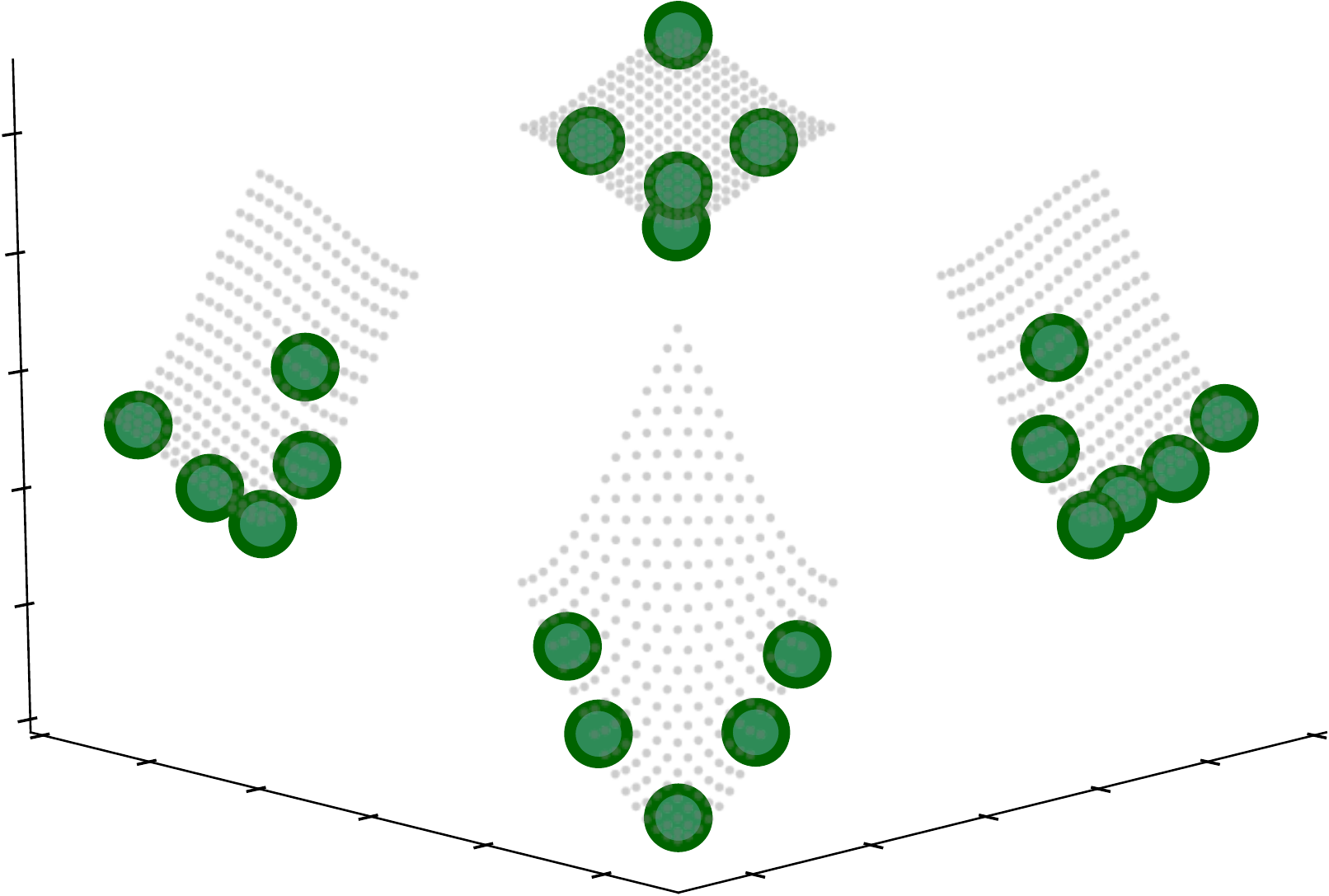}}
\subfloat[$\vector{A}_{\rm NR2}$]{\includegraphics[width=\widthvar\textwidth]{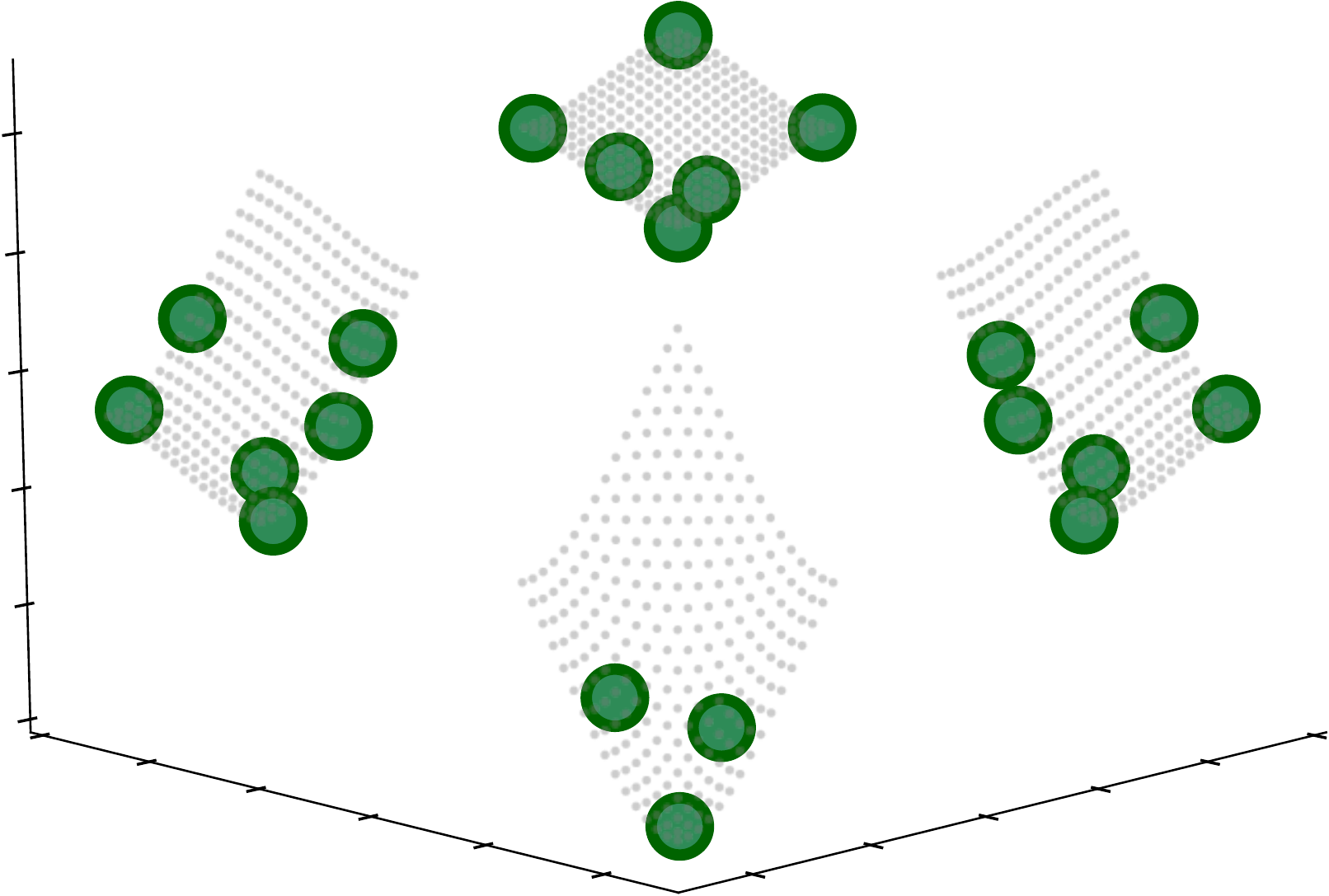}}
\subfloat[$\vector{A}_{I_{\epsilon+}}$]{\includegraphics[width=\widthvar\textwidth]{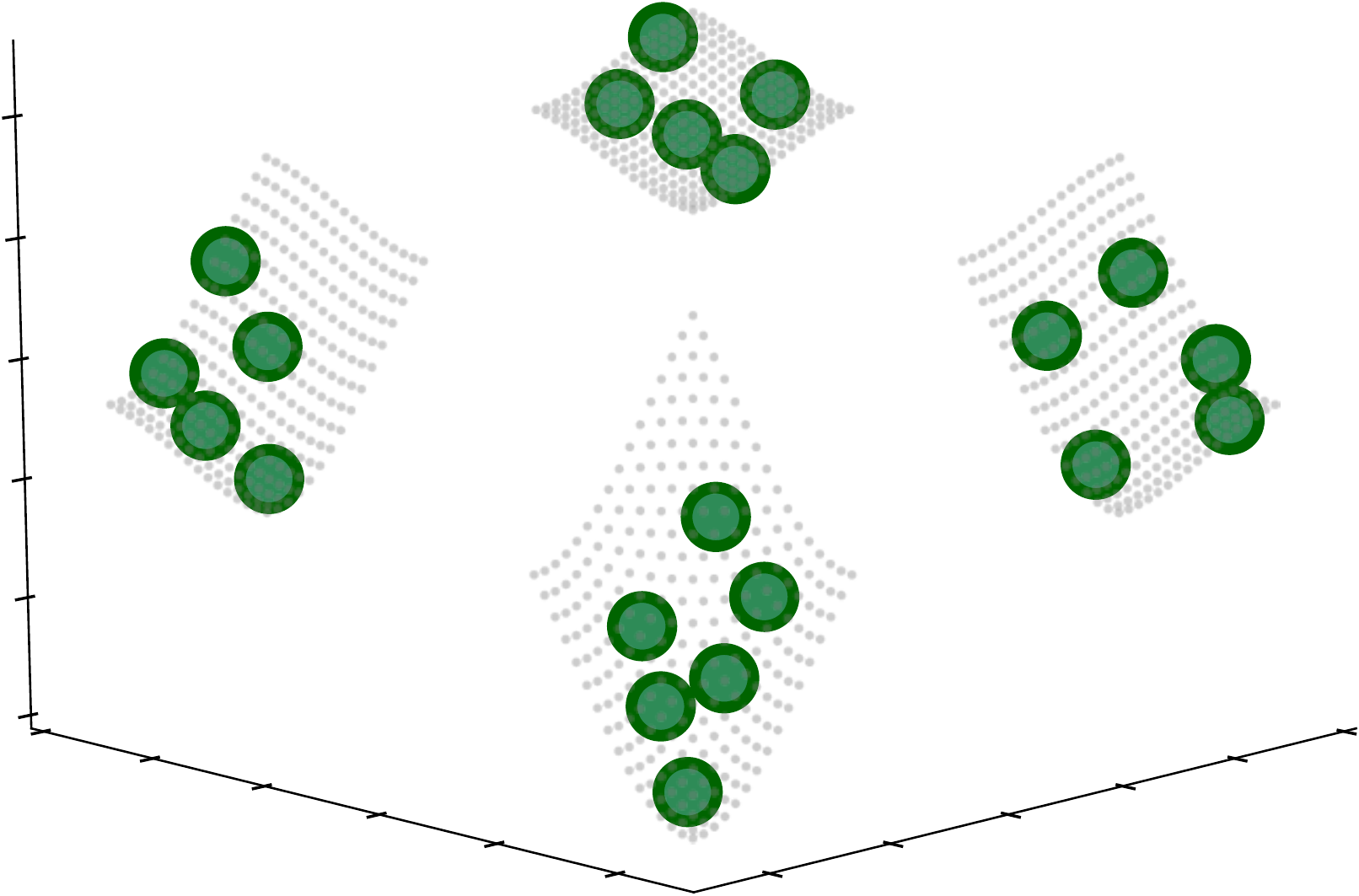}}
\\
\subfloat[$\vector{A}_{\rm SE}$]{\includegraphics[width=\widthvar\textwidth]{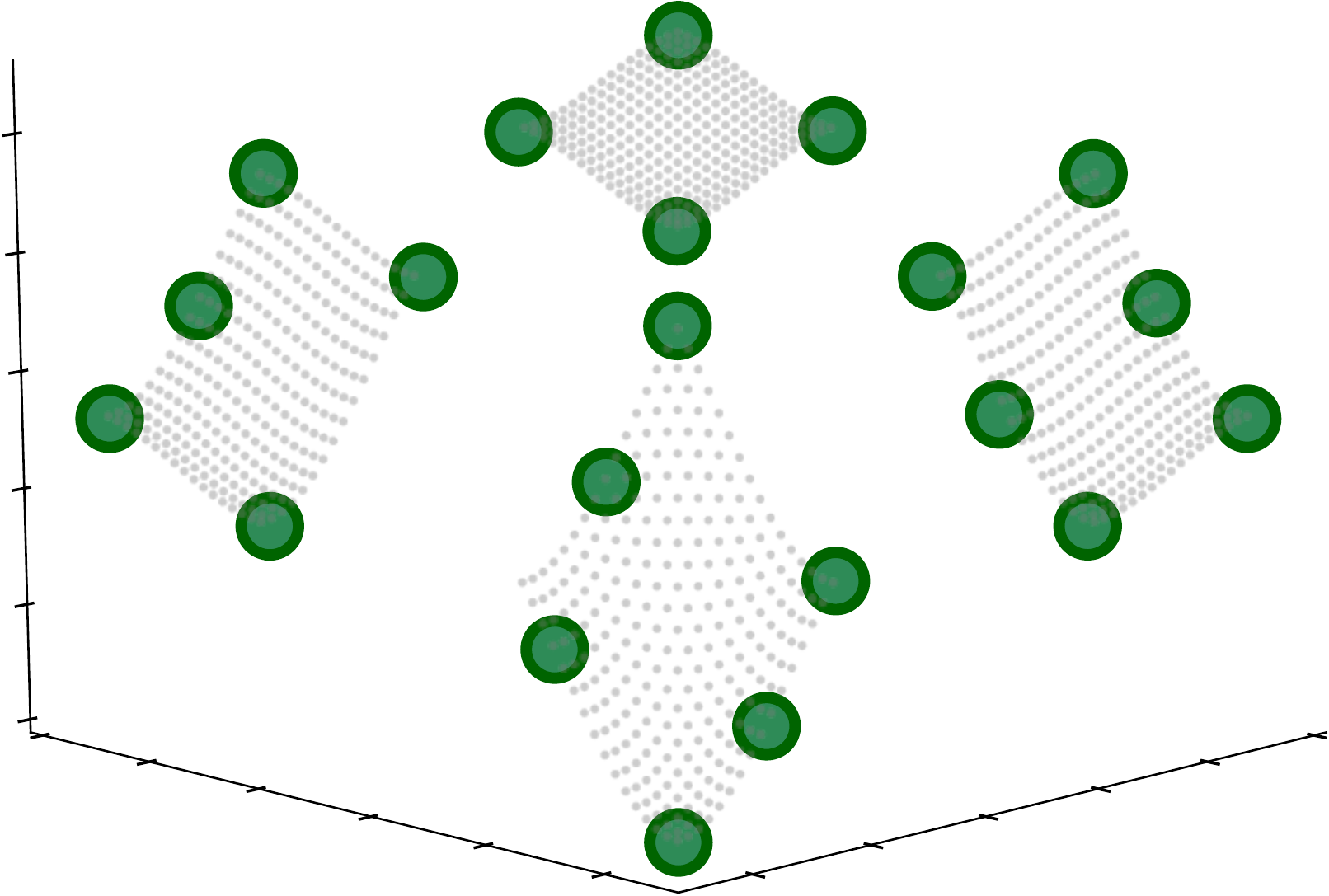}}
\subfloat[$\vector{A}_{\Delta}$]{\includegraphics[width=\widthvar\textwidth]{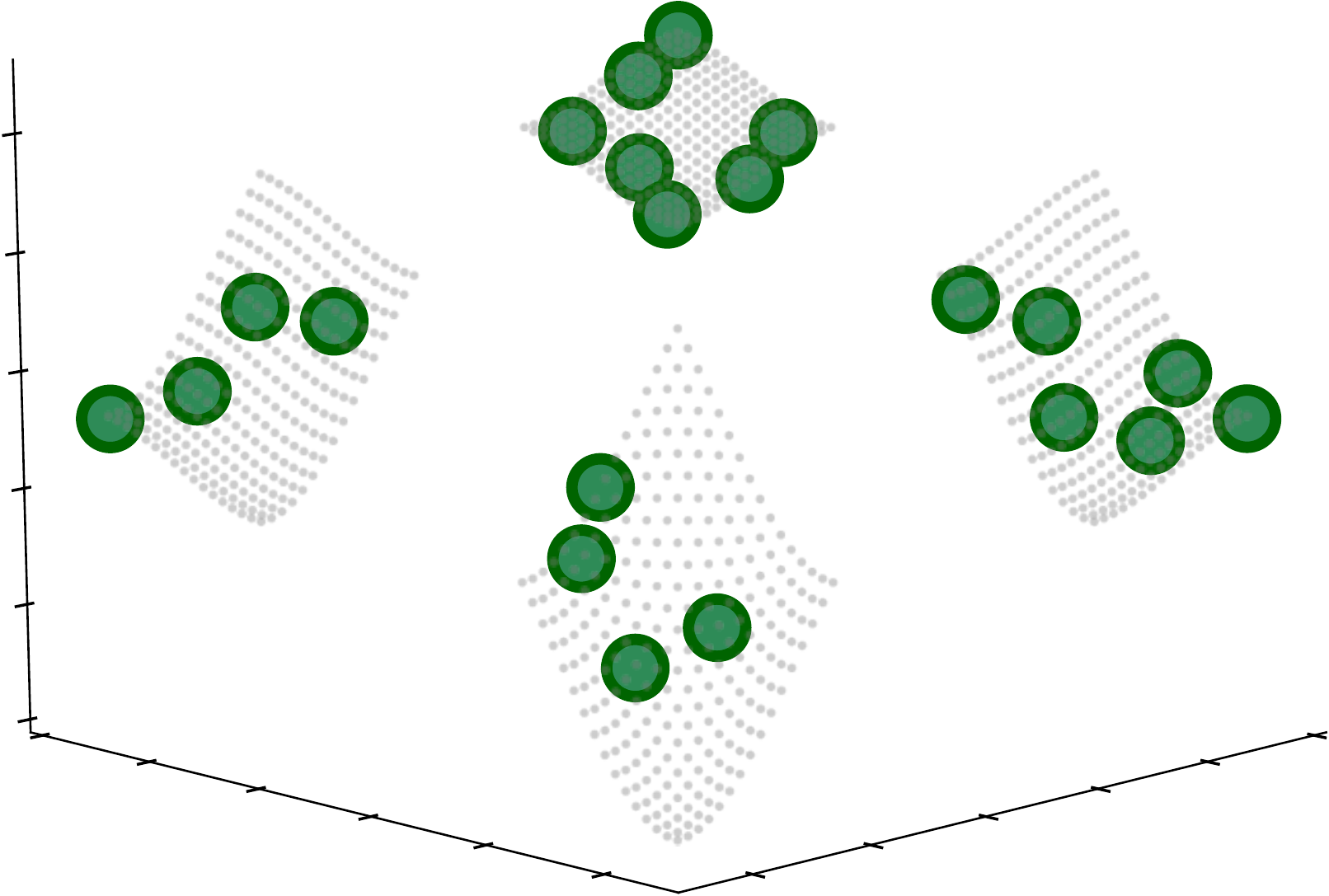}}
\subfloat[$\vector{A}_{\rm PD}$]{\includegraphics[width=\widthvar\textwidth]{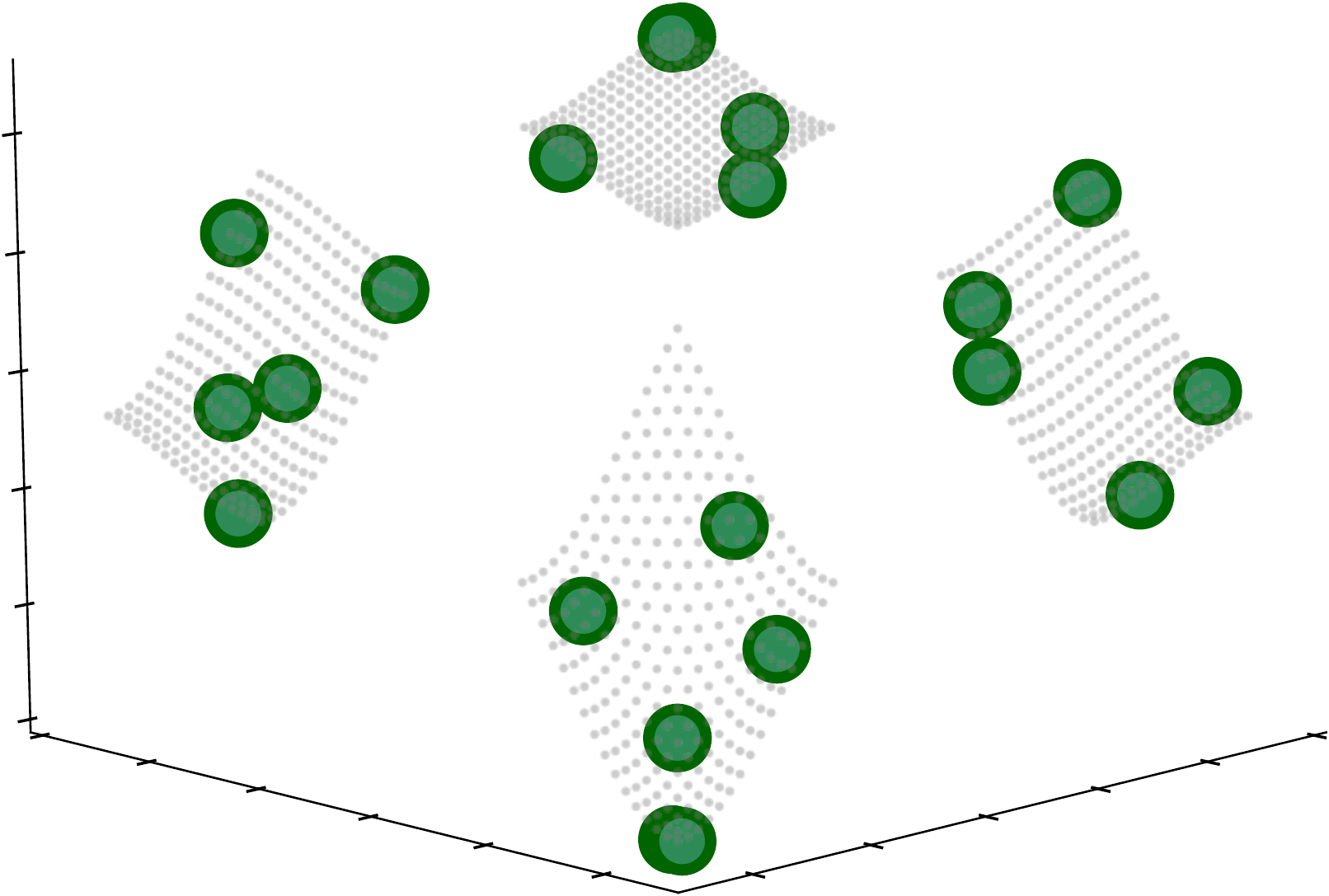}}
\caption{
\small
Approximated optimal $\mu$-distributions on $F_{\rm disconnected}$.
}
\label{fig:disconnected}
\subfloat[$\vector{A}_{\rm HV}$]{\includegraphics[width=\widthvar\textwidth]{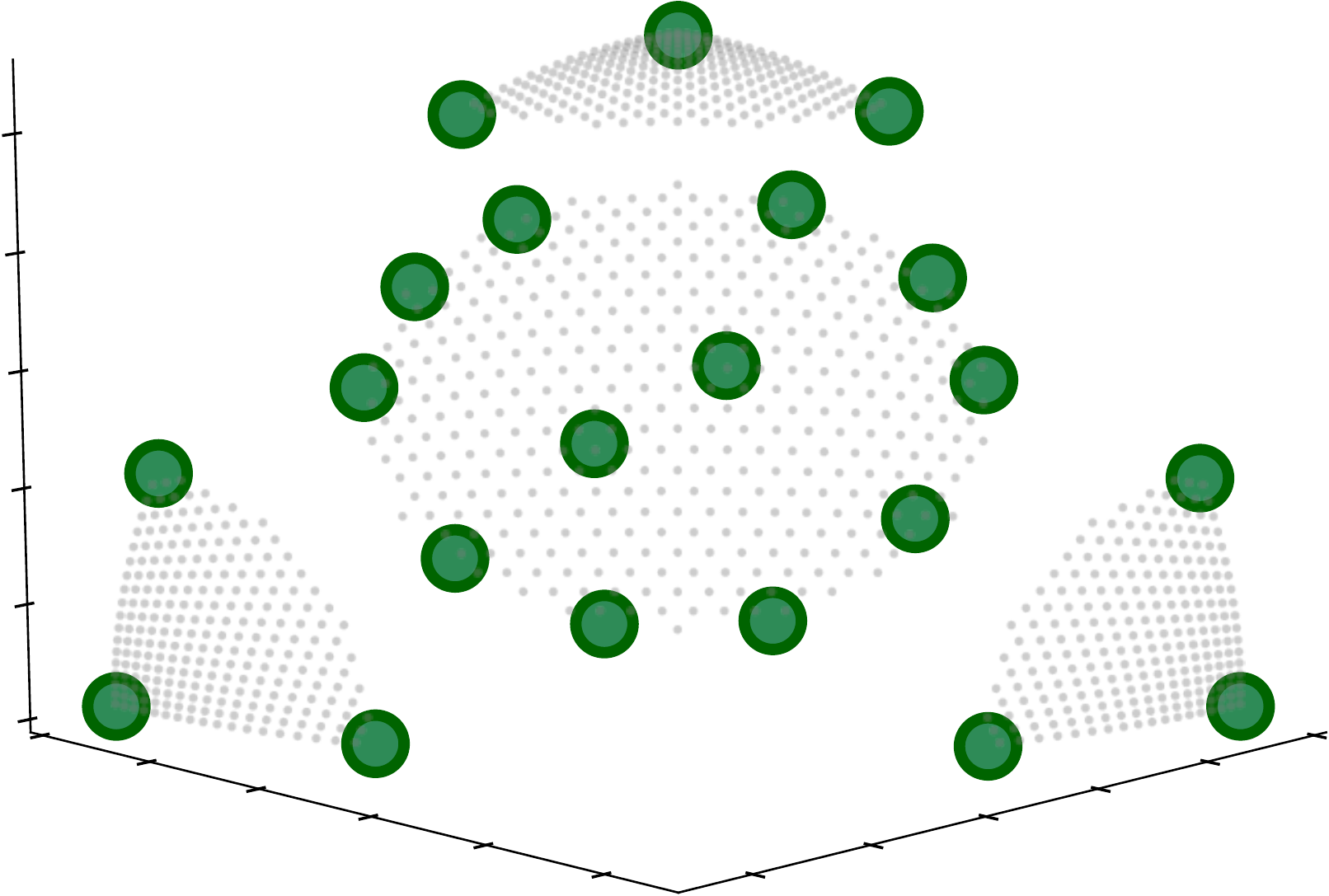}}
\subfloat[$\vector{A}_{\rm IGD}$]{\includegraphics[width=\widthvar\textwidth]{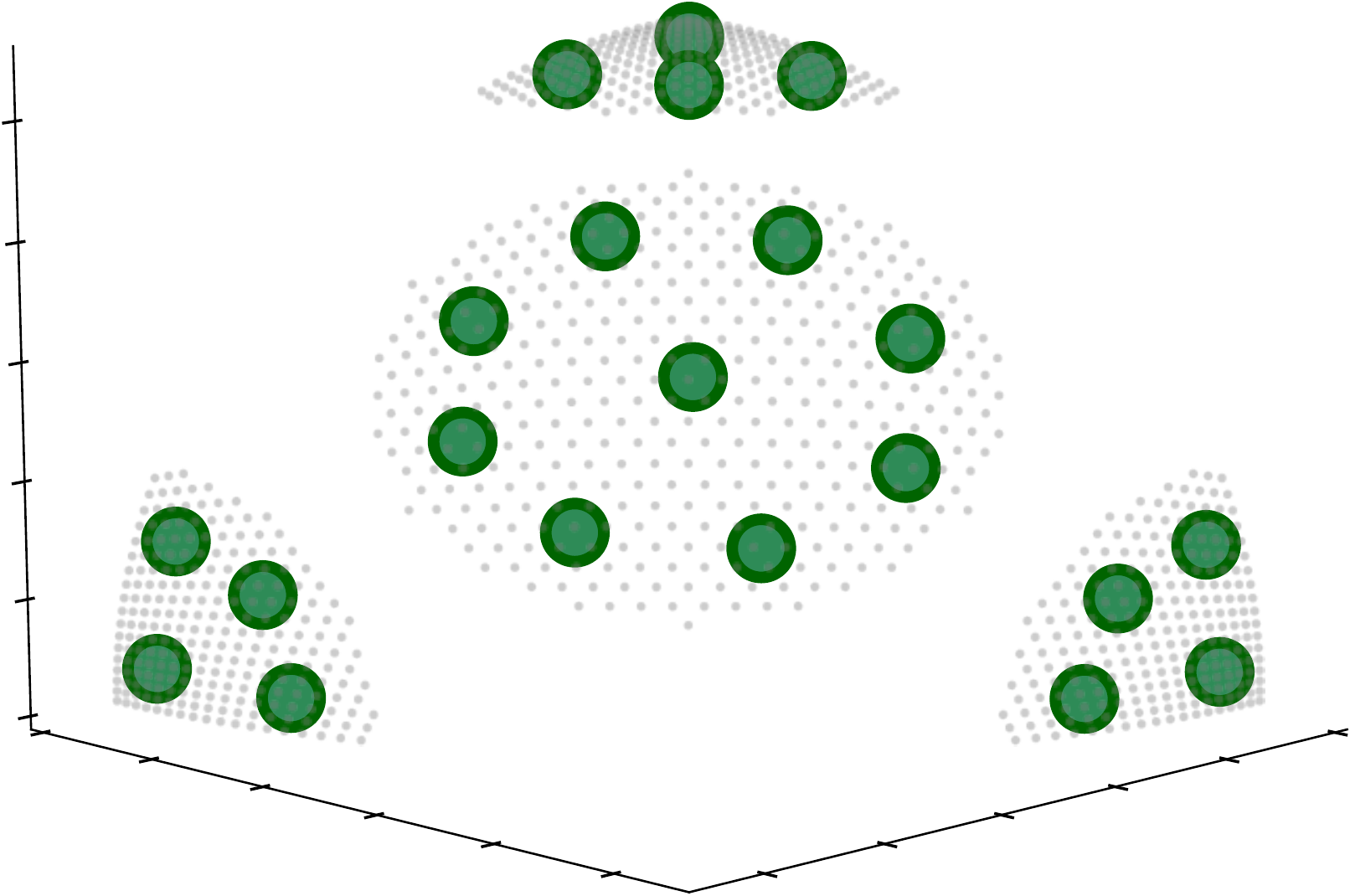}}
\subfloat[$\vector{A}_{\rm IGD^+}$]{\includegraphics[width=\widthvar\textwidth]{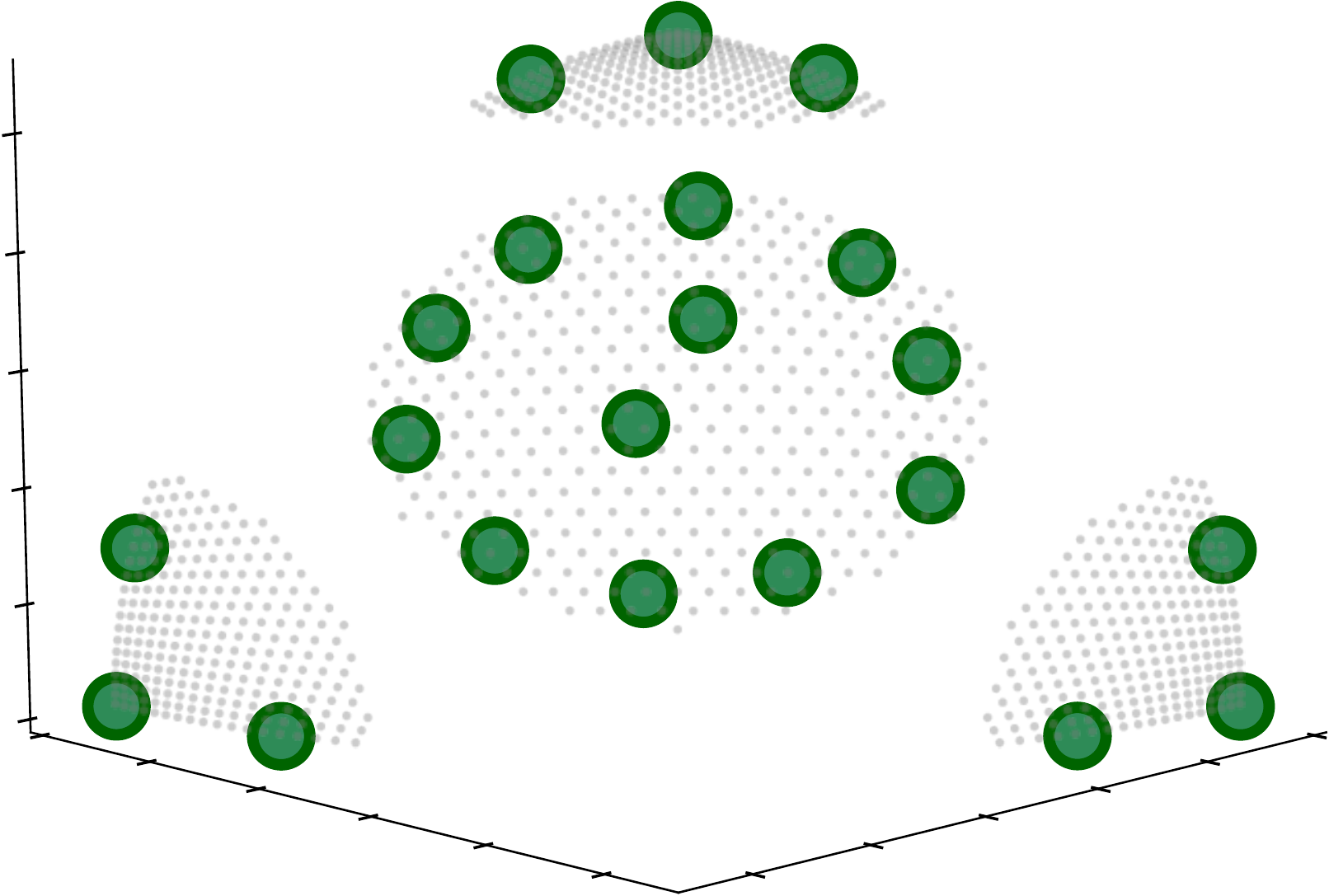}}
\\
\subfloat[$\vector{A}_{\rm R2}$]{\includegraphics[width=\widthvar\textwidth]{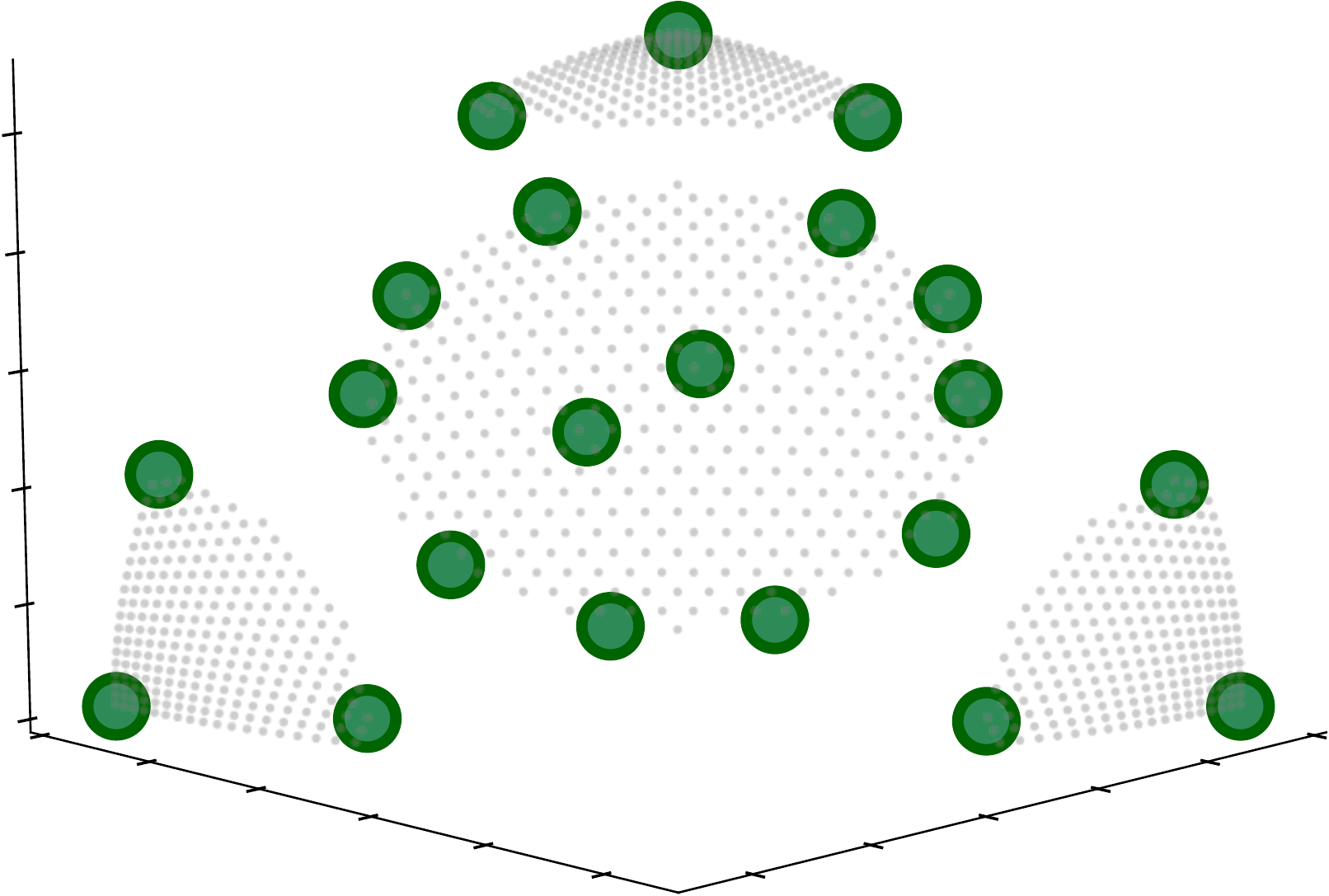}}
\subfloat[$\vector{A}_{\rm NR2}$]{\includegraphics[width=\widthvar\textwidth]{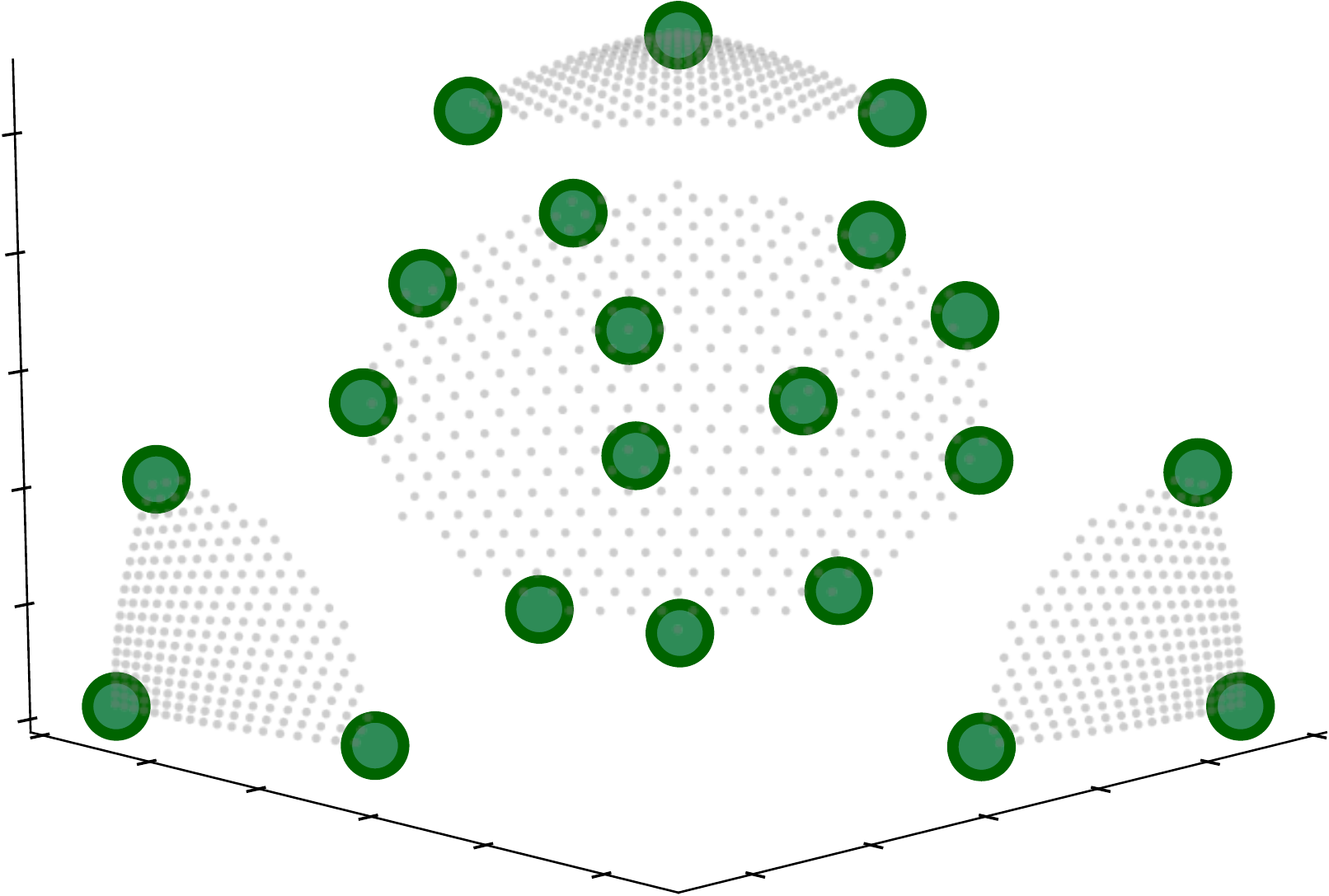}}
\subfloat[$\vector{A}_{I_{\epsilon+}}$]{\includegraphics[width=\widthvar\textwidth]{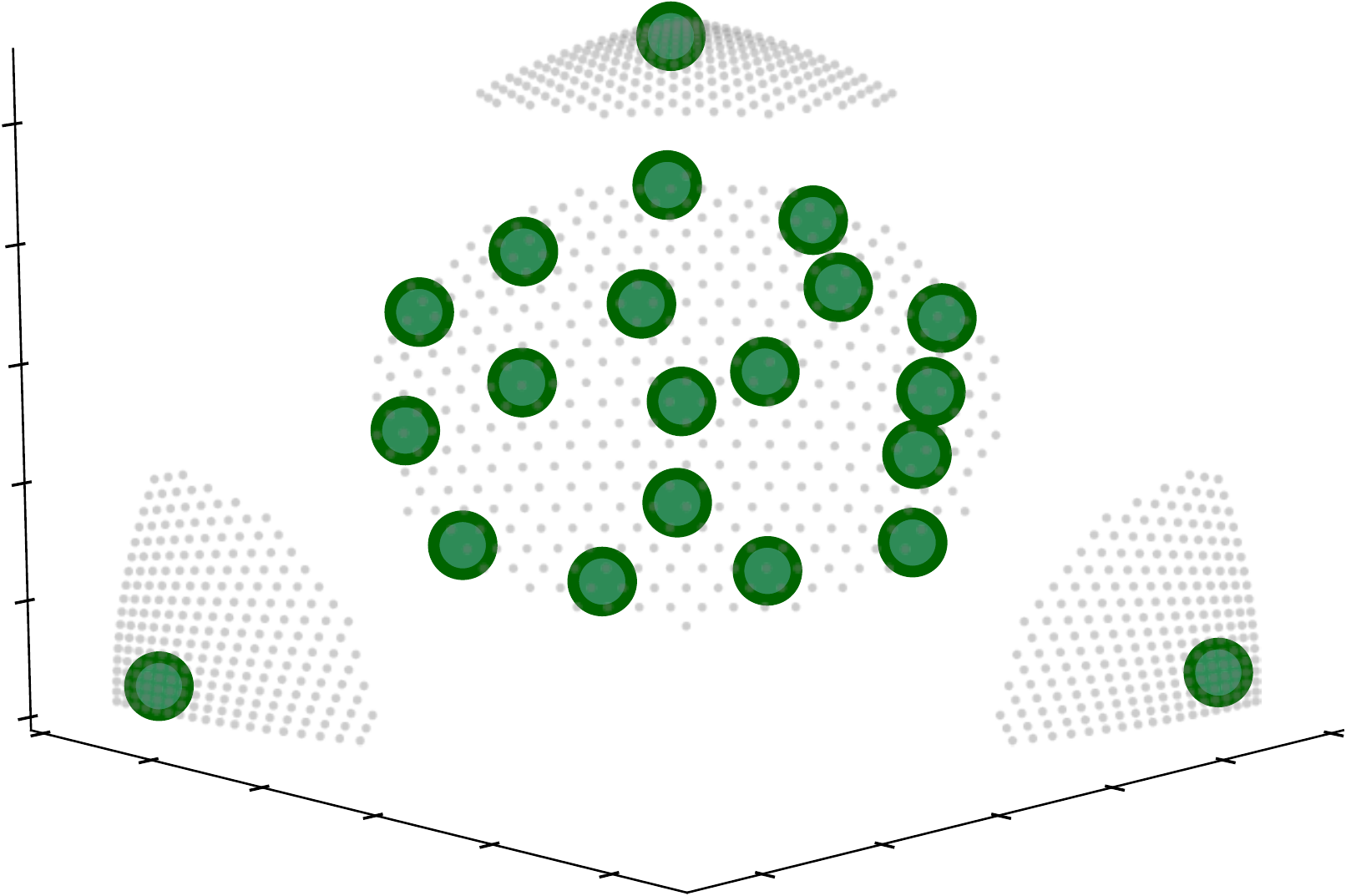}}
\\
\subfloat[$\vector{A}_{\rm SE}$]{\includegraphics[width=\widthvar\textwidth]{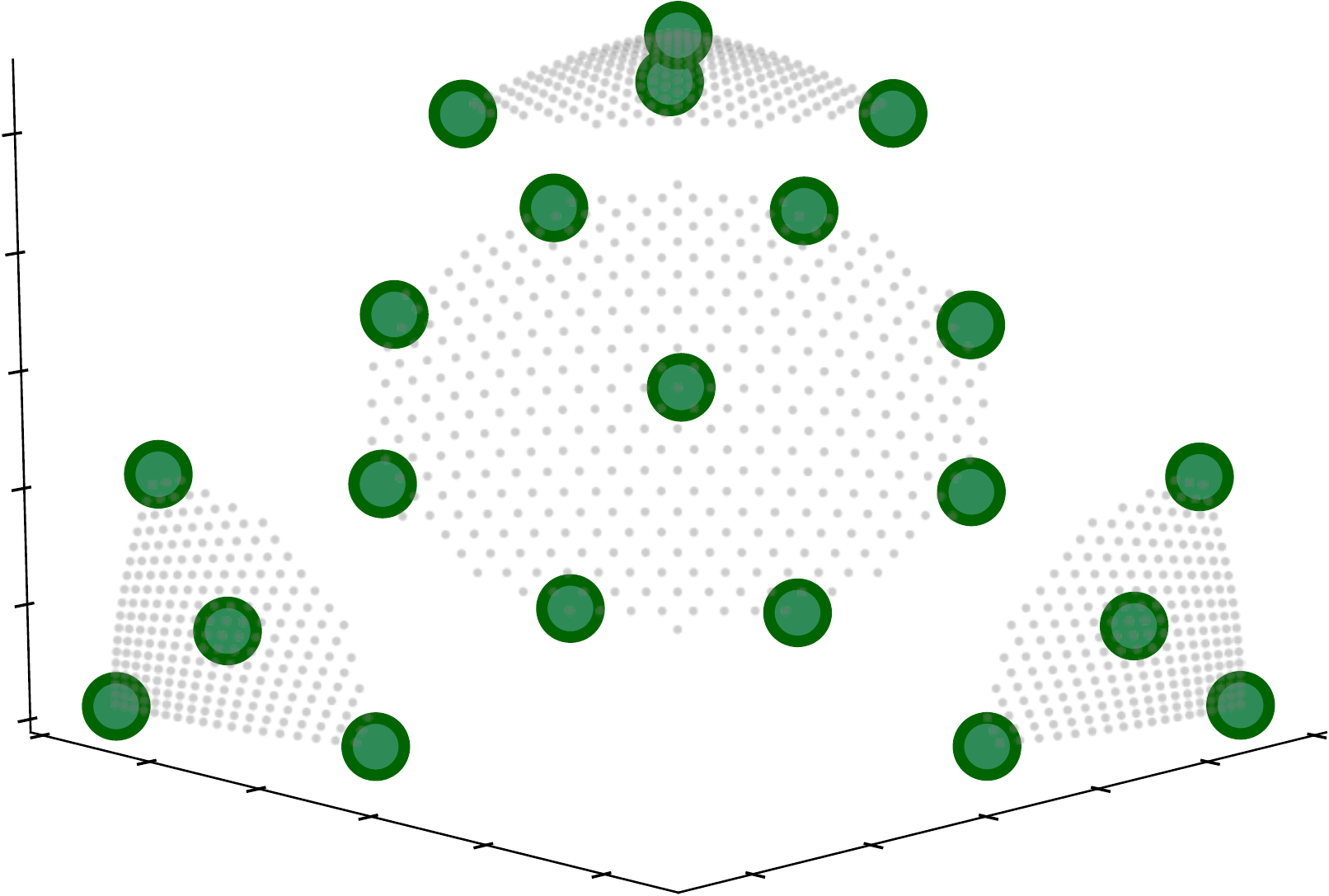}}
\subfloat[$\vector{A}_{\Delta}$]{\includegraphics[width=\widthvar\textwidth]{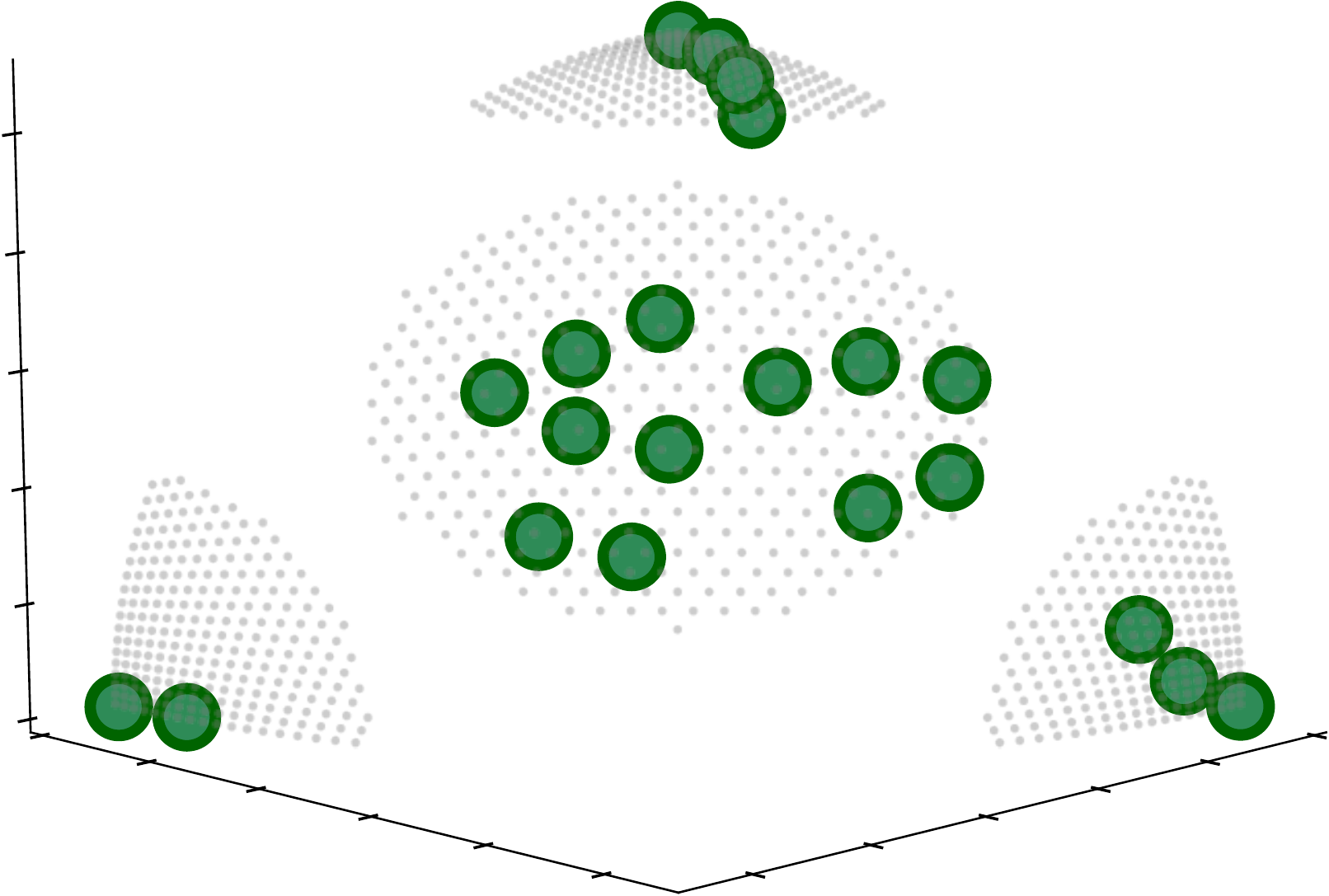}}
\subfloat[$\vector{A}_{\rm PD}$]{\includegraphics[width=\widthvar\textwidth]{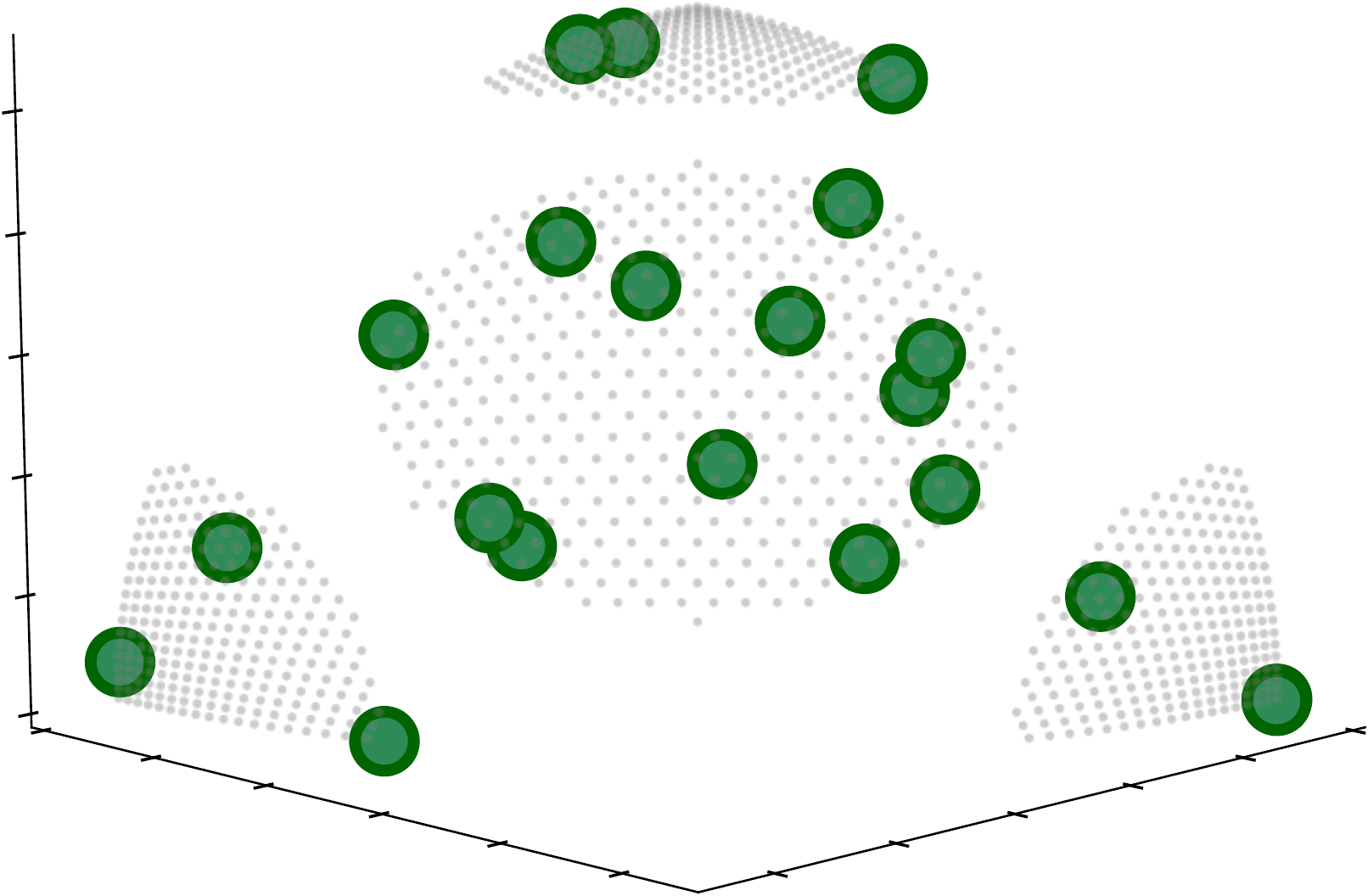}}
\caption{
\small
Approximated optimal $\mu$-distributions on $F_{\rm c\shyp concave}$.
}
\label{fig:c2-dtlz2}
\end{figure}

\section{Experimental results}
\label{sec:experimental_results}

This section analyzes the nine quality indicators using their optimal $\mu$-distributions approximated by the proposed formulation.
Subsection \ref{sec:appex_optimal_objs} discusses the optimal $\mu$-distributions for each quality indicator.
Subsection \ref{sec:ranking_analysis} examines the nine quality indicators using the ranking information.
Subsection \ref{sec:m5_8} analyzes the generality of results for $m=3$ with respect to $m$.
Subsection \ref{sec:mary} investigates a unary version of an $M$-nary quality indicator.








\subsection{Approximated optimal $\mu$-distributions for the nine quality indicators}
\label{sec:appex_optimal_objs}


Figs. \ref{fig:linear}--\ref{fig:c2-dtlz2} show approximated optimal $\mu$-distributions with $\mu=21$ obtained by the proposed approach for the nine quality indicators on the eight Pareto fronts.
In Figs. \ref{fig:linear}--\ref{fig:c2-dtlz2}, $x$, $y$, and $z$ axes represent $f_1$, $f_2$, and $f_3$, respectively.
Figs. \ref{fig:linear}--\ref{fig:c2-dtlz2} show the best objective vector set found by L-SHADE among 31 runs for each quality indicator on each front shape function $F$.
Below, we denote the best $\mu$-distribution for a quality indicator $I$ as $\vector{A}_I$ (e.g., $\vector{A}_{\rm HV}$).
Figs. S.3--S.50 in the supplementary file show approximated optimal $\mu$-distributions with the other $\mu$ values.
All reference vectors in $\vector{R}$ are shown in each figure.





For the sake of comparison, Figs. \ref{fig:linear}--\ref{fig:iconvex} (j) show a set of $\mu$ objective vectors generated by simplex-lattice design, denoted as $\vector{A}_{\rm SLD}$.
We generated $\vector{A}_{\rm SLD}$ using the same method as for generating the reference vector set $\vector{R}$ described in Section \ref{sec:experimental_settings}.
Since the number of objective vectors generated by simplex-lattice design on $F_{\rm disconnected}$ and $F_{\rm c\shyp concave}$ cannot be specified arbitrarily due to their characteristics, we do not show $\vector{A}_{\rm SLD}$ in Figs. \ref{fig:disconnected} and \ref{fig:c2-dtlz2}.
$\vector{A}_{\rm SLD}$ can be viewed as an ``ideal'' objective vector set found by decomposition-based EMOAs (e.g., NSGA-III  \cite{DebJ14}).
Although the distribution of objective vectors in $\vector{A}_{\rm SLD}$ is uniform, $\vector{A}_{\rm SLD}$ is not optimal for all the nine quality indicators (see Subsection \ref{sec:ranking_analysis}).


Since the distribution of objective vectors in some objective vector sets is not uniform, one may think that the approximated objective vector sets are far from optimal.
However, as demonstrated in Subsection \ref{sec:ranking_analysis} using Table \ref{tab:concave_ranks}, some quality indicators do not prefer uniformly distributed objective vectors.
For example, as shown in Figs. \ref{fig:nonconvex} (j) and (h), $\vector{A}_{\rm SLD}$ clearly has a better uniformity than $\vector{A}_{\Delta}$ on $F_{\rm concave}$.
Nevertheless, Table \ref{tab:concave_ranks} shows that $\Delta$ evaluates $\vector{A}_{\Delta}$ as better than $\vector{A}_{\rm SLD}$ (details are discussed later).
Such an unintuitive result may be due to a hidden property of each quality indicator.
We reemphasize that a theoretical analysis of the optimal $\mu$-distribution for $m \geq 3$ is difficult.
For this reason, the (true and approximated) optimal $\mu$-distributions for almost all quality indicators have not been investigated on the Pareto front with $m \geq 3$.




Below, we individually discuss the optimal $\mu$-distribution for each of the nine quality indicators ($\vector{A}_{\rm HV}$, $\vector{A}_{\rm IGD}$, $\vector{A}_{\rm IGD^+}$, $\vector{A}_{\rm R2}$, $\vector{A}_{\rm NR2}$,  $\vector{A}_{I_{\epsilon+}}$, $\vector{A}_{\rm SE}$, $\vector{A}_{\Delta}$, and $\vector{A}_{\rm PD}$).
Although results of HV, IGD, and IGD$^+$ are consistent with the previous studies, some results of other quality indicators provide insightful information.
Our findings are summarized in Section \ref{sec:conclusion}.


\subsubsection{$\vector{A}_{\rm HV}$}

It should be noted that the optimal $\mu$-distribution for HV significantly depends on the position of the reference vector $\vector{q}$ \cite{IshibuchiISN18ecj}.
Objective vectors in $\vector{A}_{\rm HV}$ are on the edge and the center of the Pareto front for all the front shape functions, except for $F_{\rm convex}$ and $F_{\rm i\shyp convex}$.
These observations on the convex Pareto fronts are consistent with the previous studies \cite{AugerBBZ09,JiangOZF14,IshibuchiISN18ecj}.
When $\mu$ is small, $\vector{A}_{\rm HV}$ does not contain the extreme objective vectors even for the linear front shape function $F_{\rm linear}$ (e.g., $\vector{A}_{\rm HV}$ with $\mu=10$ shown in Fig. S.3 in the supplementary file).
As $\mu$ increases, $\vector{A}_{\rm HV}$ covers the edge of the Pareto front on most front shape functions.

\subsubsection{$\vector{A}_{\rm IGD}$}

The distribution of objective vectors in $\vector{A}_{\rm IGD}$ is uniform on most front shape functions, even including $F_{\rm c\shyp concave}$.
However, objective vectors in $\vector{A}_{\rm IGD}$ on all front shape functions are not on the edge of the Pareto front.
Thus, IGD may incorrectly evaluate the spread quality of objective vector sets.
The poor spread quality of $\vector{A}_{\rm IGD}$ is consistent with the previous studies \cite{IshibuchiISN18,IshibuchiIMN19}.


\subsubsection{$\vector{A}_{\rm IGD^+}$}

Our results are consistent with the results presented in \cite{IshibuchiIMN19}.
The distribution of objective vectors in $\vector{A}_{\rm IGD^+}$ is almost the same as that in $\vector{A}_{\rm IGD}$ on $F_{\rm linear}$ and similar to that in $\vector{A}_{\rm HV}$ on the other front shape functions.
As the $\mu$ value increases, the distributions of objective vectors in $\vector{A}_{\rm IGD^+}$ and $\vector{A}_{\rm HV}$ become more similar.
For details, see Figs. S.3--S.50 in the supplementary file.

%



\subsubsection{$\vector{A}_{\rm R2}$}




Objective vectors in $\vector{A}_{\rm R2}$ are densely distributed in the center of the Pareto front for $F_{\rm linear}$, $F_{\rm concave}$, and $F_{\rm convex}$.
These results are consistent with the previous study \cite{BrockhoffWT12} for $m=2$.
However, our results on $F_{\rm i\shyp linear}$ and $F_{\rm i\shyp concave}$ are inconsistent with \cite{BrockhoffWT12}.
Only one objective vector in $\vector{A}_{\rm R2}$ is in the center of the Pareto front for $F_{\rm i\shyp linear}$.
Almost all objective vectors in $\vector{A}_{\rm R2}$ are on the edge of the Pareto front.
Similarly, all objective vectors in $\vector{A}_{\rm R2}$ are on the edge of the Pareto front for $F_{\rm i\shyp concave}$.
When $\mu$ is set to a larger value, $\vector{A}_{\rm R2}$ contains a few objective vectors on the center of the Pareto front for $F_{\rm i\shyp concave}$ as shown in Figs. S.3--S.50.



Our inconsistent results on $F_{\rm i\shyp linear}$ and $F_{\rm i\shyp concave}$ are mainly due to the inverted triangular Pareto front.
The analysis presented in \cite{IshibuchiSMN16} shows that some decomposition-based EMOAs (e.g., MOEA/D and NSGA-III) perform poorly on problems with inverted triangular Pareto fronts.
This is because the shape of the distribution of weight vectors used in the decomposition-based EMOAs is different from the shape of the Pareto front.
Since R2 uses the weight vector set $\vector{W}$ that is similar to decomposition-based EMOAs, R2 is likely to have a similar issue in decomposition-based EMOAs for $m=3$.







\subsubsection{$\vector{A}_{\rm NR2}$}

Since NR2 was designed to approximate the HV value, $\vector{A}_{\rm NR2}$ is very similar to $\vector{A}_{\rm HV}$ for all front shape functions.
Although NR2 is the modified version of R2, $\vector{A}_{\rm NR2}$ is dissimilar to $\vector{A}_{\rm R2}$.
For example, $\vector{A}_{\rm R2}$ does not contain any objective vector in the center of the Pareto front for $F_{\rm i\shyp concave}$.
In contrast, $\vector{A}_{\rm NR2}$ contains some objective vectors in the center of the Pareto front for $F_{\rm i\shyp concave}$ that is similar to $\vector{A}_{\rm HV}$.




\subsubsection{$\vector{A}_{I_{\epsilon+}}$}



As mentioned in Subsection \ref{sec:related_work_astar}, the optimal $\mu$-distributions for $I_{\epsilon+}$ on $m=2$ were investigated in  \cite{BringmannFK15}.
However, our results on $m=3$ are inconsistent with \cite{BringmannFK15}.
The distribution of objective vectors in $\vector{A}_{I_{\epsilon+}}$ is not uniform for all front shape functions (except for $F_{\rm i\shyp concave}$).
Thus, our results show that optimal $\mu$-distributions for $I_{\epsilon+}$ depend on the choice of $m$.
As pointed out in \cite{LiY19}, the $I_{\epsilon+}$ value of a given objective vector set $\vector{A}$ depends only on one objective value of one objective vector in $\vector{A}$.
This property of $I_{\epsilon+}$ may influence its optimal $\mu$-distributions on $m = 3$ in a complicated manner.
An in-depth analysis of the optimal $\mu$-distributions for $I_{\epsilon+}$ on $m = 3$ is another future work.








\subsubsection{$\vector{A}_{\rm SE}$}


Similar to $\vector{A}_{\rm IGD}$, the distribution of objective vectors in $\vector{A}_{\rm SE}$ is uniform for all front shape functions.
In contrast, objective vectors in $\vector{A}_{\rm SE}$ are more widely distributed than those in $\vector{A}_{\rm IGD}$.
For example, while objective vectors in $\vector{A}_{\rm IGD}$ are densely distributed in the center of the Pareto front for $F_{\rm convex}$, those in $\vector{A}_{\rm SE}$ cover the whole Pareto front.
We can conclude that SE evaluates the quality of objective vector sets in terms of both uniformity and spread.

However, most objective vectors in $\vector{A}_{\rm SE}$ are on the edge of the Pareto front (especially on $F_{\rm disconnected}$).
Even as the $\mu$ value increases, this characteristic can be still found (see Figs. S.3--S.50).
Thus, SE may overestimate a set of objective vectors on the edge of the Pareto front.



 \arrayrulecolor{black}
 \setlength\arrayrulewidth{1.2pt}
 \setlength{\extrarowheight}{1.2pt}
 \newcolumntype{A}{>{\columncolor[rgb]{0.9, 0.9, 0.9}}p{2em}}

\newcolumntype{C}{>{\centering\arraybackslash}X}
\newcolumntype{L}{>{\raggedright\arraybackslash}X}
\newcolumntype{R}{>{\raggedleft\arraybackslash}X}



 \definecolor{c10}{RGB}{255,50,50}
\definecolor{c9}{RGB}{255,70,70}
\definecolor{c8}{RGB}{255,90,90}
\definecolor{c7}{RGB}{255,110,110}
\definecolor{c6}{RGB}{255,130,130}

\definecolor{c5}{RGB}{140,255,140}
\definecolor{c4}{RGB}{160,255,160}
\definecolor{c3}{RGB}{180,255,180}
\definecolor{c2}{RGB}{200,255,200}
\definecolor{c1}{RGB}{225,255,225}

\begin{table*}[t]
  \centering
  \caption{\small Rankings of the nine approximated optimal $\mu$-distributions ($\vector{A}_{\rm HV}$, ..., $\vector{A}_{\rm PD}$) and $\vector{A}_{\rm SLD}$ by each quality indicator on $F_{\rm concave}$.}
  \label{tab:concave_ranks}
{\footnotesize
%
    \begin{tabularx}{47em}{|A |C|C|C|C|C|C|C|C|C|C|}
      \hline
      \rowcolor[rgb]{0.9, 0.9, 0.9} & $\vector{A}_{\rm HV}$ & $\vector{A}_{\rm IGD}$ & $\vector{A}_{\rm IGD^+}$ & $\vector{A}_{\rm R2}$ & $\vector{A}_{\rm NR2}$ & $\vector{A}_{I_{\epsilon+}}$ & $\vector{A}_{\rm SE}$ & $\vector{A}_{\Delta}$ & $\vector{A}_{\rm PD}$ & $\vector{A}_{\rm SLD}$\\
\hline
HV & \cellcolor{c1}1  & \cellcolor{c10}10  & \cellcolor{c3}3  & \cellcolor{c5}5  & \cellcolor{c2}2  & \cellcolor{c7}7  & \cellcolor{c4}4  & \cellcolor{c8}8  & \cellcolor{c9}9  & \cellcolor{c6}6  \\\hline
IGD & \cellcolor{c9}9  & \cellcolor{c1}1  & \cellcolor{c10}10  & \cellcolor{c7}7  & \cellcolor{c8}8  & \cellcolor{c2}2  & \cellcolor{c4}4  & \cellcolor{c6}6  & \cellcolor{c5}5  & \cellcolor{c3}3  \\\hline
IGD$^+$ & \cellcolor{c2}2  & \cellcolor{c9}9  & \cellcolor{c1}1  & \cellcolor{c4}4  & \cellcolor{c3}3  & \cellcolor{c7}7  & \cellcolor{c5}5  & \cellcolor{c8}8  & \cellcolor{c10}10  & \cellcolor{c6}6  \\\hline
R2 & \cellcolor{c2}2  & \cellcolor{c10}10  & \cellcolor{c4}4  & \cellcolor{c1}1  & \cellcolor{c3}3  & \cellcolor{c8}8  & \cellcolor{c5}5  & \cellcolor{c7}7  & \cellcolor{c9}9  & \cellcolor{c6}6  \\\hline
NR2 & \cellcolor{c2}2  & \cellcolor{c10}10  & \cellcolor{c3}3  & \cellcolor{c5}5  & \cellcolor{c1}1  & \cellcolor{c7}7  & \cellcolor{c4}4  & \cellcolor{c8}8  & \cellcolor{c9}9  & \cellcolor{c6}6  \\\hline
$I_{\epsilon+}$ & \cellcolor{c3}3  & \cellcolor{c5}5  & \cellcolor{c6}6  & \cellcolor{c4}4  & \cellcolor{c2}2  & \cellcolor{c1}1  & \cellcolor{c8}8  & \cellcolor{c9}9  & \cellcolor{c10}10  & \cellcolor{c7}7  \\\hline
SE & \cellcolor{c7}7  & \cellcolor{c6}6  & \cellcolor{c4}4  & \cellcolor{c3}3  & \cellcolor{c5}5  & \cellcolor{c8}8  & \cellcolor{c1}1  & \cellcolor{c9}9  & \cellcolor{c10}10  & \cellcolor{c2}2  \\\hline
$\Delta$ & \cellcolor{c9}9  & \cellcolor{c3}3  & \cellcolor{c6}6  & \cellcolor{c4}4  & \cellcolor{c7}7  & \cellcolor{c5}5  & \cellcolor{c2}2  & \cellcolor{c1}1  & \cellcolor{c10}10  & \cellcolor{c8}8  \\\hline
PD & \cellcolor{c7}7  & \cellcolor{c5}5  & \cellcolor{c8}8  & \cellcolor{c3}3  & \cellcolor{c9}9  & \cellcolor{c2}2  & \cellcolor{c6}6  & \cellcolor{c4}4  & \cellcolor{c1}1  & \cellcolor{c10}10  \\\hline
\hline
Avg. & 4.2 & 5.9 & 4.5 & 3.6 & 4.0 & 4.7 & 3.9 & 6.0 & 7.3 & 5.4 \\
\hline
\end{tabularx}
}
%
\end{table*}

\subsubsection{$\vector{A}_{\Delta}$}

Since $\vector{A}_{\Delta}$ contains the extreme objective vectors on all $F$, $\Delta$ can evaluate the spread quality of objective vector sets.
In contrast, the distribution of objective vectors in $\vector{A}_{\Delta}$ is far from uniform.
Thus, $\Delta$ may incorrectly evaluate the uniformity quality of objective vector sets.


$\Delta$ uses a pairwise distance between two closest objective vectors.
As pointed out in \cite{JiangOZF14}, this calculation scheme in $\Delta$ can provide misleading results in some cases.
This is because $\Delta$ does not consider the distance between non-closest objective vectors.
In fact, objective vectors in $\vector{A}_{\Delta}$ for $F_{\rm linear}$ seem to be connected in a chained manner (e.g., see Fig. \ref{fig:linear} (h)).
%
In contrast, SE handles the distance between all pairs of two objective vectors.
Thus, SE does not have this issue in $\Delta$.
It should be noted that the misleading results by $\Delta$ have already been reported in \cite{JiangOZF14,LiY19}.
Thus, this observation is not new.
Our contribution here is that we find and analyze $\vector{A}_{\Delta}$ on the eight front shape functions.
The above-mentioned unintuitive distribution of objective vectors in $\vector{A}_{\Delta}$ has not been investigated in the literature.












 \arrayrulecolor{black}
 \setlength\arrayrulewidth{1.2pt}
 \setlength{\extrarowheight}{1.2pt}
 \newcolumntype{A}{>{\columncolor[rgb]{0.9, 0.9, 0.9}}p{1.6em}}


\subsubsection{$\vector{A}_{\rm PD}$}





Unlike the other quality indicators, PD evaluates the dissimilarity between objective vectors.
For this reason, the distribution of objective vectors in $\vector{A}_{\rm PD}$ is not uniform on all front shape functions.
This observation is consistent with the property of PD reported in \cite{WangJY17,LiY19}.
Also, $\vector{A}_{\rm PD}$ does not contain all the three extreme objective vectors in most cases.



Interestingly, at least two objective vectors in $\vector{A}_{\rm PD}$ overlap.
Since some objective vectors in $\vector{A}_{\rm PD}$ completely overlap on $F_{\rm i\shyp concave}$ and $F_{\rm disconnected}$, it looks like that $\mu$ is less than $21$.
Thus, PD may overestimate an objective vector set that contains similar objective vectors.
Although we calculated the PD value using the translated version of the original source code as explained in Section \ref{sec:experimental_settings}, we can obtain exactly the same results even using the original one.


%


To understand the influence of overlapping objective vectors on PD, we slightly modified one of overlapping objective vectors in $\vector{A}_{\rm PD}$ on $F_{\rm linear}$ (Fig. \ref{fig:ilinear} (i)).
Two resulting objective vector sets are denoted as $\vector{A}_{\rm PD}^{\rm mod1}$ and $\vector{A}_{\rm PD}^{\rm mod2}$.
Fig. S.2 in the supplementary file shows the distribution of objective vectors in $\vector{A}_{\rm PD}^{\rm mod1}$ and $\vector{A}_{\rm PD}^{\rm mod2}$ on $F_{\rm linear}$.
The distribution of objective vectors in $\vector{A}_{\rm PD}^{\rm mod1}$ and $\vector{A}_{\rm PD}^{\rm mod2}$ is more diverse than that in $\vector{A}_{\rm PD}$.
Nevertheless, $\vector{A}_{\rm PD}^{\rm mod1}$ and $\vector{A}_{\rm PD}^{\rm mod2}$ are evaluated as worse than $\vector{A}_{\rm PD}$ as follows: ${\rm PD}(\vector{A}_{\rm PD}) = 1.74 \times 10^5$,  ${\rm PD}(\vector{A}_{\rm PD}^{\rm mod1}) = 1.32 \times 10^5$, and ${\rm PD}(\vector{A}_{\rm PD}^{\rm mod2}) = 1.22 \times 10^5$.
Recall that PD is to be maximized.




Our observation may be due to the sensitivity of PD to the order of objective vectors.
PD constructs a tree based on $\mu$ objective vectors $\vector{a}_1, ..., \vector{a}_{\mu}$ in an objective vector set $\vector{A}$.
In this procedure, each objective vector is greedily linked to its nearest unreplicated objective vector in lexical order (i.e., from $\vector{a}_1$ to $\vector{a}_{\mu}$).
The resulting tree depends on the order of the objective vectors. 
We arranged the objective vectors in $\vector{A}_{\rm PD}$ on $F_{\rm linear}$ in reverse order (i.e., from $\vector{a}_{\mu}$ to $\vector{a}_1$).
The resulting objective vector set is denoted as $\vector{A}_{\rm PD}^{\rm mod3}$.
Although $\vector{A}^{\rm mod3}_{\rm PD}$ and $\vector{A}_{\rm PD}$ contain exactly the same objective vectors, $\vector{A}^{\rm mod3}_{\rm PD}$ is evaluated as worse than $\vector{A}_{\rm PD}$ in terms of PD as follows: ${\rm PD}(\vector{A}_{\rm PD}^{\rm mod3}) = 1.29 \times 10^5$.
Overlapping objective vectors in $\vector{A}_{\rm PD}$ may help to construct a larger tree by implicitly exploiting the lexical characteristic of PD.

\subsection{Ranking information based analysis}
\label{sec:ranking_analysis}




In Subsection \ref{sec:appex_optimal_objs}, we analyzed the nine quality indicators using the optimal $\mu$-distributions.
Here, we examine the nine quality indicators using the ranking information.



Table \ref{tab:concave_ranks} shows the rankings of the nine approximated optimal $\mu$-distributions ($\vector{A}_{\rm HV}$, ..., $\vector{A}_{\rm PD}$) and $\vector{A}_{\rm SLD}$ by each quality indicator on $F_{\rm concave}$.
The average ranking values are reported at the bottom of Table \ref{tab:concave_ranks}.
Tables S.2--S.6 in the supplementary file show the rankings on $F_{\rm linear}$, $F_{\rm convex}$, $F_{\rm i\shyp linear}$, $F_{\rm i\shyp concave}$, and $F_{\rm i\shyp convex}$, respectively.
Since $\vector{A}_{\rm SLD}$ cannot be generated on $F_{\rm disconnected}$ and $F_{\rm c\shyp concave}$, the rankings for $F_{\rm disconnected}$ and $F_{\rm c\shyp concave}$ are not shown.
Note that the quality indicator called ``SLD'' does not exist.
Recall that $\vector{A}_{\rm SLD}$ is generated by simplex-lattice design.
Due to the paper length limitation, we mainly explain the results on $F_{\rm concave}$.


%
On the one hand, from each row in Table \ref{tab:concave_ranks}, we can read the ranking of the ten objective vector sets by the corresponding quality indicator on $F_{\rm concave}$.
For example, HV evaluates $\vector{A}_{\rm HV}$ and $\vector{A}_{\rm IGD}$ as the best and the worst, respectively.
On the other hand, from each column in Table \ref{tab:concave_ranks}, we can read the ranking of the corresponding objective vector set by the nine indicators.
For example, $\vector{A}_{\rm SE}$ is ranked at fourth by HV, IGD, and NR2.
%


The average ranking value of each objective vector set in the bottom row of Table \ref{tab:concave_ranks} (i.e., average value of each column) means how highly the corresponding objective vector set is evaluated by all the nine quality indicators.
$\vector{A}_{\rm R2}$ performs the best among the 10 objective vector sets on $F_{\rm concave}$ in terms of the average ranking values.
Thus, a good objective vector set with respect to R2 may be highly evaluated by the other quality indicators on $F_{\rm concave}$.
In contrast, $\vector{A}_{\rm PD}$ performs the worst in terms of the average ranking values.
This result indicates that a good objective vector set with respect to PD may be evaluated as inferior to other objective vector sets by the other quality indicators on $F_{\rm concave}$.
These results are useful for selecting a quality indicator used in indicator-based EMOAs.
Note that the best objective vector set in terms of the average ranking values depends on the choice of $F$.
For example, Table S.4 shows that $\vector{A}_{\rm R2}$ performs the second worst on $F_{\rm i\shyp linear}$.


%

Each quality indicator $I$ evaluates its optimal $\mu$-distribution $\vector{A}_I$ as the best.
For this reason, all diagonal elements in Table \ref{tab:concave_ranks} are ``1''.
The same results can be found in the rankings on the other front shape functions, except for the ranking of $\vector{A}_{\rm IGD^+}$ by IGD$^+$ on $F_{\rm linear}$ in Table S.2 and $\vector{A}_{\rm NR2}$ by NR2 on $F_{\rm i\shyp concave}$ in Table S.5.
IGD$^+$ and NR2 evaluate $\vector{A}_{\rm IGD}$ and $\vector{A}_{\rm HV}$ as the best in Tables S.2 and S.5, respectively.
These exceptional results are due to the similarity between IGD and IGD$^+$, and the similarity between HV and NR2.


As in \cite{LiefoogheD16}, Table \ref{tab:concave_Kendall_tau_value} shows the rank-based nonlinear Kendall rank correlation $\tau$ values of the nine quality indicator on $F_{\rm concave}$.
Tables S.7--S.11 in the supplementary file show the $\tau$ values of the nine quality indicator on   $F_{\rm linear}$, $F_{\rm convex}$, $F_{\rm i\shyp linear}$, $F_{\rm i\shyp concave}$, and $F_{\rm i\shyp convex}$, respectively.
The $\tau$ value in Table \ref{tab:concave_Kendall_tau_value} represents the similarity of the rankings by two quality indicators $I_1$ and $I_2$ on the 10 objective vector sets ($\vector{A}_{\rm HV}$, ..., $\vector{A}_{\rm SLD}$).
The $\tau$ values in Table \ref{tab:concave_Kendall_tau_value} are symmetric.
The range of the $\tau$ value is $[-1, 1]$.
While the non-negative $\tau$ value means that $I_1$ and $I_2$ are consistent with each other, the negative $\tau$ value means that $I_1$ and $I_2$ are inconsistent with each other.
For example, Table \ref{tab:concave_Kendall_tau_value} shows that HV is inconsistent with IGD ($\tau = -0.60$), but HV is consistent with IGD$^+$ ($\tau = 0.82$).

\begin{table}[t]
  \centering
  \caption{\small Kendall $\tau$ values of the nine indicator on $F_{\rm concave}$.}
  \label{tab:concave_Kendall_tau_value}
        {\scriptsize
  \scalebox{0.89}[1]{
%
    \begin{tabularx}{40em}{|A|C|C|C|C|C|C|C|C|C|}
      \hline
%
      \rowcolor[rgb]{0.9, 0.9, 0.9} & HV & IGD & IGD$^+$ & R2 & NR2 & $I_{\epsilon+}$ & SE & $\Delta$ & PD \\\hline
HV & \cellcolor{c8}1.00 & \cellcolor{c2}-0.60 & \cellcolor{c8}0.82 & \cellcolor{c8}0.78 & \cellcolor{c8}0.96 & \cellcolor{c6}0.33 & \cellcolor{c5}0.24 & \cellcolor{c4}-0.24 & \cellcolor{c3}-0.42 \\\hline
IGD & \cellcolor{c2}-0.60 & \cellcolor{c8}1.00 & \cellcolor{c2}-0.69 & \cellcolor{c2}-0.56 & \cellcolor{c2}-0.56 & \cellcolor{c4}-0.02 & \cellcolor{c4}-0.02 & \cellcolor{c5}0.20 & \cellcolor{c5}0.20 \\\hline
IGD$^+$ & \cellcolor{c8}0.82 & \cellcolor{c2}-0.69 & \cellcolor{c8}1.00 & \cellcolor{c7}0.69 & \cellcolor{c8}0.78 & \cellcolor{c6}0.33 & \cellcolor{c6}0.33 & \cellcolor{c4}-0.16 & \cellcolor{c3}-0.42 \\\hline
R2 & \cellcolor{c8}0.78 & \cellcolor{c2}-0.56 & \cellcolor{c7}0.69 & \cellcolor{c8}1.00 & \cellcolor{c7}0.73 & \cellcolor{c6}0.29 & \cellcolor{c6}0.29 & \cellcolor{c4}-0.11 & \cellcolor{c3}-0.29 \\\hline
NR2 & \cellcolor{c8}0.96 & \cellcolor{c2}-0.56 & \cellcolor{c8}0.78 & \cellcolor{c7}0.73 & \cellcolor{c8}1.00 & \cellcolor{c6}0.38 & \cellcolor{c6}0.29 & \cellcolor{c4}-0.20 & \cellcolor{c3}-0.47 \\\hline
$I_{\epsilon+}$ & \cellcolor{c6}0.33 & \cellcolor{c4}-0.02 & \cellcolor{c6}0.33 & \cellcolor{c6}0.29 & \cellcolor{c6}0.38 & \cellcolor{c8}1.00 & \cellcolor{c4}-0.07 & \cellcolor{c4}-0.11 & \cellcolor{c4}-0.11 \\\hline
SE & \cellcolor{c5}0.24 & \cellcolor{c4}-0.02 & \cellcolor{c6}0.33 & \cellcolor{c6}0.29 & \cellcolor{c6}0.29 & \cellcolor{c4}-0.07 & \cellcolor{c8}1.00 & \cellcolor{c5}0.16 & \cellcolor{c3}-0.47 \\\hline
$\Delta$ & \cellcolor{c4}-0.24 & \cellcolor{c5}0.20 & \cellcolor{c4}-0.16 & \cellcolor{c4}-0.11 & \cellcolor{c4}-0.20 & \cellcolor{c4}-0.11 & \cellcolor{c5}0.16 & \cellcolor{c8}1.00 & \cellcolor{c5}0.11 \\\hline
PD & \cellcolor{c3}-0.42 & \cellcolor{c5}0.20 & \cellcolor{c3}-0.42 & \cellcolor{c3}-0.29 & \cellcolor{c3}-0.47 & \cellcolor{c4}-0.11 & \cellcolor{c3}-0.47 & \cellcolor{c5}0.11 & \cellcolor{c8}1.00 \\\hline
\end{tabularx}
}
}
\end{table}

As seen from Table \ref{tab:concave_ranks}, no optimal $\mu$-distribution is ranked as the best by multiple quality indicators due to the conflicting properties of the nine quality indicators.
For example, $\vector{A}_{\rm IGD}$ is ranked as the best by IGD but the worst by HV, R2, and NR2.
%
In contrast, the rankings by HV and NR2 are almost the same on all front shape functions (except for the rankings of $\vector{A}_{\rm HV}$ and $\vector{A}_{\rm NR2}$).
In fact, the $\tau$ value of HV and NR2 in Table \ref{tab:concave_Kendall_tau_value} is $0.96$, which indicates HV and NR2 are highly correlated with each other.
Although the rankings by IGD and IGD$^+$ are almost the same on the linear Pareto fronts ($F_{\rm linear}$ and $F_{\rm i\shyp linear}$), they are dissimilar on the non-linear Pareto fronts ($F_{\rm concave}$, $F_{\rm convex}$, $F_{\rm i\shyp concave}$, and $F_{\rm i\shyp convex}$).
The $\tau$ value of IGD and IGD$^+$ in Table \ref{tab:concave_Kendall_tau_value} is also $-0.60$ on $F_{\rm concave}$.
These results of IGD and IGD$^+$ are consistent with \cite{BezerraLS17,IshibuchiIMN19}.




The simplex lattice-based $\vector{A}_{\rm SLD}$ can be viewed as the optimal objective vector set obtained by decomposition-based EMOAs.
Since $\vector{A}_{\rm SLD}$ has a good uniformity, $\vector{A}_{\rm SLD}$ is evaluated as the second best by SE on $F_{\rm concave}$.
In contrast, $\vector{A}_{\rm SLD}$ is evaluated as ``poor'' by the other quality indicators in most cases  (except for the results on $F_{\rm i\shyp  linear}$ in Table S.4).
In fact, the distribution of objective vectors in $\vector{A}_{\rm SLD}$ is not identical to optimal $\mu$-distributions for the nine quality indicators (see Figs. \ref{fig:linear}--\ref{fig:iconvex}).
The poor results of $\vector{A}_{\rm SLD}$ suggest that decomposition-based EMOAs need an adaptive weight vector method considering the optimal $\mu$-distributions for a given quality indicator.

\subsection{Analysis for $m \geq 4$}
\label{sec:m5_8}



Since it is almost impossible to clearly show the distribution of objective vectors of a problem with $m \geq 4$ {\em in an understandable manner} \cite{TusarF15,LiZY17}, this paper focused on $m=3$.
We reemphasize that our analysis for $m=3$ itself is already an important contribution to the EMO community.
Nevertheless, it is interesting to investigate whether the results for $m=3$ can be generalized to the case of $m \geq 4$.

Since the distribution of objective vectors for $m \geq 4$ is unclear, our analysis here is based on the ranking information as in Subsection \ref{sec:ranking_analysis}.
Tables S.12--S.23 in the supplementary file show the rankings of the nine approximated optimal $\mu$-distributions ($\vector{A}_{\rm HV}$, ..., $\vector{A}_{\rm PD}$) with $\mu=21$ by each quality indicator on six front shape functions ($F_{\rm linear}$, $F_{\rm concave}$, $F_{\rm convex}$, $F_{\rm i\shyp linear}$, $F_{\rm i\shyp concave}$, and $F_{\rm i\shyp convex}$) with $m=5$ and $8$, respectively.
Since $\vector{A}_{\rm SLD}$ with $\mu = 21$ cannot be generated for $m=5$ and $8$, $\vector{A}_{\rm SLD}$ is removed from this experiment.
The size of the reference point set $\vector{R}$ was set to $3\,060$ for $m=5$ and $5\,148$ for $m=8$.
We generated $\vector{R}$ for $m=8$ using the two-layered version of simplex-lattice design \cite{DebJ14}.
Tables S.24--S.35 in the supplementary file also show the Kendall rank correlation $\tau$ values of the nine quality indicators on the six front shape functions with $m=5$ and $8$, respectively.


Although most results for $m \geq 4$ are consistent with the results for $m=3$, some exceptional results can be found.
$\vector{A}_{\rm SE}$ is ranked as the best by $\Delta$ on $F_{\rm linear}$ and $F_{\rm i\shyp linear}$ with $m=8$, and $F_{\rm convex}$ and $F_{\rm i\shyp concave}$ with $m \in \{5, 8\}$.
This means that a better approximated optimal $\mu$-distribution for $\Delta$ can be found by optimizing SE.
Similarly, $\vector{A}_{\rm R2}$ is ranked as the best by $I_{\epsilon+}$ on $F_{\rm i\shyp concave}$ with $m \in \{5, 8\}$.
This may be because of the problem difficulty in finding the optimal $\mu$-distribution for $\Delta$ and $I_{\epsilon+}$.
For example, as mentioned in Subsection \ref{sec:experimental_results}, the $I_{\epsilon+}$ value of $\vector{A}$ depends only on one objective value of one objective vector in $\vector{A}$.
As in the Schwefel 2.21 function $f(\vector{x}) = \max_{i \in \{1, ..., n\}} \{|x_i|\}$ \cite{YaoLL99}, this property of $I_{\epsilon+}$ can incorporate a strong nonseparability between variables $\theta_1, ..., \theta_d$ in the proposed formulation.
In addition, this property can make some areas of the fitness landscape of the proposed formulation plateaus.
These problem difficulties become pronounced with the increase of $m$.
A landscape analysis of the proposed formulation for each quality indicator is needed for deeper understanding.


Table \ref{tab:best_avg_objvec} shows the best objective vector sets on each $F$ with $m \in \{3, 5, 8\}$ in terms of the average ranking.
We can see that the best objective vector set in terms of the average ranking depends on $m$ in addition to the choice
of $F$.
For example, $\vector{A}_{\rm R2}$, $\vector{A}_{\rm NR2}$, and $\vector{A}_{\rm IGD^+}$ perform the best on $F_{\rm concave}$ with $m=3$, $5$, and $8$, respectively.
These results indicate the necessity of an adaptive indicator selection mechanism in indicator-based EMOAs (e.g., \cite{Falcon-CardonaC18}).

\begin{table}[t]
\begin{center}
  \caption{\small Best objective vector sets on each front shape function with $m \in \{3, 5, 8\}$ in terms of the average ranking.}
{\scriptsize
  \label{tab:best_avg_objvec}
\scalebox{0.9}[1]{ 
\begin{tabular}{lllllllll}
\toprule
  %
& $F_{\rm linear}$ & $F_{\rm concave}$ & $F_{\rm convex}$ & $F_{\rm i\shyp linear}$ & $F_{\rm i\shyp concave}$ & $F_{\rm i\shyp convex}$\\
\midrule
$3$ & $\vector{A}_{\rm HV}$, $\vector{A}_{\rm NR2}$ & $\vector{A}_{\rm R2}$ & $\vector{A}_{\rm HV}$ & $\vector{A}_{\rm HV}$ & $\vector{A}_{\rm HV}$ & $\vector{A}_{\rm HV}$\\
\midrule
$5$ & $\vector{A}_{\rm NR2}$, $\vector{A}_{\Delta}$ & $\vector{A}_{\rm NR2}$ & $\vector{A}_{\rm HV}$ & $\vector{A}_{\rm NR2}$ & $\vector{A}_{\rm HV}$ & $\vector{A}_{\rm IGD^+}$\\
\midrule
$8$ & $\vector{A}_{\Delta}$ & $\vector{A}_{\rm IGD^+}$ & $\vector{A}_{\rm IGD}$ & $\vector{A}_{\rm IGD^+}$ & $\vector{A}_{\rm HV}$ & $\vector{A}_{\rm IGD^+}$\\[0.2em]
\toprule
\end{tabular}
}
}
\end{center}
\end{table}

In addition, the correlation between HV and NR2 becomes weak with the increase of $m$.
For example, the $\tau$ values of HV and NR2 on $F_{\rm linear}$ are $0.96$ for $m=3$, $0.89$ for $m=5$, and $0.22$ for $m=8$ (see Tables \ref{tab:concave_Kendall_tau_value}, S.24, and S.30, respectively).
This observation indicates that the approximation performance of NR2 is likely to deteriorate with the increase of $m$.

In summary, the results here show that some results for $m=3$ cannot be always generalized to the case of $m \geq 4$.
Thus, our observations are only for $m=3$ as emphasized in the title of this paper.
Further analysis for many-objective problems is need in future research.

\subsection{Analysis of a unary version of an $M$-nary indicator}
\label{sec:mary}

As in $I_{\epsilon+}$, some binary or $M$-nary quality indicators can be converted into their unary versions by using the reference vector set $\vector{R}$ as the compared vector set.
However, it is questionable that such a converted unary quality indicator works well.
Here, we demonstrate that the proposed formulation can be useful to analyze the unary version of an $M$-nary indicator.

Since a number of $M$-nary quality indicators have been proposed, it is very difficult to investigate all of them.
Instead, we selected the diversity comparison indicator (DCI) \cite{LiYL14dci} since it is recommended in \cite{LiY19}.
DCI relatively compares $M$ objective vector sets $\vector{A}_1$, ..., $\vector{A}_M$.
A large DCI value indicates that the corresponding $\vector{A}$ has a good diversity among $\vector{A}_1$, ..., $\vector{A}_M$.
We explain DCI in Section S.4 in the supplementary file.
For the DCI calculation, we used the source code provided by the authors of \cite{LiYL14dci}.

We converted DCI into the unary indicator using the reference vector set $\vector{R}$.
To generate $\vector{R}$, we used the same method described in Section \ref{sec:experimental_settings}.
Thus, the DCI value of $\vector{A}$ is obtained by comparing $\vector{A}$ to $\vector{R}$ by DCI.
As recommend in \cite{LiYL14dci}, we set $div$ in DCI to $19$ for $m=3$ in this study.
Since we want to examine the general property of DCI, we did not finely tune $div$ for each $F$.
We set $\mu$ to 21.
We investigate the influence of the size of $\vector{R}$ on DCI.
Below, we denote DCI with $|\vector{R}| = 21$ and $|\vector{R}| = 28$ as DCI21 and DCI28, respectively.

Fig. \ref{fig:dci} shows the distribution of objective vectors in $\vector{A}_{\rm DCI21}$ and $\vector{A}_{\rm DCI28}$ on $F_{\rm linear}$.
Figs. S.51--S.55 in the supplementary file show $\vector{A}_{\rm DCI21}$ (and $\vector{A}_{\rm DCI28}$) on $F_{\rm concave}$, $F_{\rm convex}$, $F_{\rm i\shyp linear}$, $F_{\rm i\shyp concave}$, and $F_{\rm i\shyp convex}$, respectively.
On the one hand, Fig. \ref{fig:dci} (a) shows that the objective vectors in $\vector{A}_{\rm DCI21}$ are uniformly distributed.
Thus, DCI21 is likely to prefer uniformly distributed objective vectors.
In fact, ${\rm DCI21}(\vector{A}_{\rm DCI21})=1$ and ${\rm DCI21}(\vector{A}_{\rm SLD})=1$, where $\vector{A}_{\rm SLD}$ is shown in Fig. \ref{fig:linear} (j).
On the other hand, Fig. \ref{fig:dci} (b) shows that the distribution of the objective vectors in $\vector{A}_{\rm DCI28}$ is not uniform.
This result means that DCI28 does not prefer uniformly distributed objective vectors.
Actually, DCI28 evaluates $\vector{A}_{\rm DCI28}$ as better than $\vector{A}_{\rm SLD}$ as follows: ${\rm DCI28}(\vector{A}_{\rm DCI28})=0.75$ and ${\rm DCI28}(\vector{A}_{\rm SLD})=0.67$.

In summary, we can conclude that it is possible to design a unary version of DCI, but its result is sensitive to the size of $\vector{R}$.
In the unary version of DCI, the size of $\vector{R}$ and $\mu$ should be the same.
Otherwise, non-uniformly distributed objective vectors are highly evaluated by DCI (as shown in Fig. \ref{fig:dci} (b)).

%

\begin{figure}[t]
\newcommand{\widthvar}{0.15}
\centering
\subfloat[$\vector{A}_{\rm DCI21}$]{\includegraphics[width=\widthvar\textwidth]{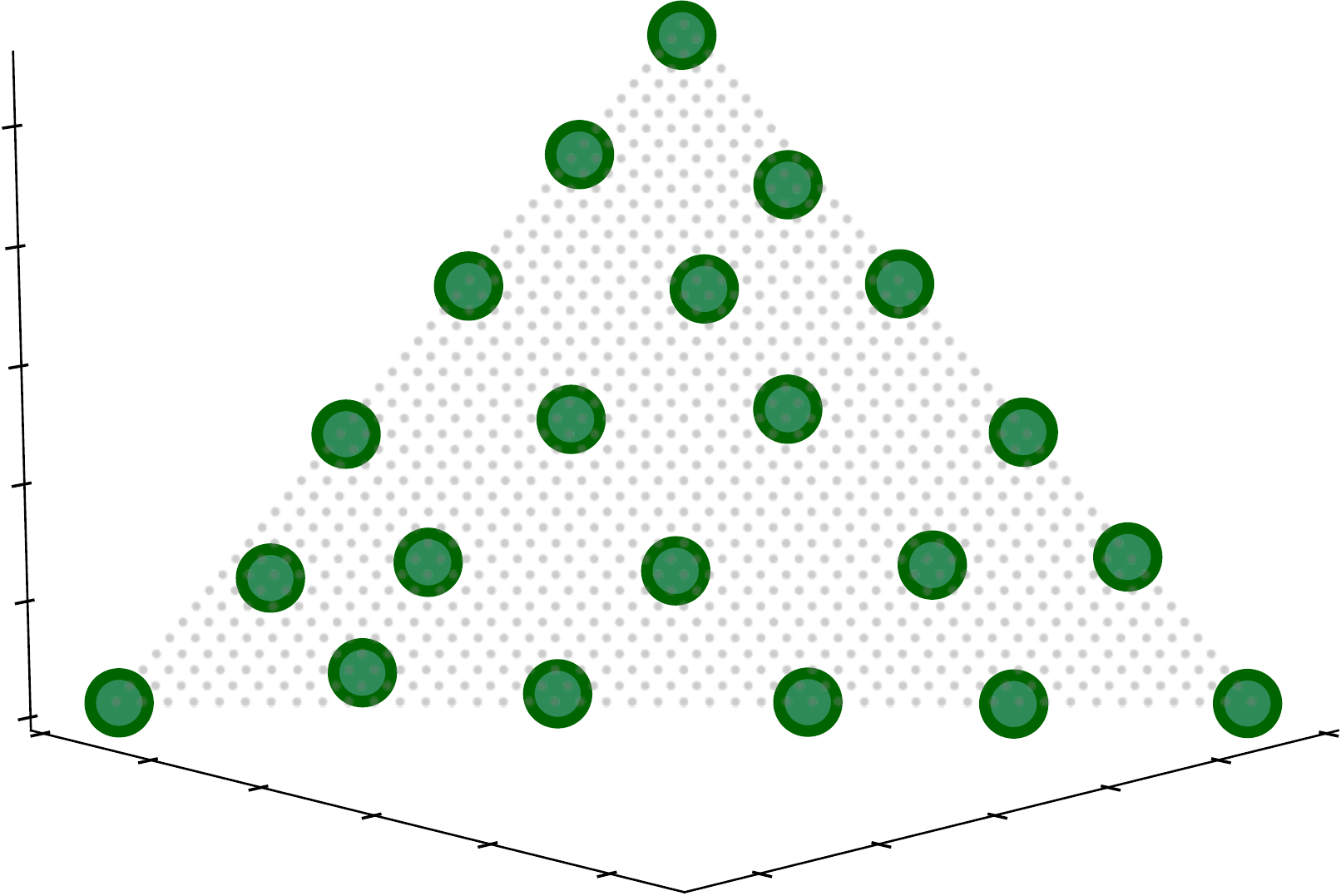}}
\subfloat[$\vector{A}_{\rm DCI28}$]{\includegraphics[width=\widthvar\textwidth]{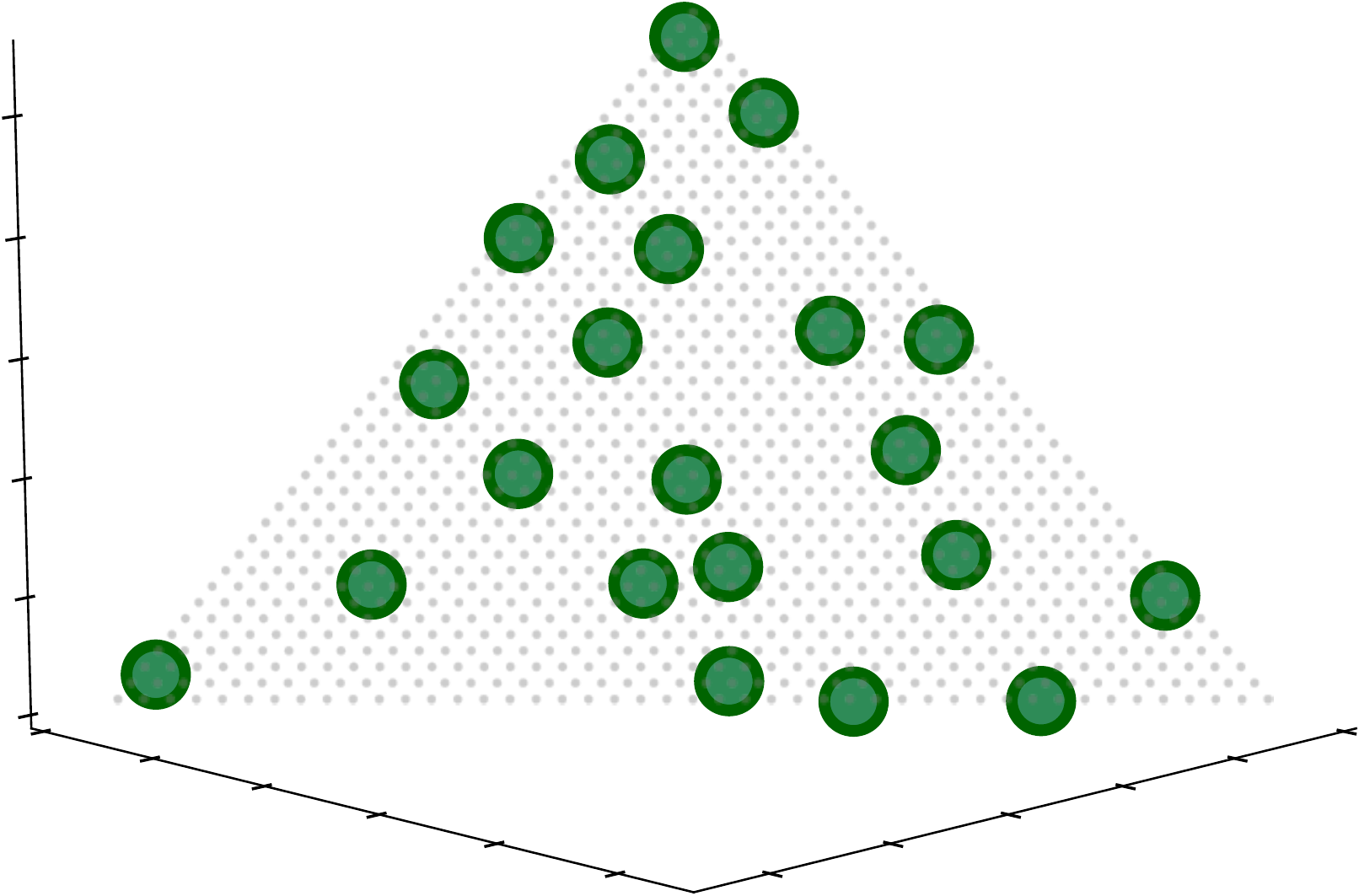}}
\caption{
\small
$\vector{A}_{\rm DCI21}$ and $\vector{A}_{\rm DCI28}$ on $F_{\rm linear}$.
}
\label{fig:dci}
\end{figure}

\section{Conclusion}
\label{sec:conclusion}

We have analyzed the nine quality indicators using their approximated optimal $\mu$-distributions for $m=3$.
First, we proposed the problem formulation of finding the optimal $\mu$-distribution for each quality indicator on Pareto fronts for arbitrary number of objectives ($m \geq 2$).
Then, we approximated the optimal $\mu$-distributions for the nine quality indicators on the Pareto fronts of eight problems with $m=3$.
We analyzed the nine quality indicators based on the optimal $\mu$-distribution based approach and the ranking information based approach.
We also examined the generality of results for $m=3$ with respect to $m$ and a unary version of an $M$-nary indicator.

Our observations for $m=3$ can be summarized as follows:



\begin{enumerate}
\renewcommand{\labelenumi}{\roman{enumi})}
\item Objective vectors in the approximated optimal $\mu$-distributions for R2 are biased on the edge of the Pareto front for $F_{\rm i\shyp linear}$ and $F_{\rm i\shyp concave}$. These results for $m=3$ are inconsistent with the results for $m=2$.
\item The approximated optimal $\mu$-distributions for NR2 and HV are similar. NR2 and HV are correlated with each other in terms of the Kendall rank correlation ($\tau = 0.96$ on $F_{\rm concave}$). Thus, NR2 can substitute for HV.
\item While the optimal $\mu$-distributions for $I_{\epsilon+}$ for $m=2$ are almost uniform, those for $m=3$ are not uniform even on the linear front.
\item Although SE can evaluate both uniformity and spread quality of objective vector sets, SE prefers objective vectors that are on the edge of the Pareto front.
\item Since the optimal $\mu$-distributions for $\Delta$ are far from uniform, $\Delta$ may incorrectly evaluate the uniformity quality of objective vector sets.
\item PD may overestimate an objective vector set with a small dissimilarity. PD is also sensitive to the order of objective vectors.
\item Most quality indicators evaluate objective vectors generated by simplex-lattice design as poor. Thus, decomposition-based EMOAs need an adaptive weight vector method when they try to search for a good objective vector set with respect to each quality indicator. 
\item A unary version of DCI is sensitive to the size of the reference vector set.
\end{enumerate}




We emphasize that the above-mentioned observations on the Pareto front for $m=3$ could not be obtained without searching for the optimal distributions of objective vectors by the proposed problem formulation.
As shown in Subsection \ref{sec:m5_8}, our results for $m=3$ cannot be always generalized to the case of $m \geq 4$.
Further analysis is needed for many-objective problems in future research.

A further analysis of the optimal $\mu$-distributions for R2 and $I_{\epsilon+}$ on the Pareto front for $m=3$ is needed.
The undesirable property of PD could be addressed by using an algorithm for finding a minimum spanning tree (e.g., \cite{Chazelle00a}).
The problem of finding the optimal $\mu$-distribution in the proposed problem formulation itself can be viewed as a ``real-world'' black-box single-objective optimization problem.
Although we used L-SHADE in this paper, we believe that benchmarking various single-objective optimizers on the proposed problem formulation is beneficial for both the EMO community and the evolutionary single-objective optimization community.

\section*{Acknowledgment}

This work was supported by National Natural Science Foundation of China (Grant No. 61876075), the Program for Guangdong Introducing Innovative and Enterpreneurial Teams (Grant No. 2017ZT07X386), Shenzhen Peacock Plan (Grant No. KQTD2016112514355531), the Science and Technology Innovation Committee Foundation of Shenzhen (Grant No. ZDSYS201703031748284), the Program for University Key Laboratory of Guangdong Province (Grant No. 2017KSYS008).

\ifCLASSOPTIONcaptionsoff
  \newpage
\fi



%



\bibliography{reference}
\bibliographystyle{IEEEtran}

\begin{IEEEbiography}[{\includegraphics[width=1in,height=1.25in,keepaspectratio]{graph/tanabe.pdf}}]{Ryoji Tanabe}
is a Research Assistant Professor with Department of Computer Science and Engineering, Southern University of Science and Technology, China, from 2017 to 2019.
He was a Post-Doctoral Researcher with ISAS/JAXA, Japan, from 2016 to 2017.
He received his Ph.D. in Science from The University of Tokyo, Japan, in 2016.
His research interests include stochastic single- and multi-objective optimization algorithms, parameter control in evolutionary algorithms, and automatic algorithm configuration.
  \end{IEEEbiography}



\begin{IEEEbiography}[{\includegraphics[width=1in,height=1.25in,keepaspectratio]{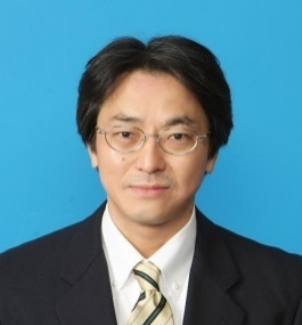}}]{Hisao Ishibuchi}  (M'93-SM'10-F'14)
  received the B.S. and M.S. degrees in precision mechanics from Kyoto University, Kyoto, Japan, in 1985 and 1987, respectively, and the Ph.D. degree in computer science from Osaka Prefecture University, Sakai, Osaka, Japan, in 1992. He was with Osaka Prefecture University in 1987-2017. Since 2017, he is a Chair Professor at Southern University of Science and Technology, China. His research interests include fuzzy rule-based classifiers, evolutionary multi-objective and many-objective optimization, memetic algorithms, and evolutionary games.

Dr. Ishibuchi was the IEEE Computational Intelligence Society (CIS) Vice-President for Technical Activities (2010-2013), an AdCom member of the IEEE CIS (2014-2019), and the Editor-in-Chief of the IEEE COMPUTATIONAL INTELLIGENCE MAGAZINE (2014-2019).
\end{IEEEbiography}

%








\end{document}